\documentclass[10pt,twocolumn,letterpaper]{article}

\usepackage[pagenumbers]{iccv}              % To produce the CAMERA-READY version

\definecolor{iccvblue}{rgb}{0.21,0.49,0.74}
\usepackage[pagebackref,breaklinks,colorlinks,allcolors=iccvblue]{hyperref}

\usepackage[accsupp]{axessibility}
\usepackage{wrapfig}
\usepackage{comment}
\usepackage[permil]{overpic}
\usepackage{float}
\usepackage{xspace}
\usepackage{graphicx}
\usepackage{amsmath}
\usepackage{utfsym}
\usepackage{bm}
\usepackage{caption}
\usepackage{colortbl}
\definecolor{best1}{RGB}{255,152,152}
\definecolor{best2}{RGB}{255,203,152}
\definecolor{best3}{RGB}{255,247,173}

\def\paperID{10617} % *** Enter the Paper ID here
\def\confName{ICCV}
\def\confYear{2025}

\title{GeoSplatting: Towards Geometry Guided Gaussian Splatting for Physically-based Inverse Rendering}

\author{
\vspace{0.5mm}
{Kai Ye$^{1,*}$, \quad Chong Gao$^{2,*}$, \quad Guanbin Li$^{2}$, \quad Wenzheng Chen$^{3,4,\dag}$, \quad Baoquan Chen$^{1,5,\dag}$} \\
\small{$^1$School of Intelligence Science and Technology, Peking University}\\
\small{$^2$School of Computer Science and Engineering, Sun Yat-sen University}\\
\small{$^3$Wangxuan Institute of Computer Technology, Peking University} \\
\small{$^4$State Key Laboratory of Multimedia Information Processing, Peking University} \\
\small{$^5$State Key Laboratory of General Artificial Intelligence, Peking University}\\
\vspace{-9mm}
}

\begin{document}

\newcommand{\name}{GeoSplatting}
\newcommand{\yaoyao}{R3DG}
\newcommand{\gir}{GIR}
\newcommand{\gaussianshader}{GS-Shader}
\newcommand{\sugar}{SuGaR}
\newcommand{\meshsampler}{MGadapter}

\newcommand{\wz}[1]{\textcolor{red}{\textsf{[wz: #1]}}}
\newcommand{\gc}[1]{\textcolor{green}{\textsf{[gc: #1]}}}
\newcommand{\yk}[1]{\textcolor{blue}{\textsf{[yk: #1]}}}

\makeatletter
\DeclareRobustCommand\onedot{\futurelet\@let@token\@onedot}
\def\@onedot{\ifx\@let@token.\else.\null\fi\xspace}

\def\eg{\emph{e.g}\onedot} \def\Eg{\emph{E.g}\onedot}
\def\ie{\emph{i.e}\onedot} \def\Ie{\emph{I.e}\onedot}
\def\cf{\emph{cf}\onedot} \def\Cf{\emph{Cf}\onedot}
\def\etc{\emph{etc}\onedot} \def\vs{\emph{vs}\onedot}
\def\wrt{w.r.t\onedot} \def\dof{d.o.f\onedot}
\def\iid{i.i.d\onedot} \def\wolog{w.l.o.g\onedot}
\def\etal{\emph{et al}\onedot}
\makeatother

\newcommand{\point}{\mathbf{x}}
\newcommand{\threeDPosition}{\point}

\newcommand{\threeDGaussian}{{G}}
\newcommand{\centerPos}{\boldsymbol{\mu}}
\newcommand{\centerPosScalar}{{\mu}}
\newcommand{\threeDCov}{\bm{\Sigma}}
\newcommand{\threeDScaling}{\mathbf{S}}
\newcommand{\threeDRotation}{\mathbf{R}}

\newcommand{\opacity}{o}
\newcommand{\gaussianColor}{c}
\newcommand{\gaussianColorDiffuse}{\gaussianColor^d}
\newcommand{\gaussianColorSpecular}{\gaussianColor^s}
\newcommand{\gaussianColorResidual}{\gaussianColor^r}
\newcommand{\gaussianMovement}{v}

\newcommand{\weightBlending}{\alpha}

\newcommand{\sdfValue}{s}
\newcommand{\sdfFunc}{\bm{\zeta}}
\newcommand{\meshSamplerFunc}{\mathcal{T}}

\newcommand{\network}{F}%\mathsf{F}}
\newcommand{\networkParam}{\bm{\vartheta}}

\newcommand{\mesh}{\mathbf{M}}
\newcommand{\face}{\mathbf{F}}

\newcommand{\eqcomma}{\ \ ,}
\newcommand{\eqstop}{\ \ .}

\newcommand{\materialDiffuse}{\mathbf{k}_\mathrm{d}}
\newcommand{\materialSpecular}{\mathbf{k}_\mathrm{s}}
\newcommand{\materialRoughness}{\rho}
\newcommand{\materialMetalness}{m}
\newcommand{\materialResidualSH}{s}

\newcommand{\image}{I}
\newcommand{\mask}{M}
\newcommand{\loss}{\mathcal{L}}

\newcommand{\yes}{\usym{1F5F8}}
\newcommand{\no}{\usym{2613}}

\newcommand{\dir}{\bm{\omega}}
\newcommand{\dirLight}{\dir_L}
\newcommand{\dirView}{\dir_V}
\newcommand{\normal}{\mathbf{n}}
\newcommand{\dirNormal}{\normal}
\newcommand{\dirReflect}{\dir_R}
\newcommand{\shininess}{\alpha}

\newcommand{\x}{\mathbf{x}}
\newcommand{\wi}{\dir_\mathrm{i}}
\newcommand{\wo}{\dir_\mathrm{o}}
\newcommand{\wh}{\dir_\mathrm{h}}
\newcommand{\Li}{\mathbf{L}_\mathrm{i}}
\newcommand{\Le}{\mathbf{L}_\mathrm{e}}
\newcommand{\Lind}{\mathbf{L}_\mathrm{ind}}
\newcommand{\Ldir}{\mathbf{L}_\mathrm{dir}}
\newcommand{\Lo}{\mathbf{L}_\mathrm{o}}
\newcommand{\thetai}{\theta_\mathrm{i}}
\newcommand{\dd}{\mathrm{d}}
\newcommand{\brdf}{\mathbf{f}_\mathrm{r}}
\newcommand{\hemis}{{\mathcal{H}^2}}
\newcommand{\sphere}{{\mathcal{S}^2}}
\newcommand{\Omesh}{O_\mathrm{mesh}}
\newcommand{\Ogs}{O_\mathrm{3dgs}}

\newcommand{\albedo}{a}
\newcommand{\diffAlbedo}{\mathbf{a}}
\newcommand{\ndf}{D}
\newcommand{\fresnel}{F}
\newcommand{\geom}{G}
\newcommand{\microfacet}{M}

\newcommand{\envmap}{\Li}

\twocolumn[{
\renewcommand\twocolumn[1][]{#1}
\maketitle
\begin{center}
    \vspace*{-6.5mm}
    \centering
    \includegraphics[width=0.94\linewidth,trim={2.0cm 3.3cm 3.5cm 3.2cm},clip]{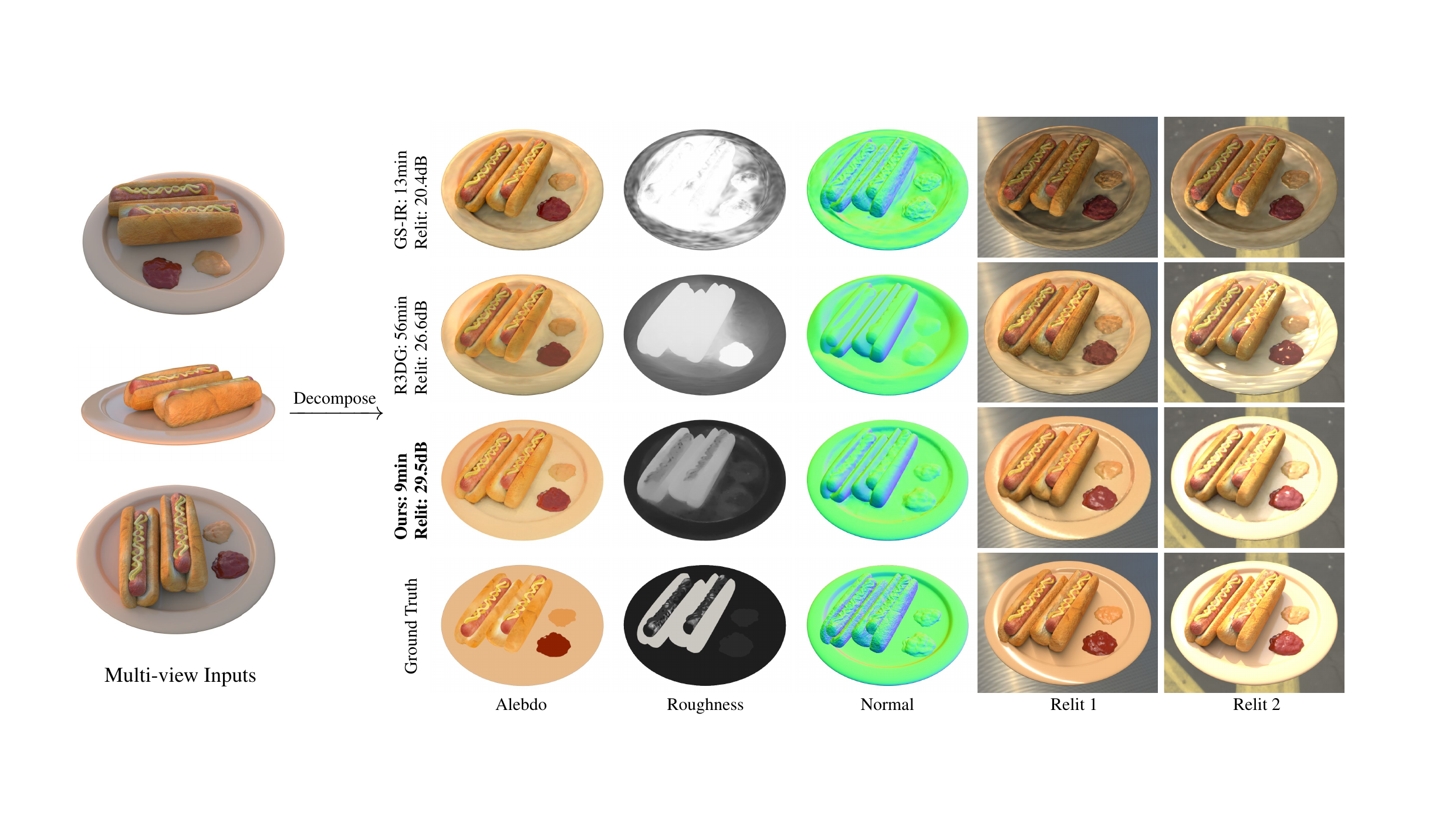}
    \vspace*{-3mm}
    \captionof{figure}{
        \small{
            \textbf{Decomposition and Relighting Results.} Existing 3DGS-based inverse rendering methods, such as Relightable 3DGS (R3DG)~\cite{R3DG2023} and GS-IR~\cite{liang2024gs}, typically rely on approximated normal directions, which can lead to inaccuracies in light transport modeling and subsequently result in noisy material decomposition and flawed relighting results. In contrast, we propose {\name}, a novel approach that augments 3DGS with explicit geometry guidance to improve normal estimation and light transport modeling, resulting in superior decomposition and relighting quality, while also delivering excellent training efficiency.
            }
        \label{fig:teaser}
    }
    \vspace{1mm}
\end{center}
}]

\renewcommand{\thefootnote}{\fnsymbol{footnote}}
\footnotetext[1]{Equal contribution. $^\dag$Equal advisory.}

\begin{abstract}
Recent 3D Gaussian Splatting (3DGS) representations~\cite{kerbl3Dgaussians} have demonstrated remarkable performance in novel view synthesis; further, material-lighting disentanglement on 3DGS warrants relighting capabilities and its adaptability to broader applications. While the general approach to the latter operation lies in integrating differentiable physically-based rendering (PBR) techniques to jointly recover BRDF materials and environment lighting, achieving a precise disentanglement remains an inherently difficult task due to the challenge of accurately modeling light transport. Existing approaches typically approximate Gaussian points' normals, which constitute an implicit geometric constraint. However, they usually suffer from inaccuracies in normal estimation that subsequently degrade light transport, resulting in noisy material decomposition and flawed relighting results.
To address this, we propose {\name}, a novel approach that augments 3DGS with explicit geometry guidance for precise light transport modeling. By differentiably constructing a surface-grounded 3DGS from an optimizable mesh, our approach leverages well-defined mesh normals and the opaque mesh surface, and additionally facilitates the use of mesh-based ray tracing techniques for efficient, occlusion-aware light transport calculations. This enhancement ensures precise material decomposition while preserving the efficiency and high-quality rendering capabilities of 3DGS.
Comprehensive evaluations across diverse datasets demonstrate the effectiveness of {\name}, highlighting its superior efficiency and state-of-the-art inverse rendering performance.
The project page can be found at \url{https://pku-vcl-geometry.github.io/GeoSplatting/}.
\end{abstract}
\vspace*{-12pt}
\section{Introduction}

3D reconstruction, fueled by powerful representations~\cite{mildenhall2020nerf,kerbl20233d}, has achieved significant progress in recent years.
While the reconstructed objects exhibit detailed geometry and photorealistic appearance, enabling striking renderings from novel viewpoints~\cite{wang2021neus,barron2022mip}, the lack of material and lighting decomposition limits their adaptability to varying lighting conditions. 
Consequently, traditional 3D reconstruction methods fall short of producing versatile 3D assets suitable for downstream applications such as gaming, filmmaking, and AR/VR.
To address this limitation, several approaches grounded in Physically-Based Rendering (PBR) theory~\cite{kajiya1986} extend beyond traditional appearance recovery to material-lighting disentanglement, framing this problem as physically-based inverse rendering~\cite{zhang2021nerfactor, boss2021nerd, R3DG2023, jiang2023gaussianshader}.

Existing approaches tackle inverse rendering by combining differentiable PBR techniques with powerful 3D representations, enabling joint estimation of material and lighting through gradient descent. Earlier works rely on explicit 3D representations, such as triangular meshes, to seamlessly integrate with differentiable PBR pipelines~\cite{Munkberg_2022_CVPR,hasselgren2022nvdiffrecmc}. However, these approaches can suffer from less robust optimization, as mesh-based differentiable rendering methods typically generate gradients only at triangle edges~\cite{Laine2020diffrast}. In contrast, the success of NeRF and NeuS in novel view synthesis has inspired efforts to combine PBR techniques with neural fields, leading to more robust inverse rendering. Nevertheless, due to the heavy ray sampling involved in their volumetric rendering process, these methods are computationally expensive and require hours to converge~\cite{liu2023nero,yang2023sire}.

More recently, 3D Gaussian Splatting (3DGS) \cite{kerbl3Dgaussians} has emerged as a powerful 3D representation, offering both high-quality novel-view synthesis and impressive efficiency.
However, the lack of explicit geometry boundaries and well-defined surface normals prevents vanilla 3DGS from accurately modeling light transport, thereby hindering effective material-lighting disentanglement. 
Motivated by geometry enhancement techniques for 3DGS~\cite{guedon2023sugar, Huang2DGS2024, dai2024high,lyu20243dgsr, yu2024gsdf3dgsmeetssdf, xiang2024gaussianroomimproving3dgaussian, Chen2024PGSRPG}, several approaches have attempted to extend its use to inverse rendering tasks. A common strategy involves squashing 3DGS points into surfels, from which surface normals can be approximated via implicit geometric constraints, such as depth-normal regularization~\cite{jiang2023gaussianshader, liang2024gs, shi2023gir, R3DG2023}.
Nevertheless, accurately modeling light transport requires both precise normal directions and opaque surfaces, which define light propagation directions and light-surface intersections, respectively. 
As a result, existing 3DGS-based inverse methods—relying on approximated normals and the inherently semi-transparent nature of Gaussian points—typically struggle with light transport modeling, leading to noisy material decompositions and flawed relighting.

To address these challenges in 3DGS-based inverse rendering frameworks, we propose {\name}, a principled solution that augments 3DGS with explicit geometry guidance. 
Unlike existing 3DGS-based inverse rendering methods that rely on implicit geometric constraints to iteratively approximate Gaussian point normals, {\name} differentiably constructs surface-aligned Gaussian points from an optimizable, explicit mesh. Consequently, {\name} inherently leverages the well-defined mesh normals and the opaque mesh surface for more accurate light transport modeling, resulting in superior material-lighting disentanglement and improved relighting quality.

Specifically, we incorporate isosurface techniques~\cite{shen2023flexicubes} to utilize an optimizable triangular mesh as the geometry guidance. By introducing a Mesh-to-Gaussian adaptor ({\meshsampler}) that differentiably constructs structured Gaussian points grounded on the mesh geometry, we can leverage the explicit mesh normals and the opaque mesh surface to enhance 3DGS with accurate light transport modeling. The fully differentiable nature of {\meshsampler} allows for end-to-end optimization of the geometry guidance during training, ensuring consistency between the mesh and 3DGS. Such consistency further facilitates the use of mesh-based ray tracing techniques for efficient light transport calculations, delivering excellent optimization efficiency while effectively accounting for shadow effects and inter-reflection.

We conduct extensive experiments to validate the effectiveness of our method, demonstrating that {\name} significantly outperforms previous 3DGS-based inverse rendering baselines in terms of geometry accuracy and material-lighting disentanglement while delivering exceptional efficiency compared to both 3DGS-based and implicit-field-based inverse rendering methods.

\begin{figure*}
	\vspace*{-0.5cm}
	\begin{center}
		\includegraphics[width=0.98\textwidth,trim={0.5cm 1.0cm 0.2cm 0.5cm},clip]{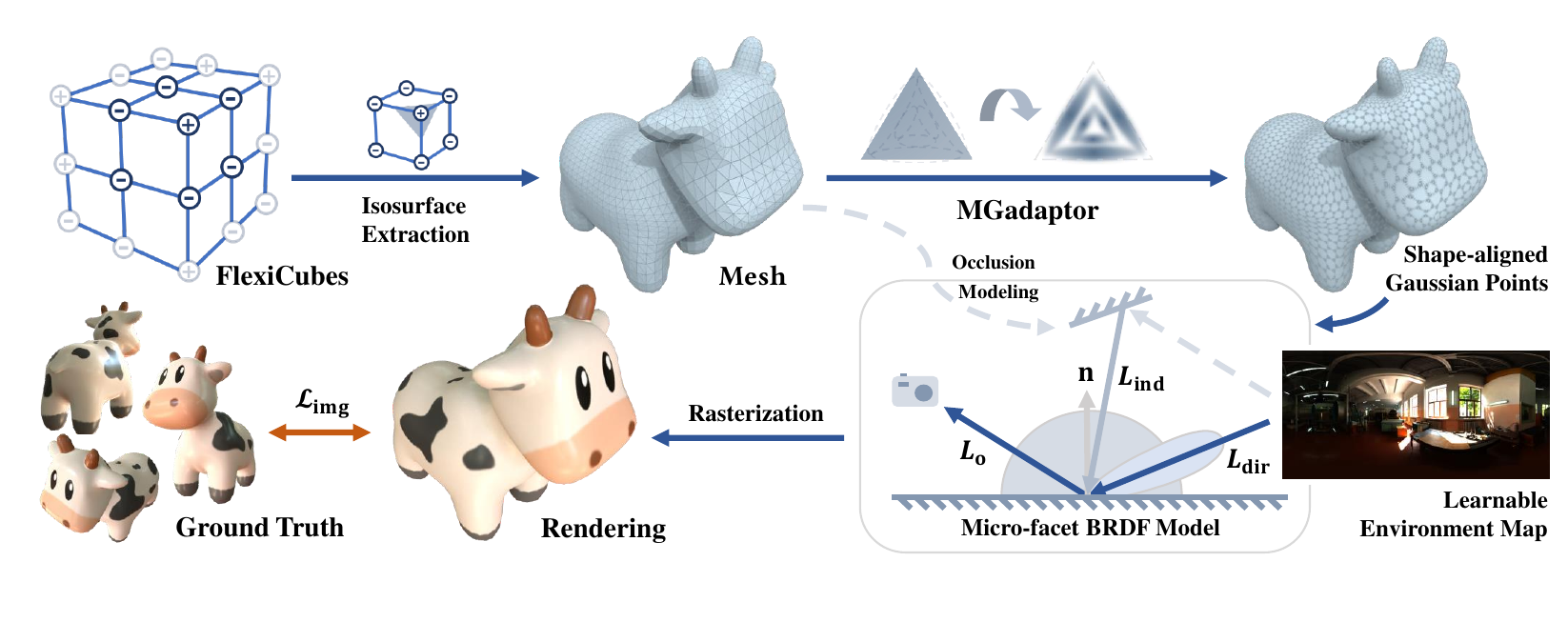}
	\end{center}
	\vspace*{-0.7cm}
	\caption{\label{fig:pipeline} \textbf{Pipeline}. {\name} first extracts an intermediate mesh from the scalar field, upon which Gaussian points are sampled and rendered using PBR equations. Finally, they are composited into images through the Gaussian rasterization pipeline. \textbf{The entire process is fully differentiable and can be trained end-to-end.}}
	\vspace*{-0.4cm}
\end{figure*}

\vspace{-1mm}
\section{Related Work}

\subsection{Geometry Modeling for Inverse Rendering} 

\paragraph{Implicit Field Representations.} 
Neural Radiance Fields (NeRF) \cite{mildenhall2020nerf} pioneered in utilizing implicit neural networks and differentiable volume rendering~\cite{volumerender1988} for high-quality novel view synthesis. Building on this, several works have extended NeRF by modeling implicit surfaces, linking density values in volumetric rendering with Signed Distance Functions (SDF)~\cite{yariv2021volume, oechsle2021unisurf, wang2021neus}. While these representations offer well-defined normal directions through SDF gradients, making them promising candidates for inverse rendering tasks~\cite{jin2023tensoir,yang2023sire,liu2023nero}, their reliance on dense sampling can lead to inefficiencies during both training and inference.

\paragraph{Mesh Representations.} 
Explicit representations~\cite{shen2021dmtet, shen2023flexicubes} extract triangular meshes from SDF or scalar fields, providing well-defined normal directions and making them well-suited for the Physically-Based Rendering (PBR) framework~\cite{Munkberg_2022_CVPR, hasselgren2022nvdiffrecmc}. However, mesh-based differentiable rendering~\cite{DIBR19, chen2021dibrpp, Laine2020diffrast} often produces sparse gradients, primarily concentrated around triangle edges. Consequently, mesh optimization tends to be more challenging than with implicit fields, particularly when handling complex geometries (\eg, thin structures) and high-frequency lighting-material interactions (\eg, specular surfaces).

\vspace{-1mm}
\subsection{3DGS-based Inverse Rendering} 

More recently, 3D Gaussian Splatting (3DGS)~\cite{kerbl20233d} has emerged as a powerful representation for novel view synthesis, offering significantly superior rendering efficiency compared to NeRF and its variants. However, 3DGS models the scene with Gaussian points which often misalign with the ground truth surface and are prone to artifacts like floaters. To address these issues, some research extends the vanilla 3DGS with geometry enhancement techniques, either by introducing additional regularizations to ensure Gaussian points adhere closely to the surface and remain disk-shaped~\cite{guedon2023sugar, Huang2DGS2024, dai2024high, Chen2024PGSRPG}, or by jointly training 3DGS with implicit SDF fields~\cite{lyu20243dgsr, yu2024gsdf3dgsmeetssdf, xiang2024gaussianroomimproving3dgaussian}.

Inspired by these geometry enhancement techniques, several methods are proposed to further extend the vanilla 3DGS for inverse rendering tasks. For example, R3DG~\cite{R3DG2023} learns additional normal attributes and regularizes normal directions using rendered depth maps. GaussianShader~\cite{jiang2023gaussianshader} utilizes the shortest axis direction, while GIR~\cite{shi2023gir} employs eigendecomposition to determine surface orientations. These methods also incorporate the Disney BRDF~\cite{disneybrdf} with split-sum simplification~\cite{splitsum} and consider indirect lighting through ray tracing or residual terms.
However, the approximated normals are less accurate and can be spatially inconsistent, therefore leading to poor decomposition results, especially for reflective surfaces.
In contrast, we propose a unified framework that combines explicit mesh representation with 3DGS, achieving shape consistency between meshes and Gaussian points. This innovation enables explicit modeling of normal directions and enhances Gaussian primitives to better capture light transport, resulting in more effective material and lighting decomposition for inverse rendering tasks.

\vspace{-1mm}
\subsection{Rendering Equation Evaluation}

While most inverse rendering approaches rely on the physically-based rendering equation to model material-lighting disentanglement, directly solving the rendering equation, which involves a complex and computationally expensive integral, remains impractical. Instead, many approaches incorporate approximation techniques to accelerate computations.
Among these, the split-sum approximation~\cite{splitsum} stands out for achieving both high efficiency and excellent render quality and has been utilized in numerous inverse rendering approaches~\cite{Munkberg_2022_CVPR, jiang2023gaussianshader, liang2024gs}. However, its lack of self-occlusion modeling often leads to inaccurate decomposition, especially under strong shadow effects. In contrast, methods that incorporate Spherical Gaussian (SG) with ray tracing can effectively handle shadow effects and inter-reflection~\cite{boss2021nerd, physg2021, zhang2022invrender, jin2023tensoir, R3DG2023}, but struggle with reflective cases due to the low-frequency nature of SG.

To address both the need for self-occlusion awareness and reflective surface modeling, our approach employs Monte Carlo sampling to evaluate the rendering equation in a ray tracing manner, building on prior works~\cite{hasselgren2022nvdiffrecmc, liu2023nero}. Rather than directly conducting ray tracing on 3D Gaussians~\cite{R3DG2023}, our approach utilizes the explicit mesh extracted from our isosurface to serve as an effective proxy, benefiting from the more efficient BVH-accelerated ray tracing.

\vspace{-1mm}
\section{Methodology}
\label{sec:method}

{\name} effectively bridges 3DGS and mesh representations by differentiably constructing a set of structured Gaussian points grounded on the mesh surface, leveraging the accurate geometry and normals of the mesh, along with the efficiency and superior rendering capabilities of 3DGS.
As illustrated in Fig.~\ref{fig:pipeline}, it begins by utilizing FlexiCubes \cite{shen2023flexicubes} to extract a triangular mesh from the learnable isovalue grids. We then introduce a {\meshsampler} that generates 3DGS points on the mesh surface in a differentiable manner (Sec.~\ref{sec:method_mesh2gaussian}).
Next, we present an efficient 3DGS PBR framework, which computes the rendering equation~\cite{renderingequation} to produce per-Gaussian PBR colors (Sec.~\ref{sec:method_render}). A BVH-accelerated ray tracing is performed on the extracted mesh to evaluate light transport (Sec.~\ref{sec:method_raytracing}). Once the Gaussian points are shaded, the 3DGS rasterization algorithm is used to generate the final rendering results.
Additionally, we provided implementation details in Sec.~\ref{sec:method_implement}.

\subsection{Geometry Guided Gaussian Points Generation}
\label{sec:method_mesh2gaussian}

\paragraph{Background.} The vanilla 3DGS~\citep{kerbl3Dgaussians} represents a Gaussian ellipsoid by a full 3D covariance matrix $\threeDCov$ and its center position $\centerPos$: $\threeDGaussian(\point) = e^{-\frac{1}{2} (\point-\centerPos)^T \threeDCov^{-1} (\point-\centerPos)}$, where $\point$ is the location of a 3D point. 
To ensure a valid positive semi-definite covariance matrix,  
$\threeDCov$ is decomposed into the scaling matrix $\threeDScaling$ and the rotation matrix $\threeDRotation$ that characterizes the geometric shape of a 3D Gaussian. 
Beyond $\centerPos$, $\threeDScaling$ and $\threeDRotation$, each Gaussian point maintains additional learnable parameters including % 3D center position $\centerPos \in \mathbb{R}^3$, 
Through periodic densification and culling heuristics, 3DGS adaptively controls the distribution of Gaussian points during optimization. While capable of high-fidelity rendering, 3D Gaussians inherently lack deterministic geometry boundaries and explicit normal directions. This significantly complicates light transport computation, thereby hindering their direct application to physically-based inverse rendering tasks.

\begin{figure}[t]
	\vspace{-3mm}
	\begin{center}
	\setlength{\tabcolsep}{1pt}
	\setlength{\fboxrule}{1pt}
	\begin{tabular}{cc}
		\begin{tabular}{c}
			\vspace{-0.5mm}
			\includegraphics[width=0.48\linewidth]{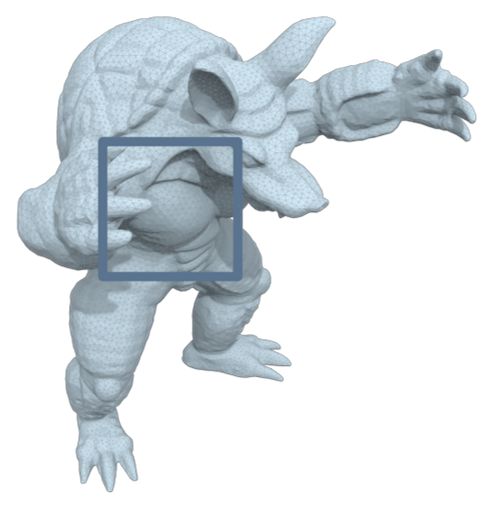}\\
			{\footnotesize{(a) Underlying Mesh}}
		\end{tabular}
		&
		\begin{tabular}{cc}
			\includegraphics[width=0.22\linewidth]{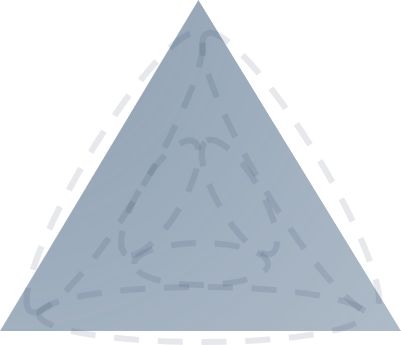} &
			\includegraphics[width=0.22\linewidth]{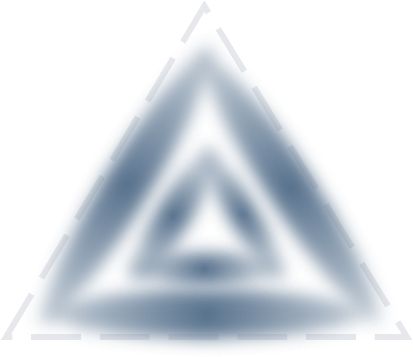}\\
			\vspace{1mm}
			{\scriptsize{(b) Triangle Facet}} & {\scriptsize{(c) Gaussian Pattern}}\\
			\includegraphics[width=0.22\linewidth]{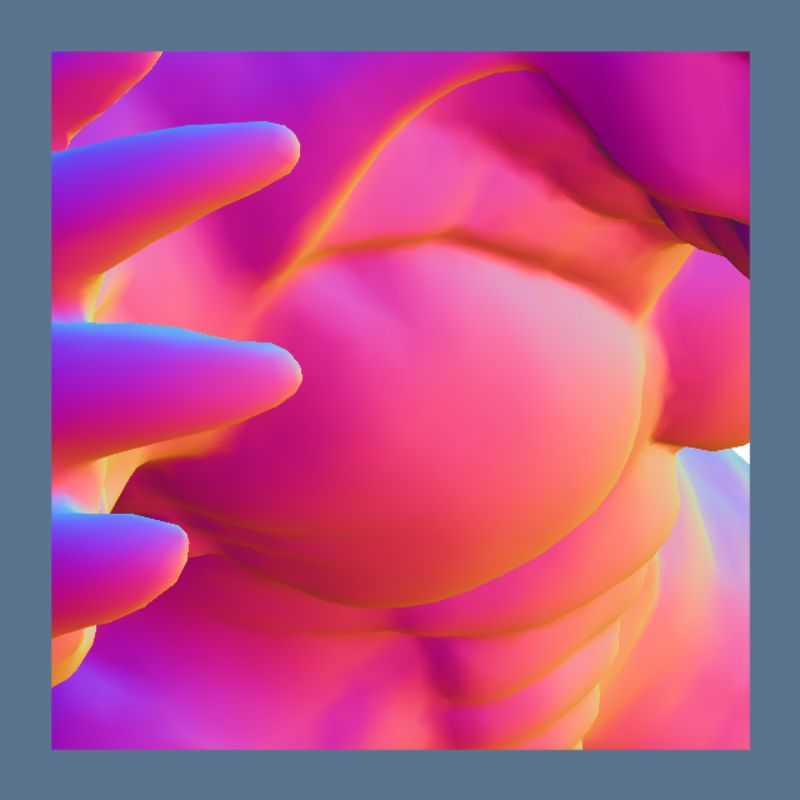} &
			\includegraphics[width=0.22\linewidth]{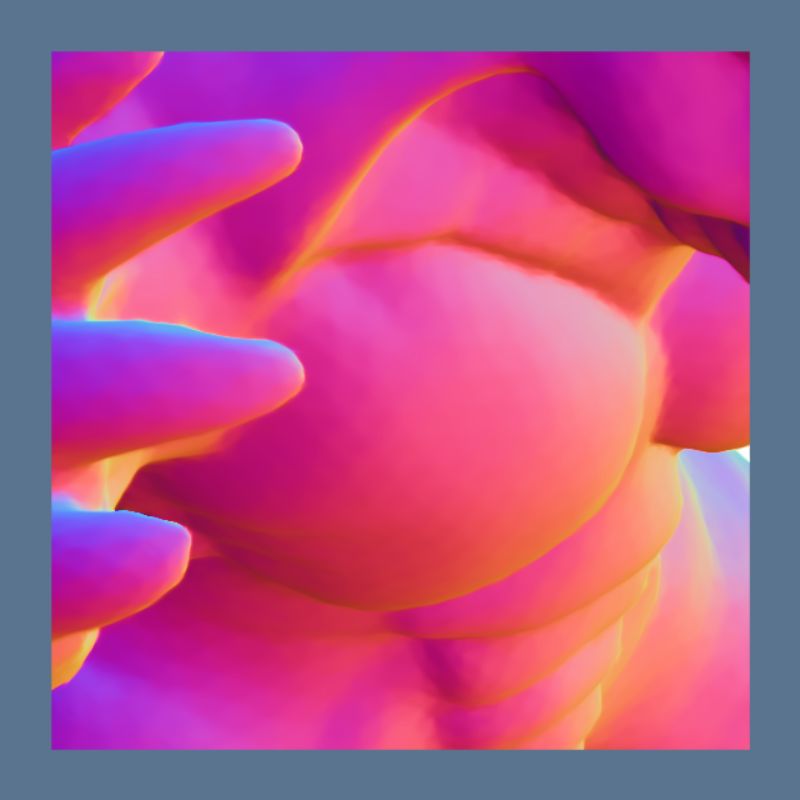} \\
			{\scriptsize{(d) Mesh Normals}} & {\scriptsize{(e) Gaussian Normals}}
		\end{tabular}
	\end{tabular}
  	\end{center}
	\vspace{-6mm}
	\caption{ \label{fig:sample}
		\textbf{{\meshsampler} Overview.}
		(a) Given the underlying mesh, (b) {\meshsampler} is applied to each triangle facet, (c) following a predefined pattern to draw six Gaussian points. The pattern is predefined in barycentric coordinate space to maintain consistency between (d) the mesh normals and (e) the Gaussian normals without additional optimization, inherently enabling accurate modeling of Gaussian normals.
	}
	\vspace{-3mm}
\end{figure}

\vspace{-3mm}
\paragraph{Method.}
Our goal is to introduce explicit geometric guidance to 3D Gaussian Splatting. 
To achieve this, we leverage the isosurface technique FlexiCubes \citep{shen2023flexicubes} for geometry guidance and propose {\meshsampler}, which generates Gaussian points guided by the mesh geometry.
Starting with a scalar field $\sdfFunc: \mathbb{R}^3 \rightarrow \mathbb{R}$ which maps spatial locations to scalar values, isosurfaces can be represented by discretizing $\sdfFunc$ as learnable values stored at grid vertices.
We then apply the differentiable isosurface extraction technique, \ie, FlexiCubes, to obtain a triangular mesh $\mesh$ from $\sdfFunc$. 
With this intermediate mesh $\mesh$ serving as explicit guidance, we propose {\meshsampler}, formulated as the function $\meshSamplerFunc$, to generate Gaussian points attributes $\left\{\left(\centerPos, \threeDScaling, \threeDRotation, \normal \right)\right\}$ from each triangles $\mathbf{P}$. Specifically, the {\meshsampler} produces $K$ Gaussian points for each triangle $\mathbf{P}$:
\vspace{-1mm}
\begin{equation}
	\left\{\left(\centerPos_i, \threeDScaling_i, \threeDRotation_i, \normal_i \right)|~i=1,\dots,K\right\} = \meshSamplerFunc(\mathbf{P}) \eqstop
 \label{eq:meshsampler}
\end{equation} 
$\meshSamplerFunc$ computes the shape attributes of each generated Gaussian point, including its position $\centerPos_i \in \mathbb{R}^{3}$, scale $\threeDScaling_i \in \mathbb{R}^{3}$, and rotation $\threeDRotation_i \in \mathbb{R}^{3\times3}$, along with the normal $\normal_i \in \mathbb{R}^{3}$. 
Note that the shape parameters of the Gaussian points are fully determined by each triangle $\mathbf{P}$. We empirically set $K=6$ and divide the triangle into 6 sub-regions, placing one Gaussian point in each region (Fig.~\ref{fig:sample}). The position $\centerPos_i$ and normal $\normal_i$ are computed via barycentric interpolation, while the scale $\threeDScaling_i$ and rotation $\threeDRotation_i$ are decided by the orientation and shape of the triangle $\mathbf{P}$. The opacity is set to 1, as the mesh surface is opaque. This {\meshsampler} sampling process is fully differentiable, allowing gradients to be effectively propagated to the isosurface $\sdfFunc$. We illustrate the {\meshsampler} in Fig.~\ref{fig:sample} and provide the details in Appendix~\ref{app:meshsampler}.

\vspace{-3mm}
\paragraph{Discussion.} Rather than learning approximated normals and supervising with normal-depth consistency loss~\cite{R3DG2023,jiang2023gaussianshader}, our surface-grounded Gaussian points leverage the well-defined mesh normals, resulting in improved normal accuracy. This leads to significantly better inverse rendering performance compared to previous 3DGS-based methods that rely on approximated normals. Furthermore, the {\meshsampler} is specifically designed to produce highly structured Gaussian points that maintain shape consistency relative to the triangular mesh. This shape alignment naturally enables the intermediate mesh $\mesh$ to serve as a proxy for self-occlusion evaluation, facilitating effective light transport modeling for occlusion-aware decomposition, which we explain later in Sec.~\ref{sec:method_raytracing}.

\subsection{Physically-based Gaussian Rendering}
\label{sec:method_render}

\paragraph{Background.}
The vanilla 3DGS assigns each Gaussian point a $k$-order spherical harmonic parameter to represent view-dependent rendering effects, which lacks physical interpretability. In contrast, our {\name} represents high-order lighting effects through the physically-based rendering equation~\citep{renderingequation} (Eq.~\ref{eq:re}), where the BRDF material is formulated by GGX microfacet model~\citep{GGX} (Eq.~\ref{eq:ggx}).
\begin{equation} 
\Lo(\x,\wo)= \int_\hemis \brdf(\x,\wi,\wo)\Li(\x,\wi)|\normal \cdot \wi|\dd \wi\eqcomma \label{eq:re}
\end{equation}
\vspace{-3mm}
\begin{equation} 
\brdf(\x,\wi,\wo)= (1-\materialMetalness)\frac{\diffAlbedo}{\pi} + \frac{DFG}{4|\normal \cdot \wi||\normal \cdot \wo|} \eqstop\label{eq:ggx}
\end{equation}
In Eq.~\ref{eq:re}, the outgoing radiance $\Lo(\x,\wo)$ in the direction $\wo$ is computed as the integral of the BRDF function $\brdf(\x,\wi,\wo)$, the incoming light $\Li(\x,\wi)$, and the cosine term $|\normal \cdot \wi|$, which accounts for the angle between the surface normal $\normal$ and the incoming light direction $\wi$, over the hemisphere $\hemis$.
In Eq.~\ref{eq:ggx}, the GGX model defines the BRDF function $\brdf(\x,\wi,\wo)$ as two components: the diffuse term $(1-\materialMetalness)\frac{\diffAlbedo}{\pi}$ and the specular term $\frac{DFG}{4|\normal \cdot \wi||\normal \cdot \wo|}$, which are all determined by PBR material attributes, including albedo $\diffAlbedo \in [0,1]^3$, metalness $\materialMetalness\in[0,1]$ and roughness $\materialRoughness\in[0,1]$.

\vspace{-3mm}
\paragraph{Method.}
Our goal is to assign PBR attributes to Gaussian points and then evaluate the rendering equation to produce PBR colors. Following prior works~\cite{Munkberg_2022_CVPR,hasselgren2022nvdiffrecmc}, we introduce multi-resolution hash grids~\citep{muller2022instant} $\mathcal{E}_\mathrm{d}$ and $\mathcal{E}_\mathrm{s}$ for PBR attribute querying. Specifically, given a Gaussian point with position $\centerPos_i$, its PBR attributes are generated as follows:
\vspace{-1mm}
\begin{equation}
	\diffAlbedo_i = \mathcal{E}_\mathrm{d}(\centerPos_i)\eqcomma\ \ %
	(\materialRoughness_i, \materialMetalness_i) = \mathcal{E}_\mathrm{s}(\centerPos_i)\eqstop
	\label{eq:mlptexture}
\end{equation}

\vspace{-5mm}
\paragraph{Discussion.}
Once PBR attributes are obtained, the BRDF function $\brdf$ is determined. We then differentiably solve Eq.~\ref{eq:re} with Monte Carlo sampling to obtain PBR colors for each Gaussian point, which are subsequently processed using standard 3DGS rasterization and rendered into final images via alpha compositing. Subsequently, a backpropagation process is applied to progressively optimize the PBR attributes through photometric supervision.

\subsection{Light Transport Modeling}
\label{sec:method_raytracing}

\paragraph{Background.}
During the evaluation of Eq.~\ref{eq:re}, accurately modeling the incident lighting $\Li(\x,\wi)$ remains challenging, as it comprises not only the environment illumination but also the indirect illumination from multi-bounce inter-reflection. Effective light transport modeling is required to capture higher-order effects, such as shadowing and inter-reflections. Instead of relying on computationally expensive path tracing techniques for multi-bounce light transport modeling~\cite{L1995MMAMCAFPBR,MILO}, existing approaches often approximate the lighting with a single-bounce model, disentangling $\Li(\x,\wi)$ into two components: the direct term $\Ldir(\wi)$, representing the environmental illumination, and the indirect term $\Lind(\x,\wi)$, accounting for inter-reflection. A scalar factor $O(\x,\wi)$ is introduced to model self-occlusion, weighting the ratio between these two lighting components:
\begin{equation}
\vspace{-1mm}
\begin{aligned}
	\Li(\x,\wi) =~& (1-O(\x,\wi))\Ldir(\wi)\\ & +O(\x,\wi)\Lind(\x,\wi) \eqstop
\end{aligned}
\label{eq:light_decomposition}
\end{equation}
Here, the direct term $\Ldir(\wi)$ is usually represented with a $H\times W\times3$ environment map which can be queried by latitude and longitude coordinates, while the indirect term is modeled as a set of Spherical Harmonic coefficients.

\begin{figure}[t]
\vspace{-5mm}
\begin{center}
\resizebox{\linewidth}{!}{
\setlength{\tabcolsep}{1pt}
\setlength{\fboxrule}{1pt}
\hspace{-2mm}
\begin{tabular}{c}
	\begin{tabular}{cc}
		\vspace{-1mm}
		\includegraphics[width=0.32\linewidth,trim={0.5cm 3cm 0.5cm 5cm},clip]{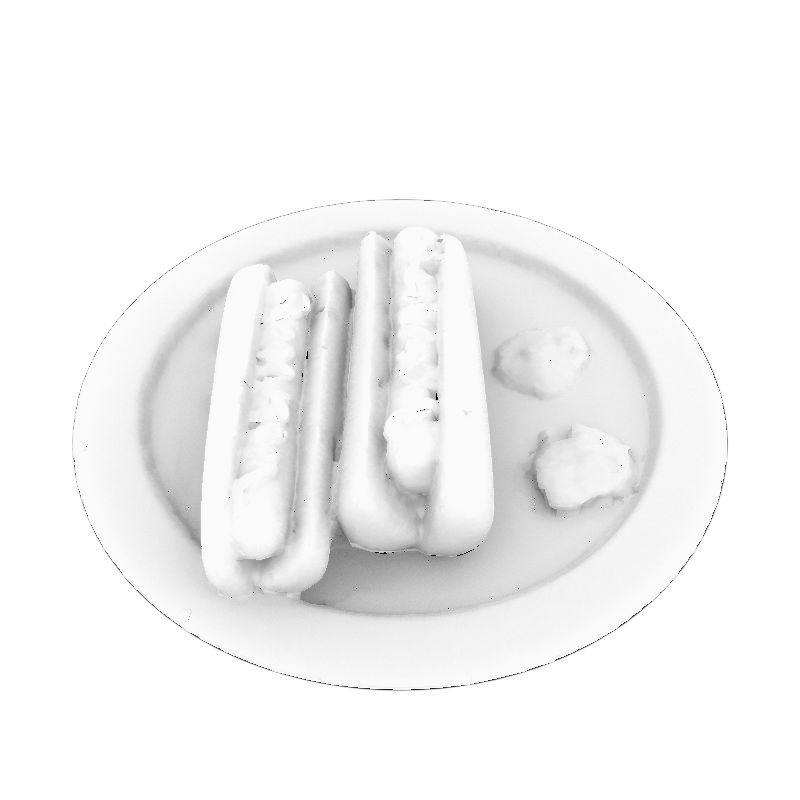} &
		\includegraphics[width=0.32\linewidth,trim={0.5cm 3cm 0.5cm 5cm},clip]{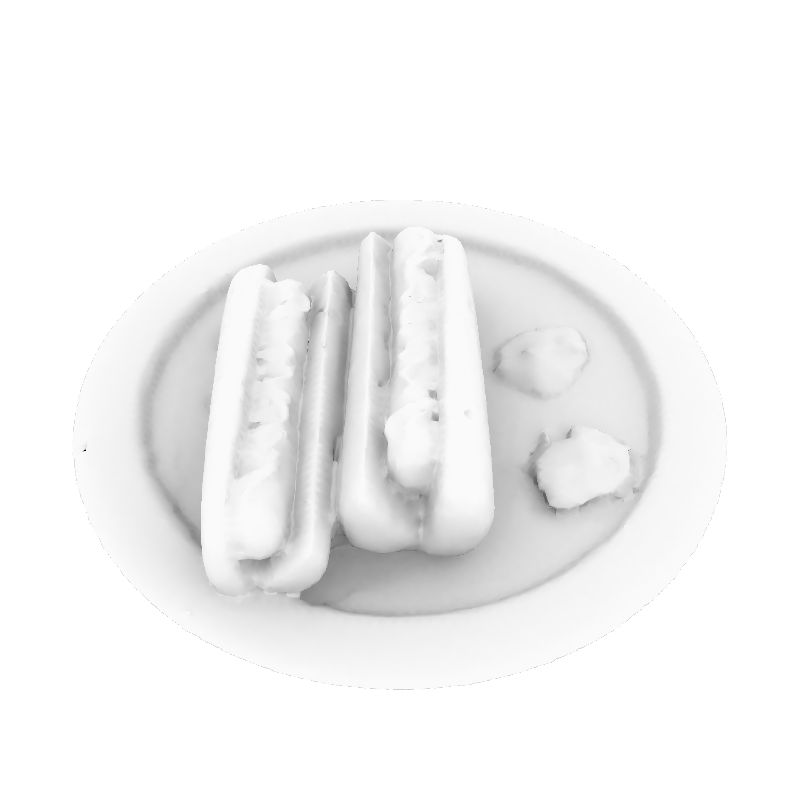} \\
		\footnotesize{(a) $\Ogs$} &
		\footnotesize{(b) $\Omesh$}
	\end{tabular}
	\\
	\begin{tabular}{ccc}
		\vspace{-1mm}
		\includegraphics[width=0.32\linewidth,trim={0.5cm 3cm 0.5cm 5cm},clip]{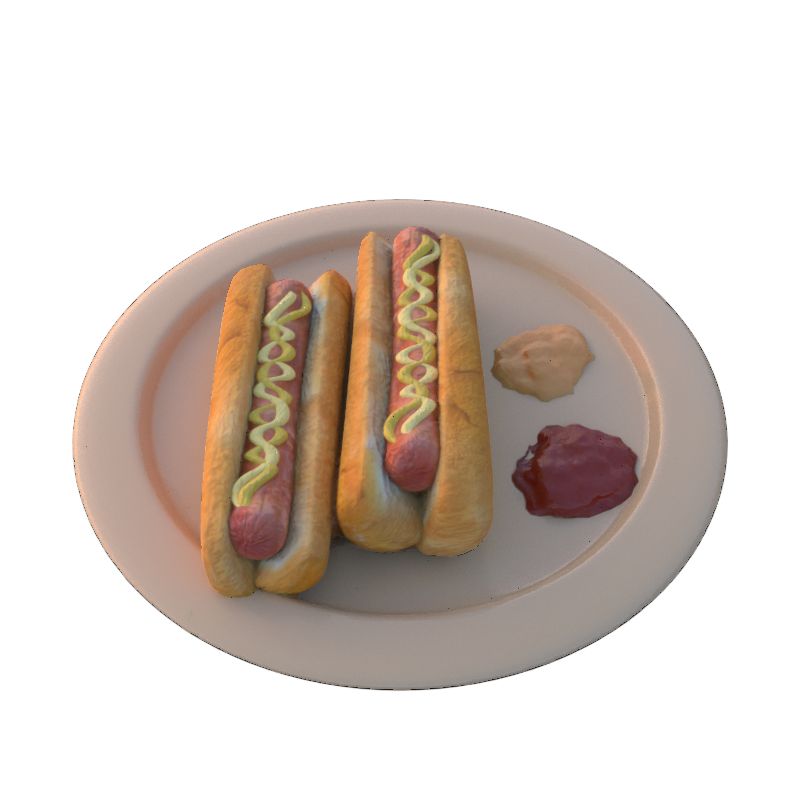} &
		\includegraphics[width=0.32\linewidth,trim={0.5cm 3cm 0.5cm 5cm},clip]{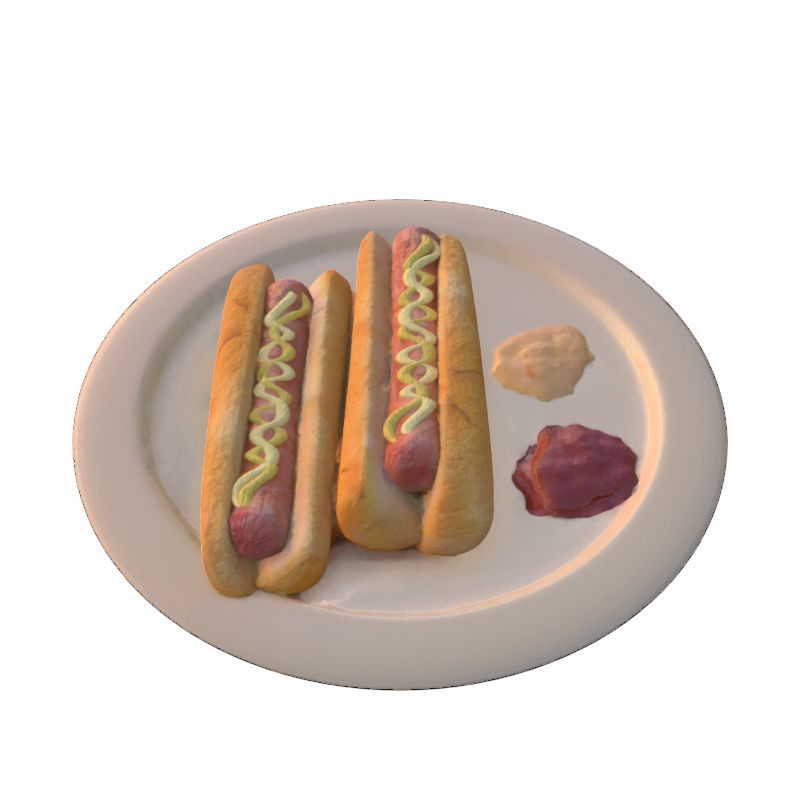} &
		\includegraphics[width=0.32\linewidth,trim={0.5cm 3cm 0.5cm 5cm},clip]{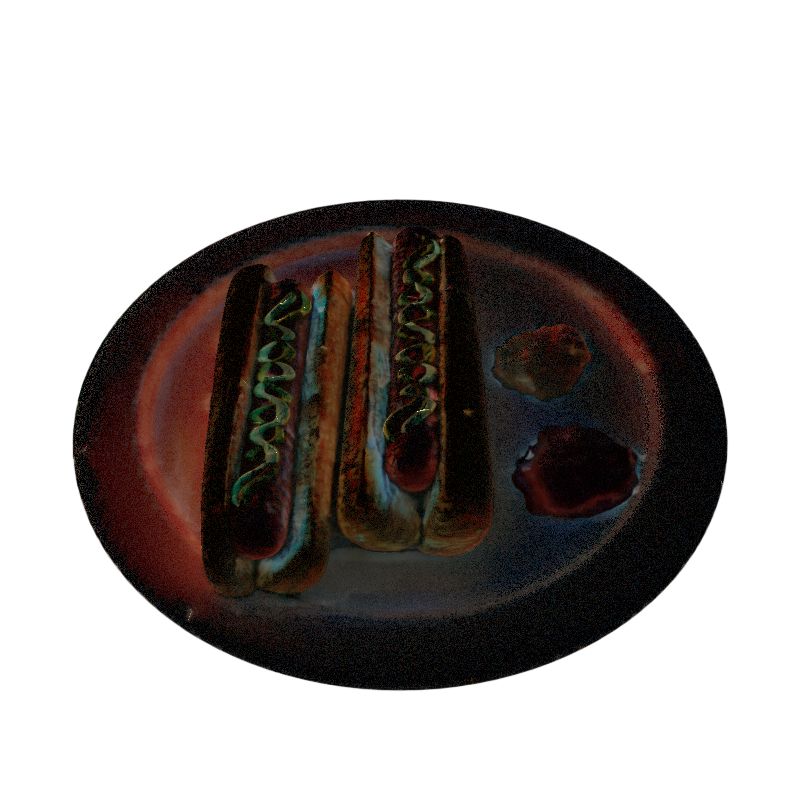}\\
		\footnotesize{(c) Render w/ full $\Li$} &
		\footnotesize{(d) Render w/ only $\Ldir$} &
		\footnotesize{(e) Render w/ only $\Lind$}
	\end{tabular}
\end{tabular}
}
\end{center}
\vspace{-5mm}
\caption{ \label{fig:light_transport}
	\textbf{{Self-Occlusion Evaluation}}.
	We visualize the occlusion term evaluated on (a) generated Gaussian points ($\Ogs$) and (b) the underlying mesh ($\Omesh$), respectively. The Gaussian-mesh shape alignment ensures consistency between $\Ogs$ and $\Omesh$, enabling the mesh to serve as an effective proxy for accelerated self-occlusion evaluation. We additionally illustrate (c) full PBR, (d) PBR without self-occlusion, and (e) Inter-reflection, to demonstrate the effectiveness of our self-occlusion modeling.
}
\vspace{-2mm}
\end{figure}

\vspace{-3mm}
\paragraph{Method.}
Our goal is to achieve effective light transport modeling. The key to this lies in estimating the occlusion term $O(\x,\wi)$ both accurately and efficiently. Prior 3DGS-based inverse rendering approaches~\cite{R3DG2023,moenne20243d} typically model occlusion term as a contiguous form $\Ogs(\x,\wi)\in[0,1]$ and estimate it by accumulating Gaussian opacities along the ray $\mathbf{r}(t)=\x+t\wi$, which is computationally expensive and leads to low efficiency. In contrast, by leveraging explicit geometry guidance, we replace $\Ogs(\x,\wi)\in[0,1]$ with a binary term $\Omesh(\x,\wi)\in\{0,1\}$, which is estimated on the deterministic mesh surface. This approach enables mesh-based ray tracing, accelerated by BVH, for efficient occlusion evaluation. As illustrated in Fig~\ref{fig:light_transport}, this replacement does not introduce noticeable errors because our {\meshsampler} ensures shape alignment between the 3DGS and the mesh, thereby maintaining consistency between the approximated $\Omesh(\x,\wi)$ and the ground truth $\Ogs(\x,\wi)$. A more detailed analysis of the error bound can be found in Appendix~\ref{app:meshsampler:analysis}.

\vspace{-3mm}
\paragraph{Discussion.}
With explicit geometry guidance, we propose ray tracing techniques tailored to our hybrid representations for self-occlusion evaluation. Compared to existing methods which also rely on ray tracing for occlusion-aware light transport modeling~\cite{R3DG2023,jin2023tensoir,liu2023nero}, our solution significantly reduces optimization time, delivering exceptional efficiency.

\subsection{Implementation Details}
\label{sec:method_implement}

\paragraph{Loss Functions.} 
The pipeline of {\name} is fully differentiable and can be trained end-to-end with photometric supervision.
Thanks to the geometry guidance, we do not require regularization terms, such as dist loss~\citep{barron2022mip} or pseudo depth normal loss~\citep{jiang2023gaussianshader,R3DG2023}, to provide additional geometric constraints.
However, similar to NVdiffrecmc, {\name} also relies on an object mask loss, as optimizing explicit surface is more challenging. 
Specifically, the photometric loss is computed as:
$\loss_\mathrm{img} = \loss_{1} + \lambda_\mathrm{ssim}\loss_\mathrm{SSIM} + \lambda_\mathrm{mask}\loss_\mathrm{mask}$,
where $\loss_\mathrm{mask}$ is an MSE loss between the rendered alpha map and the ground truth object mask.
Furthermore, we add an entropy loss to constrain the shape, following DMTet and FlexiCubes~\citep{shen2021dmtet,shen2023flexicubes}.
To achieve better decomposition, we apply smoothness regularization on $\materialDiffuse$ and $\materialSpecular$ along with a light regularization, following NVdiffrecmc and R3DG~\citep{hasselgren2022nvdiffrecmc,R3DG2023}.
The final loss $\loss$ is a combination of the photometric loss and the regularization losses:
$\loss = \loss_\mathrm{img} + \lambda_\mathrm{entropy}\loss_\mathrm{entropy} + \lambda_\mathrm{smooth}\loss_\mathrm{smooth} + \lambda_\mathrm{light}\loss_\mathrm{light}$.
More details are provided in Appendix~\ref{app:loss}.

\begin{comment}
\paragraph{Split-sum Approximation}
Direct evaluation of the integral in Eq. \ref{eq:re} is computationally expensive and may lead to optimization instability, especially during the initial stages of training when the geometric shape is still noisy. To mitigate this issue, we follow prior works \cite{liu2023nero} to adopt the split-sum approximation technique during the first half of the training process. Specifically, the split-sum technique approximates Eq. \ref{eq:re} as follows:
\begin{equation}
\Lo(\wo) \approx \int_\hemis \brdf(\wi,\wo)|\normal\cdot \wi|\,\dd \wi
\int_\hemis \Li(\wi) D |\normal\cdot \wi|\,\dd \wi \eqstop \label{eq:chap4:splitsum}
\end{equation}
Eq.~\ref{eq:chap4:splitsum} enables fast pre-computation. The left BRDF integral can be stored in a 2D lookup table and queried using $|\normal \cdot \wi|$ and $\materialRoughness$, while the right lighting integral is represented by pre-integrated environment maps and can also be queried by $\materialRoughness$. 
This approach enables the direct computation of outgoing radiance \(\Lo\) using material parameters without requiring any ray sampling, significantly enhancing evaluation speed. Additional details are provided in Appendix. In the latter half of the training stage, as the geometric shape stabilizes, we transition to our Monte Carlo sampling process for a more accurate evaluation of Eq. \ref{eq:re}, incorporating shadow effects and inter-reflections.
\end{comment}

\vspace{-3mm}
\paragraph{Split-Sum Warm-up}
Directly evaluating the integral in Eq.~\ref{eq:re} is inefficient and may lead to optimization instability, especially during the initial stages of training when the geometric shape used to model light transport is still noisy. To mitigate this issue, we follow prior works~\cite{liu2023nero} to adopt the split-sum approximation~\cite{splitsum} technique during the early training stage.
The split-sum technique assumes no self-occlusion, approximating Eq.~\ref{eq:re} as follows:
\vspace{-1mm}
\begin{equation}
\begin{aligned}
	\Lo(\wo) \approx& \int_\hemis \brdf(\wi,\wo)|\normal\cdot \wi|\,\dd \wi\\
	&\cdot\int_\hemis \Ldir(\wi) D |\normal\cdot \wi|\,\dd \wi \eqstop \label{eq:chap4:splitsum}
\end{aligned}
\end{equation}
While Eq.~\ref{eq:chap4:splitsum} enables fast pre-computation, significantly enhancing evaluation speed, its lack of self-occlusion modeling can lead to suboptimal decomposition results (\eg, baking shadow into albedo). Therefore, after this warm-up stage, we transition to our Monte Carlo sampling process for a more accurate evaluation of Eq.~\ref{eq:re}, incorporating shadow effects and inter-reflection.

\begin{figure}[t]
	\vspace{-1mm}
	\begin{center}
	\setlength{\tabcolsep}{1pt}
	\setlength{\fboxrule}{1pt}
	\begin{tabular}{c}
		\begin{tabular}{cc}
			\vspace{-1mm}
			\includegraphics[width=0.45\linewidth,trim={2cm 0cm 0cm 3cm},clip]{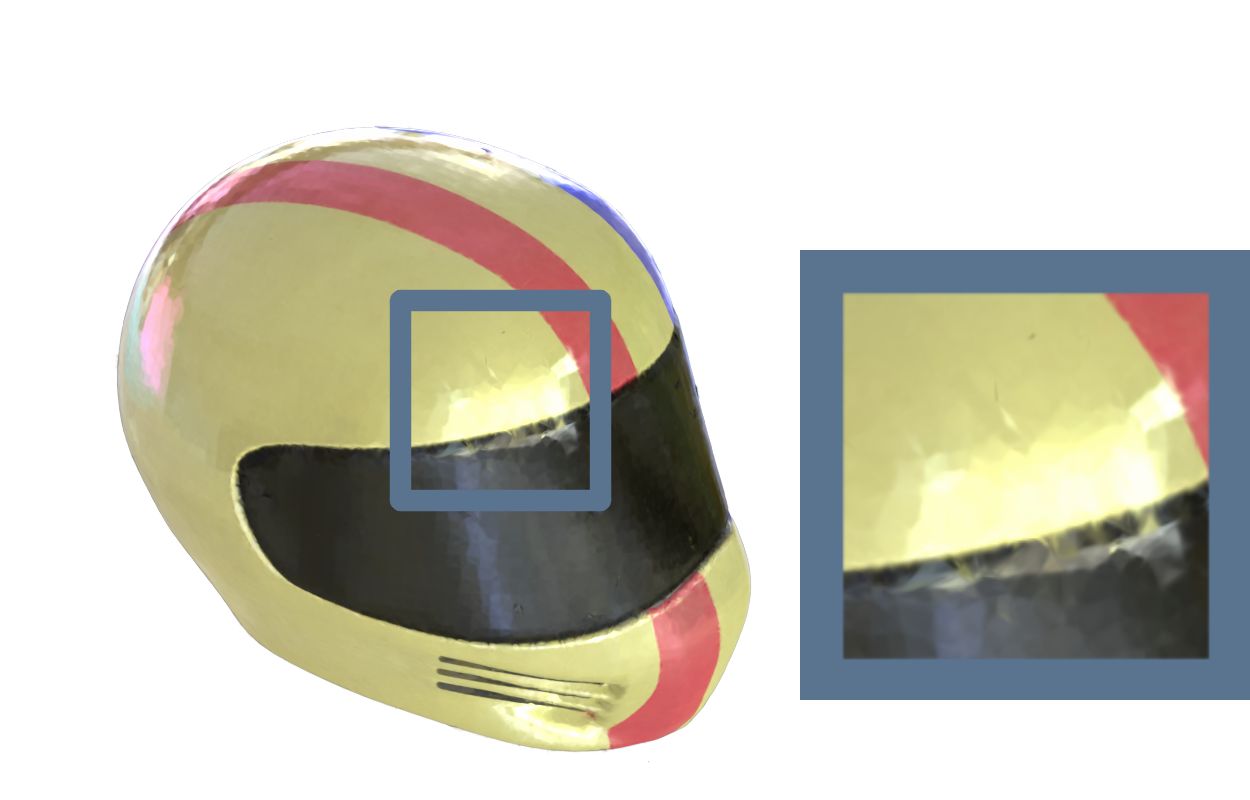} &
			~\includegraphics[width=0.45\linewidth,trim={2cm 0cm 0cm 3cm},clip]{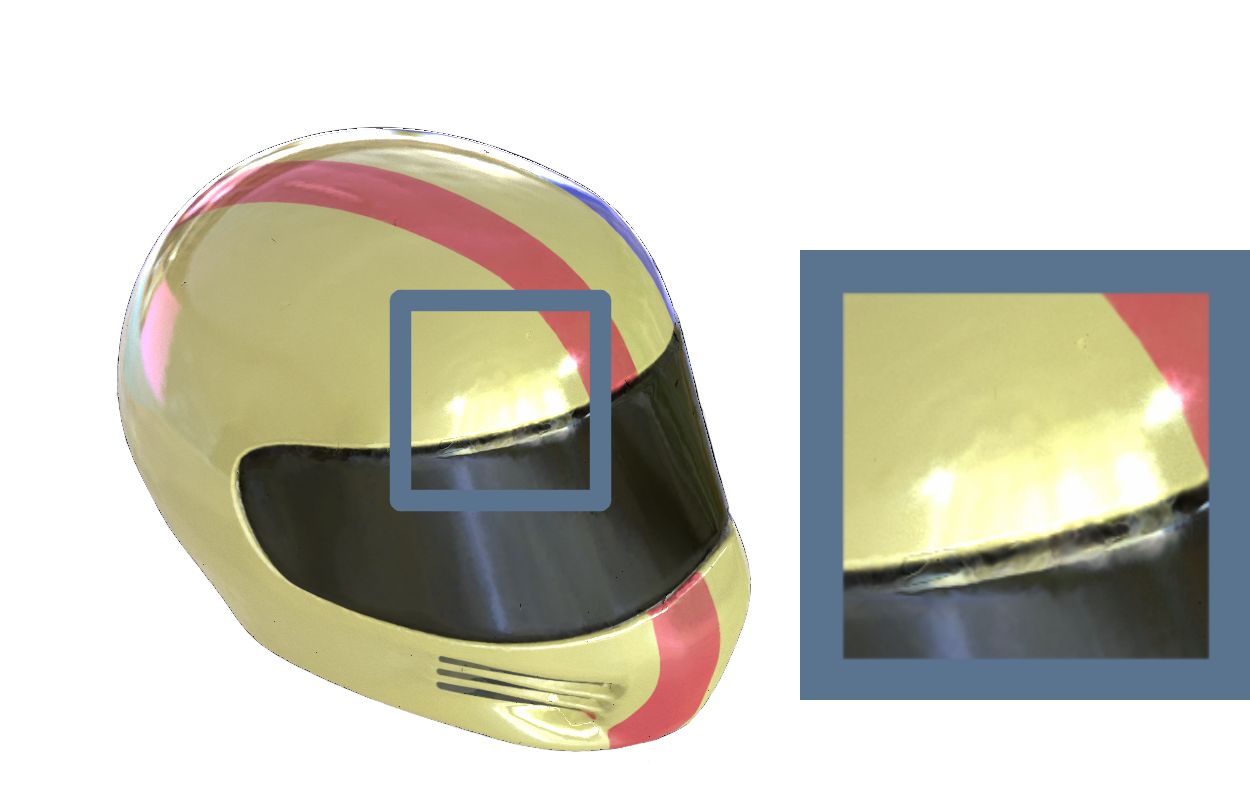} \\
			\footnotesize{(a) Forward Shading} &
			\footnotesize{(b) Deferred Shading}
		\end{tabular}
	\end{tabular}
	\end{center}
	\vspace{-5mm}
	\caption{ \label{fig:shading}
		\textbf{{Appearance Refinement.}}
		(a) The rendering details of forward shading are limited by the spatial density of Gaussian points. (b) In contrast, deferred shading conducts pixel-wise PBR, delivering much more detailed results for reflective objects.
	}
	\vspace{-3mm}
\end{figure}

\vspace{-3mm}
\paragraph{Appearance Refinement.}
{\name} conducts the per-Gaussian shading, \ie, first computes PBR colors on Gaussian points and then rasterizes the per-Gaussian colors into pixels. Since each Gaussian point only delivers a single PBR color, the rendering detail of a surface region is constrained by its Gaussian point density, making it difficult to accurately model high-frequency variations in appearance (\eg, strong reflections). This issue is particularly pronounced for our highly structured Gaussian points. To address it, we consider the deferred shading techniques which conduct pixel-wise PBR for appearance refinement. Similar to existing works~\cite{ye2024gsdr,liang2024gus}, we allow {\name} to switch to deferred shading and slightly adjust displacement of Gaussian points at the end of the training, as illustrated in Fig.~\ref{fig:shading}. An ablation study on this refinement can be found in Sec.~\ref{exp:abl} and further details are provided in Appendix~\ref{app:appref}.

\begin{table*}
\vspace{-3mm}
\begin{center}
\resizebox{\textwidth}{!}{ % Resizes the table to fit within the text width
\begin{tabular}{cc|ccccccc|cccccc|ccc}
\toprule
\multicolumn{2}{c|}{Dataset} & \multicolumn{7}{c|}{Synthetic4Relight} & \multicolumn{6}{c|}{TensoIR Synthetic} & \multicolumn{3}{c}{Shiny Blender} \\

Method\rule{0pt}{3ex} & Training & \multicolumn{3}{c}{Relighting} & \multicolumn{3}{c}{Albedo} & \multicolumn{1}{c|}{Roughness} & \multicolumn{3}{c}{Relighting} & \multicolumn{3}{c|}{Albedo} & \multicolumn{3}{c}{NVS}\\
    & minutes$\downarrow$& PSNR $\uparrow$ & SSIM $\uparrow$ & LPIPS $\downarrow$ & PSNR $\uparrow$ & SSIM $\uparrow$ & LPIPS $\downarrow$ & \multicolumn{1}{c|}{MSE $\downarrow$} & PSNR $\uparrow$ & SSIM $\uparrow$ & LPIPS $\downarrow$ & PSNR $\uparrow$ & SSIM $\uparrow$ & \multicolumn{1}{c|}{LPIPS $\downarrow$} & PSNR $\uparrow$ & SSIM $\uparrow$ & LPIPS $\downarrow$ \\
\midrule
NVdiffrec & 72 & 28.89 & 0.953 & 0.061 & 28.66 & 0.941 & 0.066\cellcolor{best3} & 0.026 & 24.64 & 0.916 & 0.078 & 25.84 & 0.922 & 0.096 & 28.70 & 0.945 & 0.119\\
NVdiffrecmc & 82 & 30.23\cellcolor{best3} & 0.946 & 0.080 & 29.14\cellcolor{best3} & 0.945\cellcolor{best3} & 0.067 & 0.010\cellcolor{best3} & 26.51 & 0.923 & 0.087 & 27.71 & 0.931 & 0.083 & 28.03 & 0.932 & 0.125\\
TensoIR & $\sim$270 & 29.94& 0.948 & 0.082 & 29.60\cellcolor{best2} & 0.934 & 0.073 & 0.015 & 28.51\cellcolor{best3} & 0.945\cellcolor{best1} & 0.059\cellcolor{best2} & 28.35\cellcolor{best3} & 0.953\cellcolor{best1} & 0.059\cellcolor{best1} & 27.89 & 0.907 & 0.156\\
NeRO & $\sim$800 & 30.15 & 0.961\cellcolor{best3} & 0.052\cellcolor{best3} & 26.69 & 0.936 & 0.067 & 0.006\cellcolor{best2} & 26.31 & 0.933\cellcolor{best3} & 0.071\cellcolor{best3} & 25.98 & 0.908 & 0.100 & 29.84\cellcolor{best3} & 0.962\cellcolor{best1} & 0.072\cellcolor{best1}\\
GS-IR & 20\cellcolor{best2} & 23.81 & 0.902 & 0.086 & 26.66 & 0.936 & 0.085 & 0.825 & 24.35 & 0.884 & 0.091 & 26.80 & 0.892 & 0.116 & 27.01 & 0.892 & 0.133\\
GS-Shader & 63\cellcolor{best3} & 22.32 & 0.924 & 0.084 & N/A & N/A & N/A & 0.050 & 22.42 & 0.872 & 0.103  & N/A & N/A & N/A & 30.64\cellcolor{best2} & 0.952\cellcolor{best3} & 0.081\cellcolor{best3}\\
R3DG & $\sim$110 & 31.00\cellcolor{best2} & 0.964\cellcolor{best2} & 0.050\cellcolor{best2} & 28.31 & 0.951\cellcolor{best1} & 0.058\cellcolor{best1} & 0.013 & 28.55\cellcolor{best2} & 0.927 & 0.072 & 28.74\cellcolor{best2} & 0.945\cellcolor{best2} & 0.072\cellcolor{best3} & 28.83 & 0.942 & 0.098\\
Ours & 14\cellcolor{best1} & 34.10\cellcolor{best1} & 0.971\cellcolor{best1} & 0.037\cellcolor{best1} & 29.90\cellcolor{best1} & 0.949\cellcolor{best2} & 0.062\cellcolor{best2} & 0.004\cellcolor{best1} & 29.95\cellcolor{best1} & 0.943\cellcolor{best2} & 0.052\cellcolor{best1} & 29.41\cellcolor{best1} & 0.941\cellcolor{best3} & 0.067\cellcolor{best2} & 31.14\cellcolor{best1} & 0.955\cellcolor{best2} & 0.080\cellcolor{best2}\\
\bottomrule
\end{tabular}
}
\end{center}
\vspace{-5mm}
\caption{
    \label{table:exp:relight} 
\textbf{Quantitative Results of Inverse Rendering.}
Our method achieves state-of-the-art performance in relighting quality and material recovery for general cases from the Synthetic4Relight and TensoIR Synthetic datasets, along with NVS for reflective cases from the Shiny Blender dataset. Note that {\gaussianshader} does not provide disentangled albedo but rather a diffuse color merged with lighting, so we leave it as N/A. Also, note that we apply the albedo scaling introduced in~\citep{zhang2021nerfactor} to perform a fair comparison.
    }
\vspace{-4mm}
\end{table*}
\begin{figure*}
\vspace{-6mm}
    \hspace*{-0mm}
    \begin{center}
    \resizebox{\textwidth}{!}{
        \setlength{\tabcolsep}{1pt}
        \setlength{\fboxrule}{1pt}
        \begin{tabular}{c}
            
            \begin{tabular}{ccccccccccccc}
                & \footnotesize{Render} & \footnotesize{Albedo} & \footnotesize{Roughness} & \footnotesize{Envmap}
                & \footnotesize{Render} & \footnotesize{Albedo} & \footnotesize{Roughness} & \footnotesize{Envmap} \\
                \rotatebox{90}{\footnotesize{{NVdiffrecmc}}} &
                \multicolumn{1}{c}{\includegraphics[width=0.12\linewidth,trim={0cm 8cm 2cm 4cm},clip]{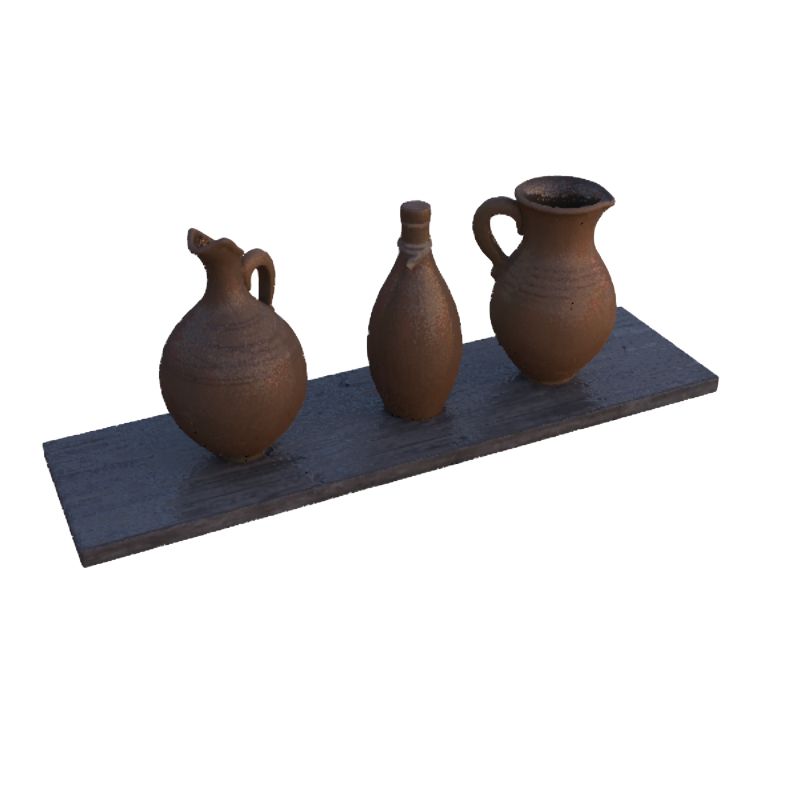}}  &  
                \multicolumn{1}{c}{\includegraphics[width=0.12\linewidth,trim={0cm 8cm 2cm 4cm},clip]{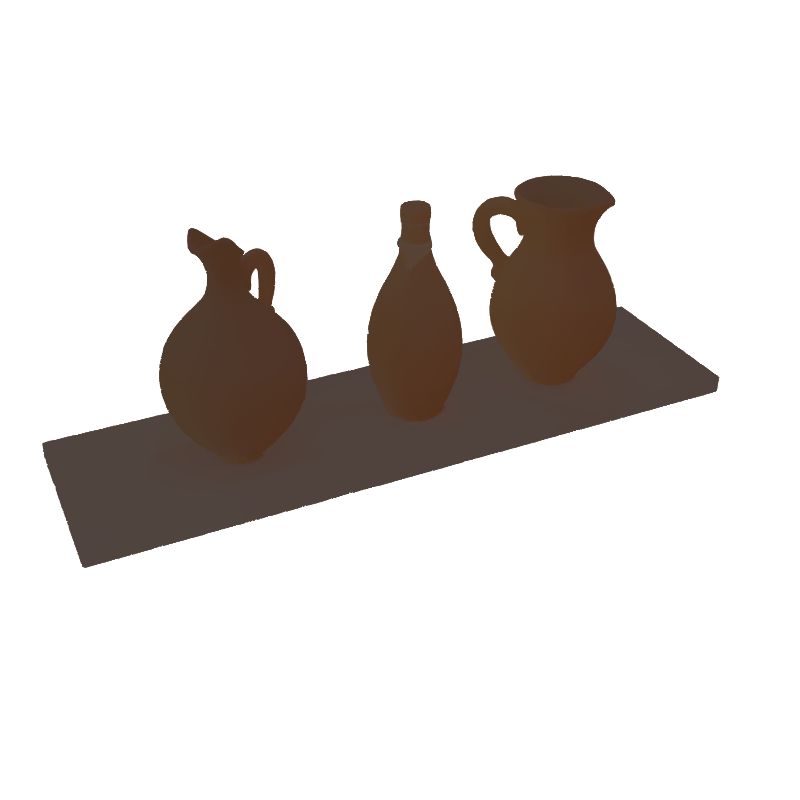}}  &  
                \multicolumn{1}{c}{\includegraphics[width=0.12\linewidth,trim={0cm 8cm 2cm 4cm},clip]{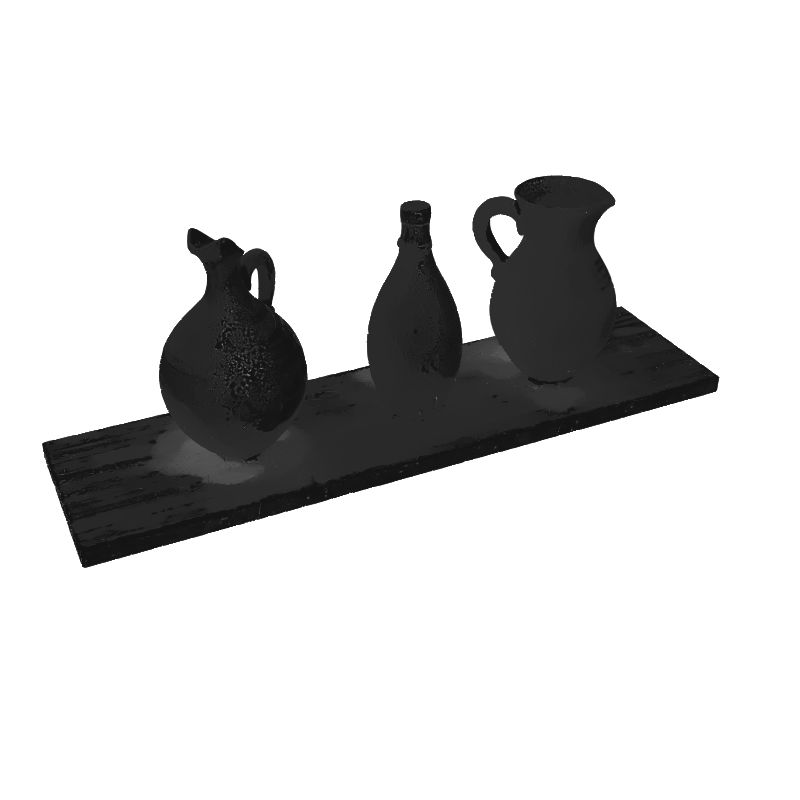}}  &  
                \multicolumn{1}{c}{\includegraphics[width=0.12\linewidth,trim={0.5cm 6cm 0.5cm 4cm},clip]{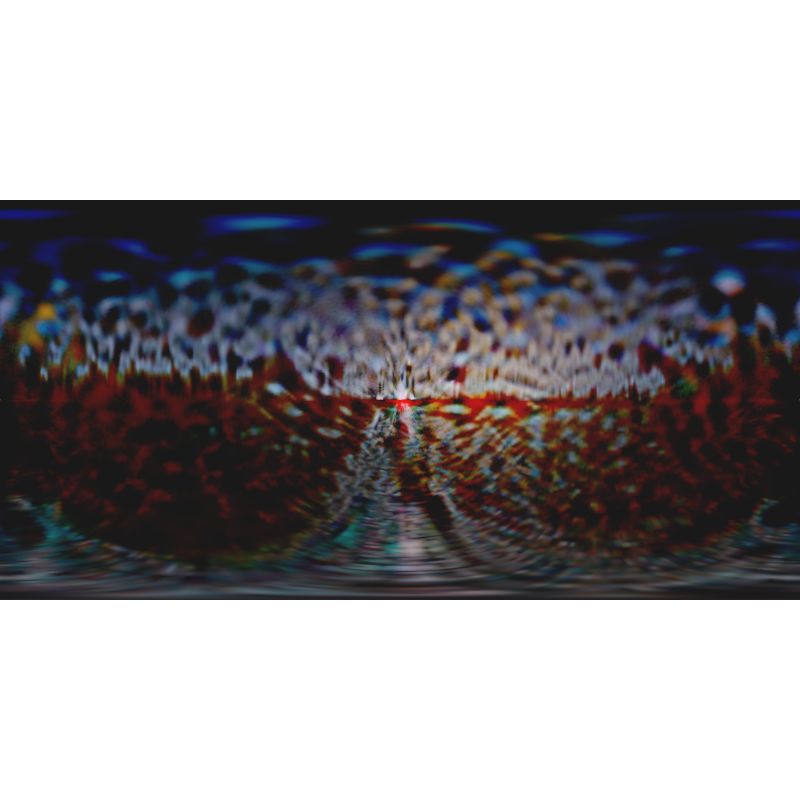}}  &
                \multicolumn{1}{c}{\includegraphics[width=0.12\linewidth,trim={0cm 3.5cm 0cm 4cm},clip]{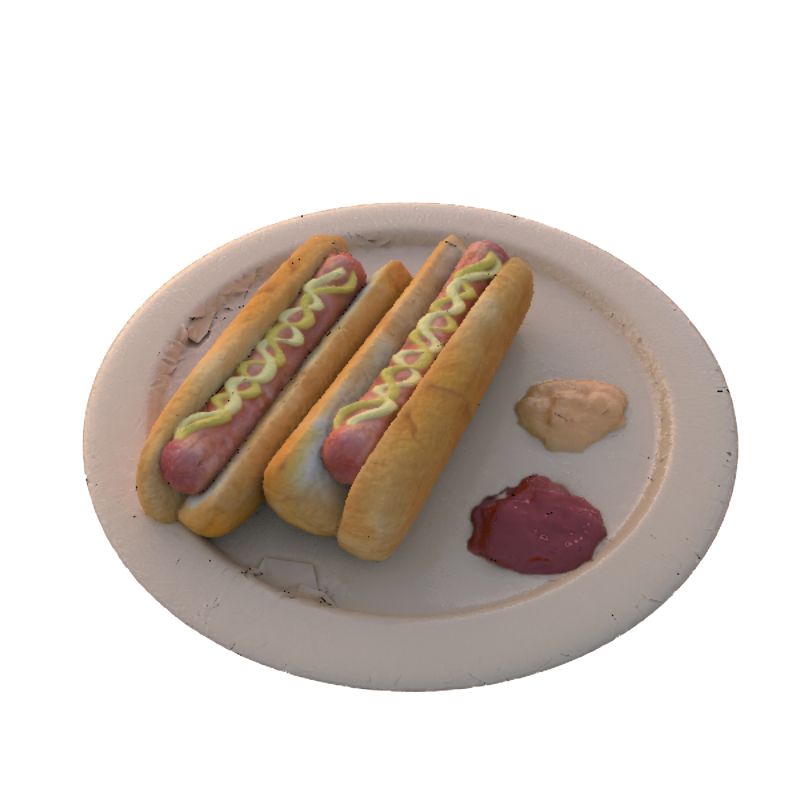}}  &  
                \multicolumn{1}{c}{\includegraphics[width=0.12\linewidth,trim={0cm 3.5cm 0cm 4cm},clip]{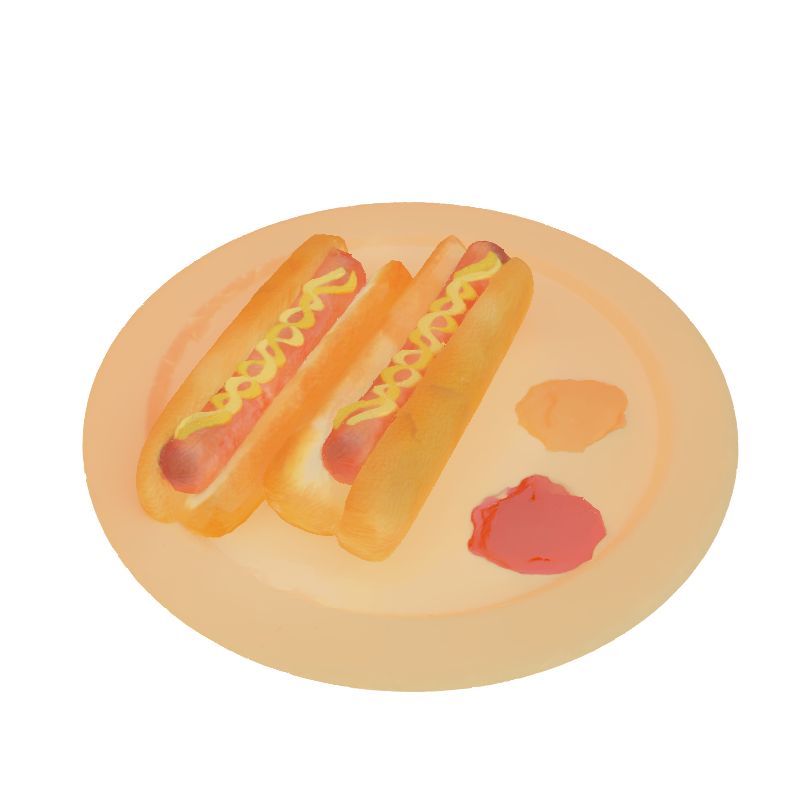}}  &  
                \multicolumn{1}{c}{\includegraphics[width=0.12\linewidth,trim={0cm 3.5cm 0cm 4cm},clip]{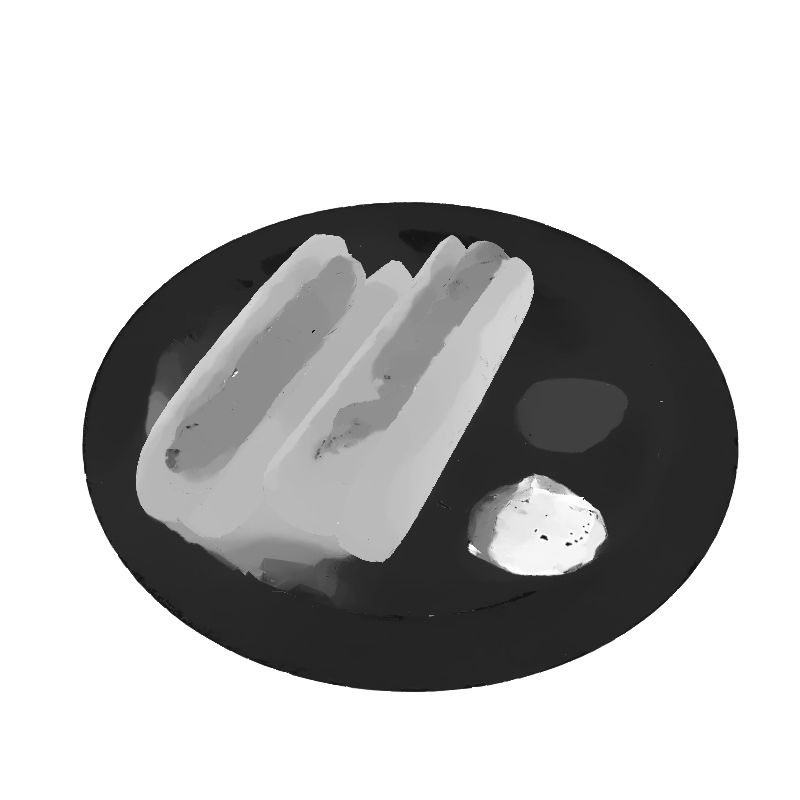}}  &  
                \multicolumn{1}{c}{\includegraphics[width=0.12\linewidth,trim={0.5cm 6cm 0.5cm 4cm},clip]{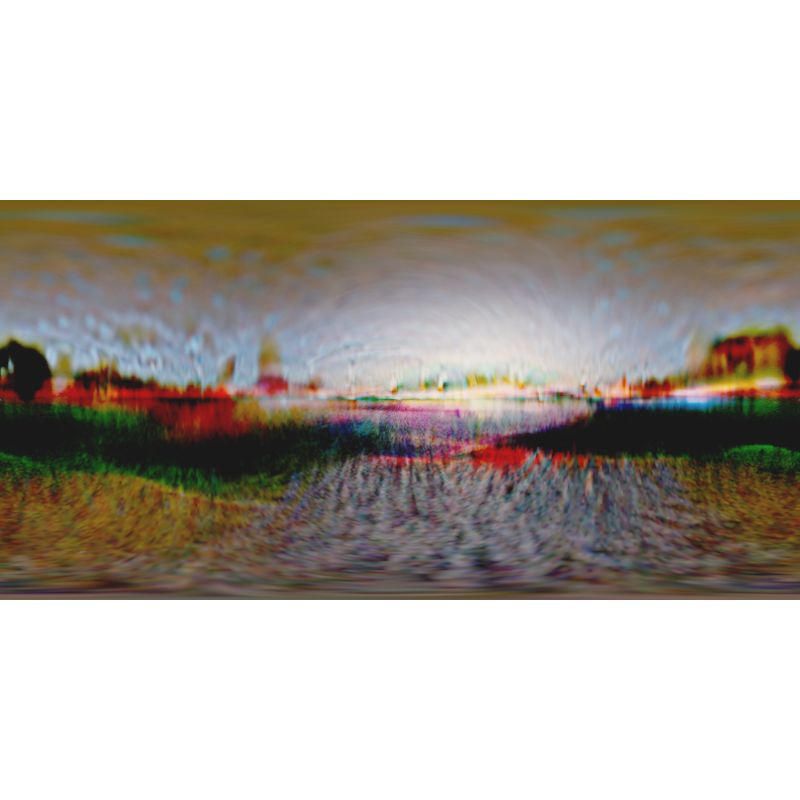}}  \\
                
                \rotatebox{90}{\footnotesize{~~~{R3DG}}} &
                \multicolumn{1}{c}{\includegraphics[width=0.12\linewidth,trim={0cm 8cm 2cm 4cm},clip]{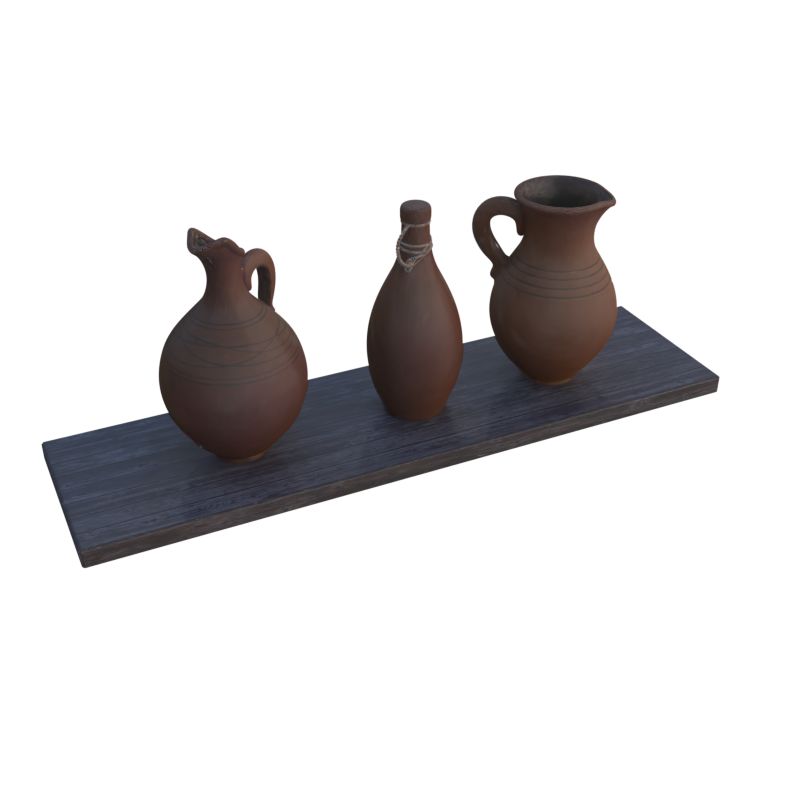}}  &  
                \multicolumn{1}{c}{\includegraphics[width=0.12\linewidth,trim={0cm 8cm 2cm 4cm},clip]{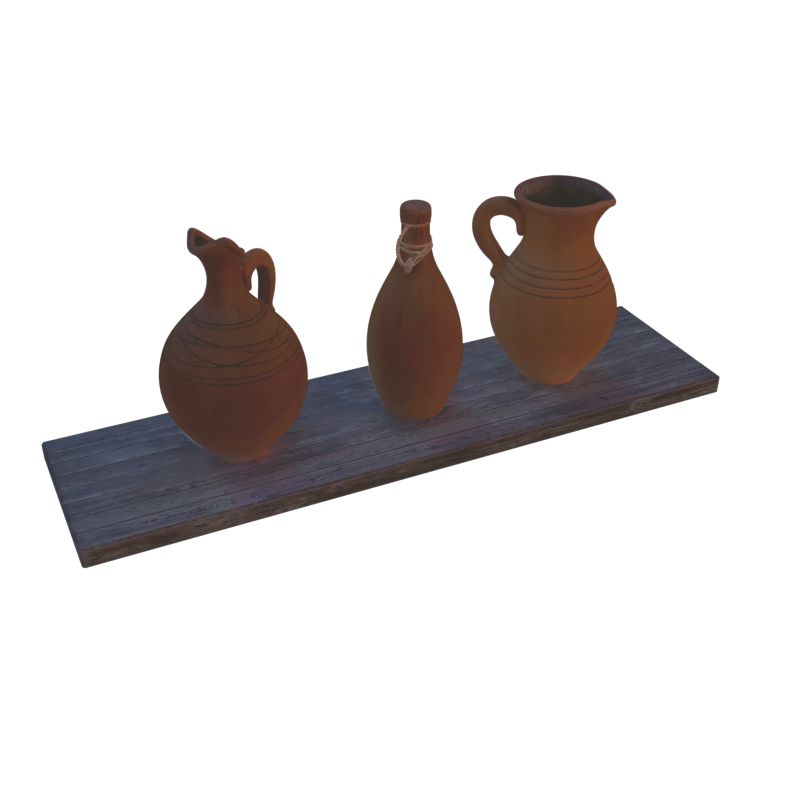}}  &  
                \multicolumn{1}{c}{\includegraphics[width=0.12\linewidth,trim={0cm 8cm 2cm 4cm},clip]{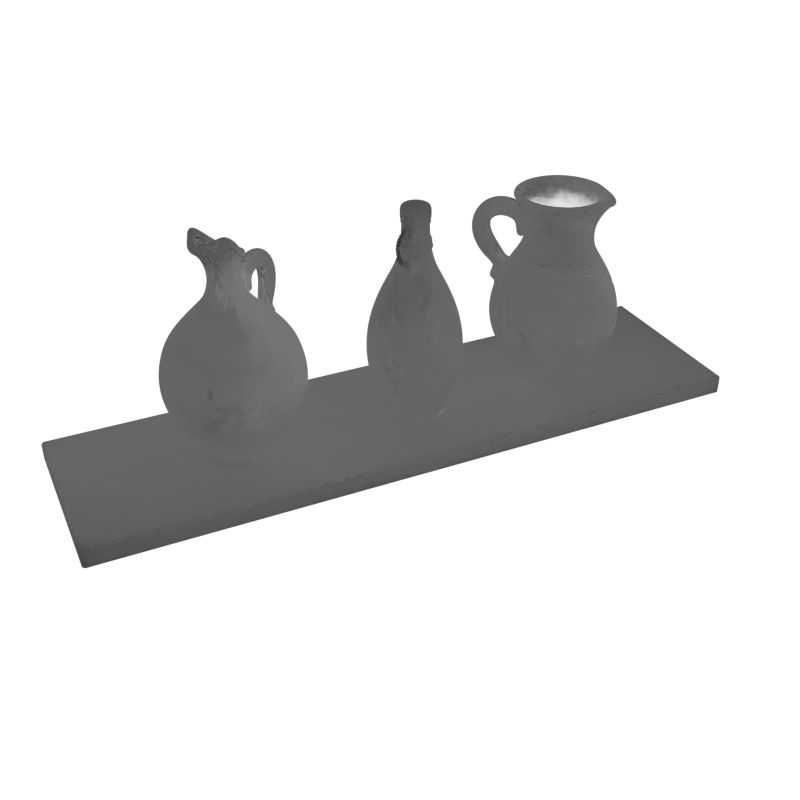}}  &  
                \multicolumn{1}{c}{\includegraphics[width=0.12\linewidth,trim={0.5cm 6cm 0.5cm 4cm},clip]{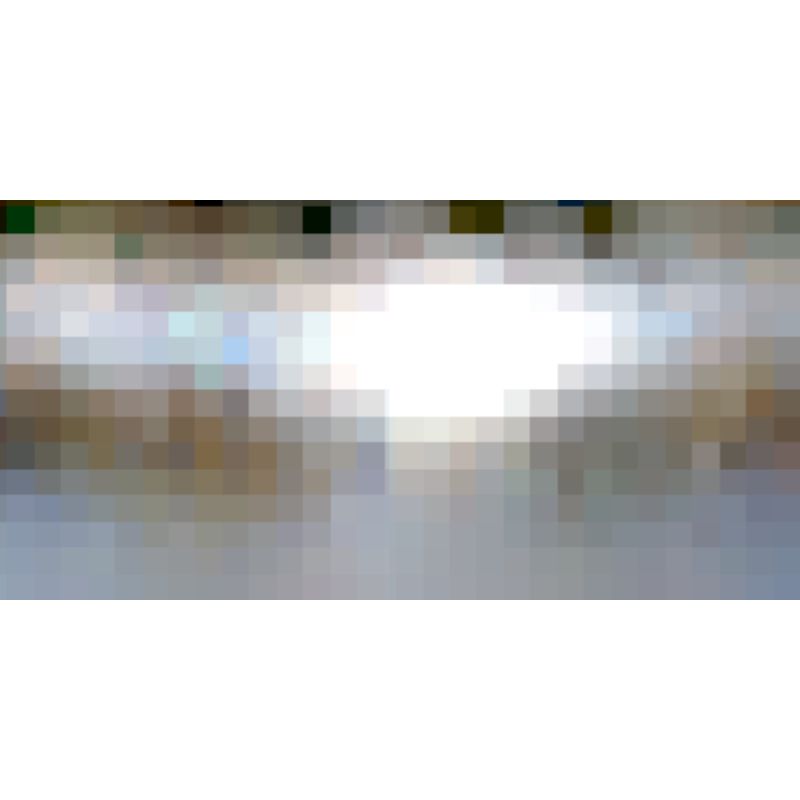}}  &
                \multicolumn{1}{c}{\includegraphics[width=0.12\linewidth,trim={0cm 3.5cm 0cm 4cm},clip]{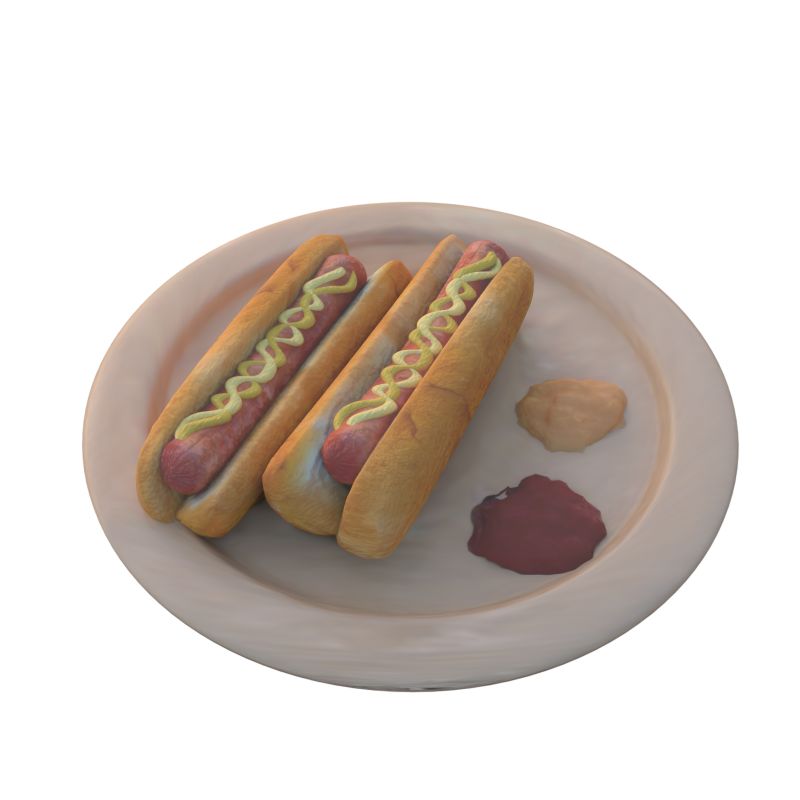}}  &  
                \multicolumn{1}{c}{\includegraphics[width=0.12\linewidth,trim={0cm 3.5cm 0cm 4cm},clip]{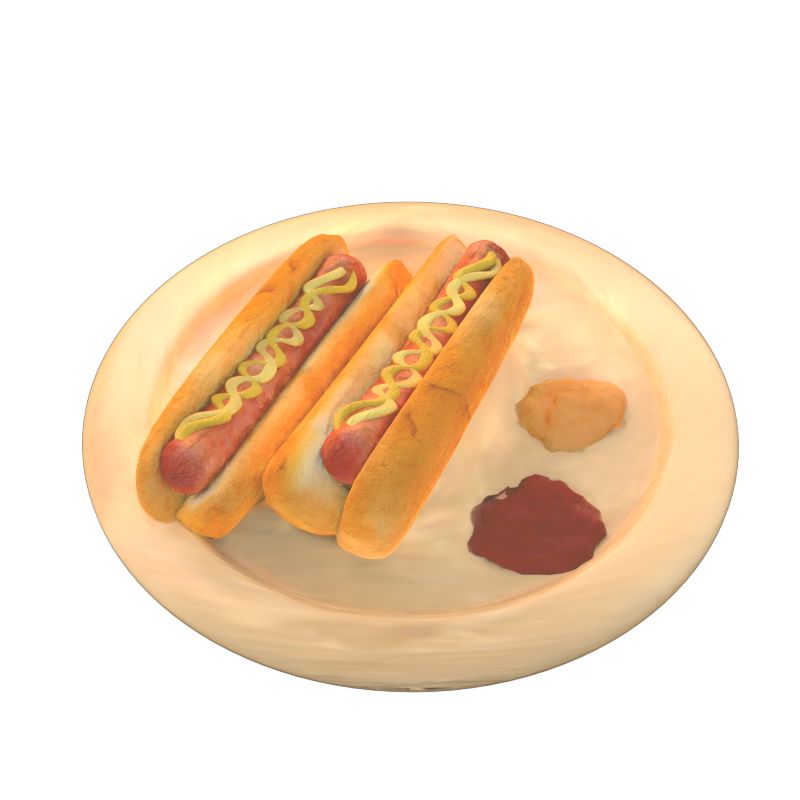}}  &  
                \multicolumn{1}{c}{\includegraphics[width=0.12\linewidth,trim={0cm 3.5cm 0cm 4cm},clip]{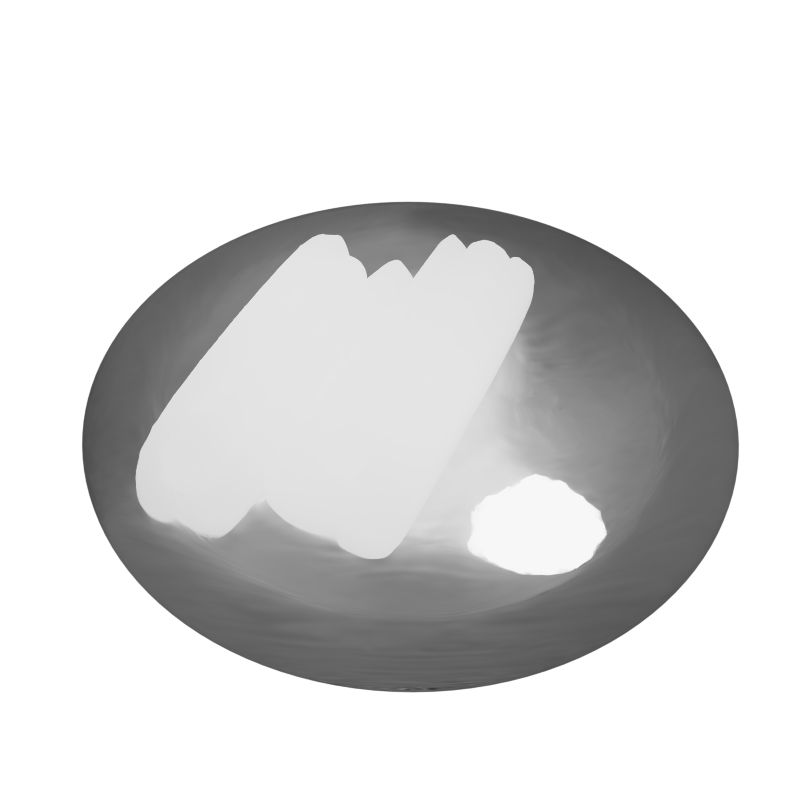}}  &  
                \multicolumn{1}{c}{\includegraphics[width=0.12\linewidth,trim={0.5cm 6cm 0.5cm 4cm},clip]{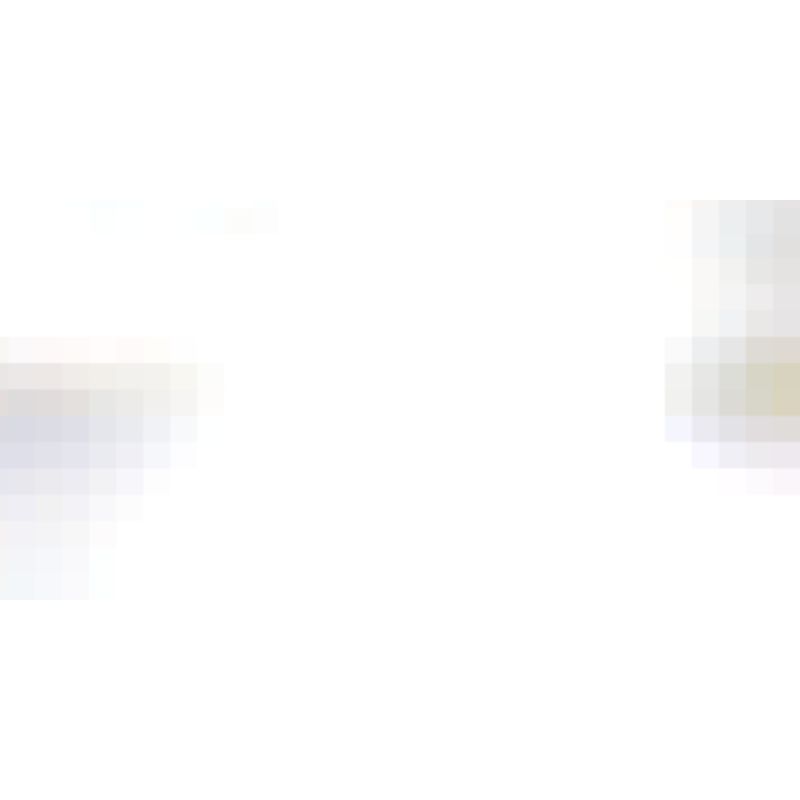}}  \\

                \rotatebox{90}{\footnotesize{~~~{TensoIR}}} &
                \multicolumn{1}{c}{\includegraphics[width=0.12\linewidth,trim={0cm 8cm 2cm 4cm},clip]{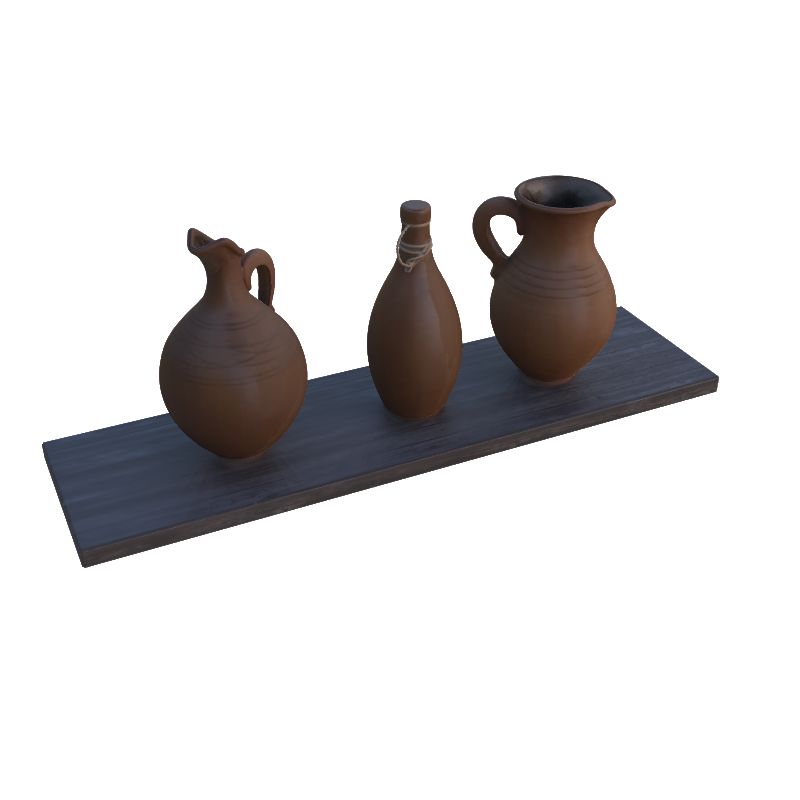}}  &  
                \multicolumn{1}{c}{\includegraphics[width=0.12\linewidth,trim={0cm 8cm 2cm 4cm},clip]{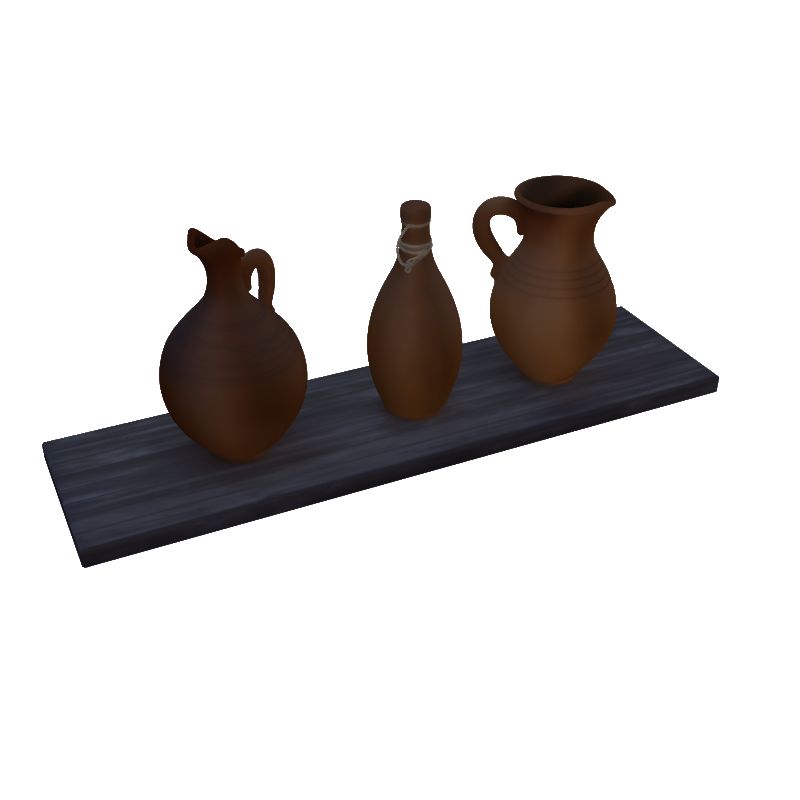}}  &  
                \multicolumn{1}{c}{\includegraphics[width=0.12\linewidth,trim={0cm 8cm 2cm 4cm},clip]{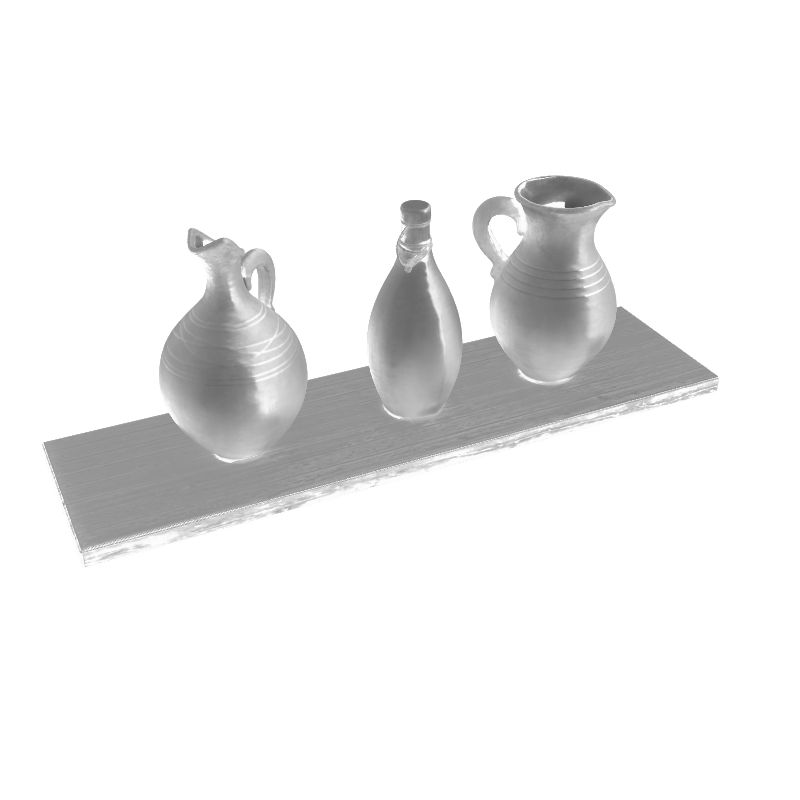}}  &  
                \multicolumn{1}{c}{\includegraphics[width=0.12\linewidth,trim={0.5cm 6cm 0.5cm 4cm},clip]{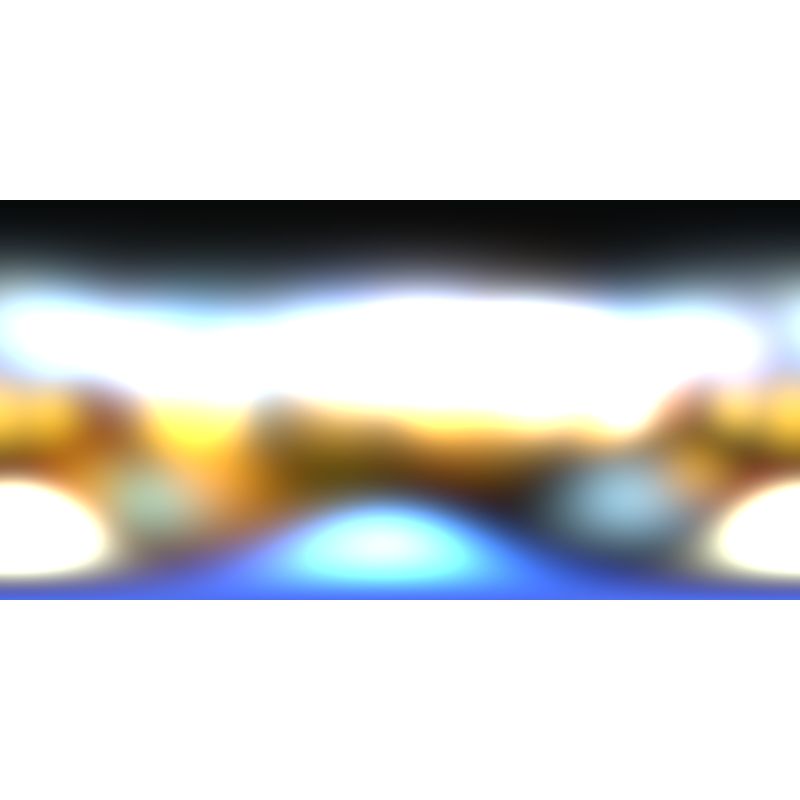}}  &
                \multicolumn{1}{c}{\includegraphics[width=0.12\linewidth,trim={0cm 3.5cm 0cm 4cm},clip]{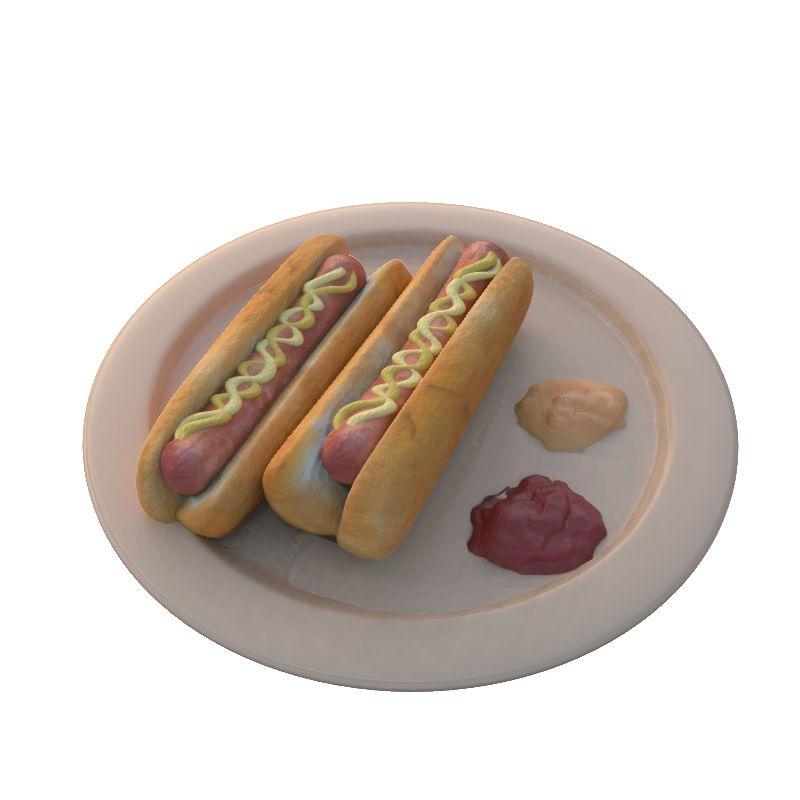}}  &  
                \multicolumn{1}{c}{\includegraphics[width=0.12\linewidth,trim={0cm 3.5cm 0cm 4cm},clip]{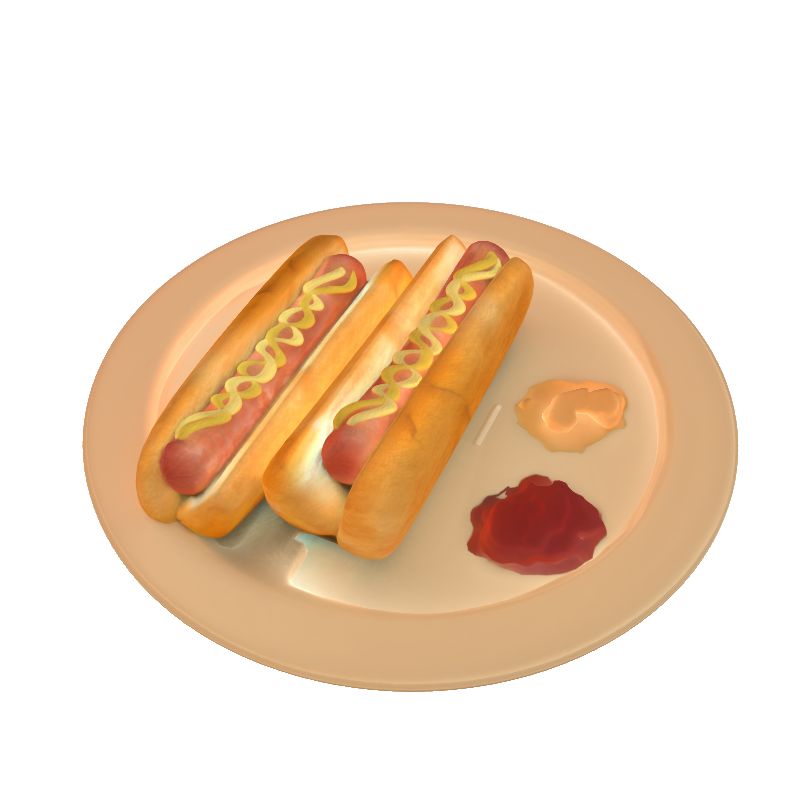}}  &  
                \multicolumn{1}{c}{\includegraphics[width=0.12\linewidth,trim={0cm 3.5cm 0cm 4cm},clip]{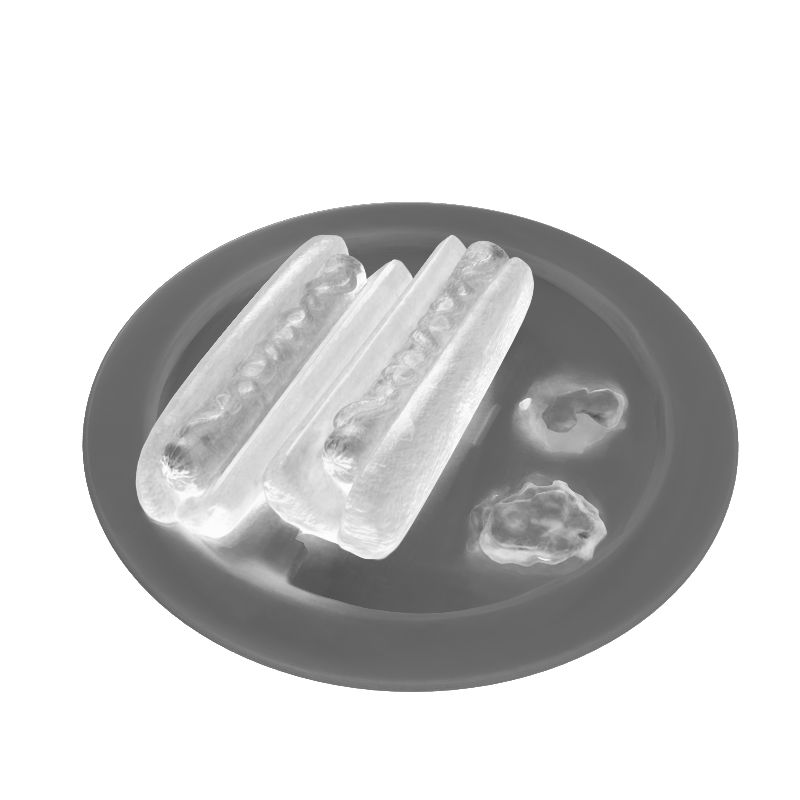}}  &  
                \multicolumn{1}{c}{\includegraphics[width=0.12\linewidth,trim={0.5cm 6cm 0.5cm 4cm},clip]{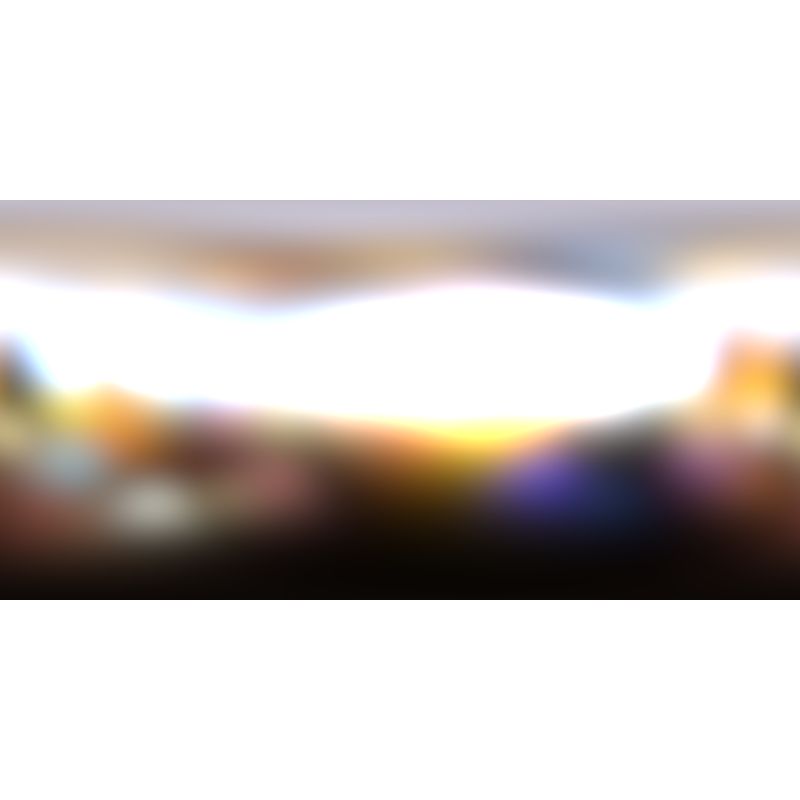}}  \\
                
                \rotatebox{90}{\footnotesize{~~~{Ours}}} &
                \multicolumn{1}{c}{\includegraphics[width=0.12\linewidth,trim={0cm 8cm 2cm 4cm},clip]{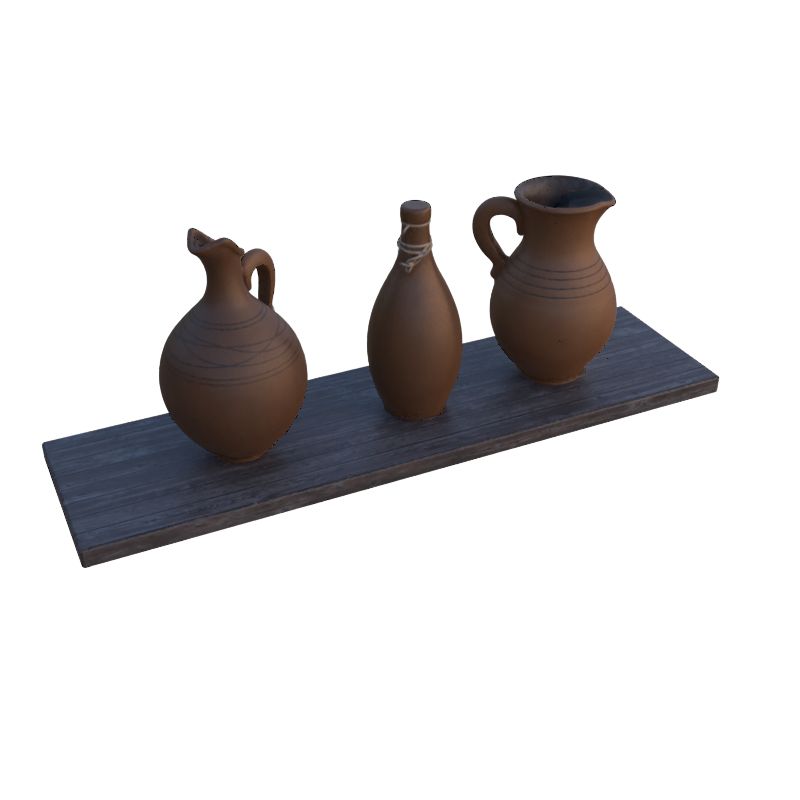}}  &  
                \multicolumn{1}{c}{\includegraphics[width=0.12\linewidth,trim={0cm 8cm 2cm 4cm},clip]{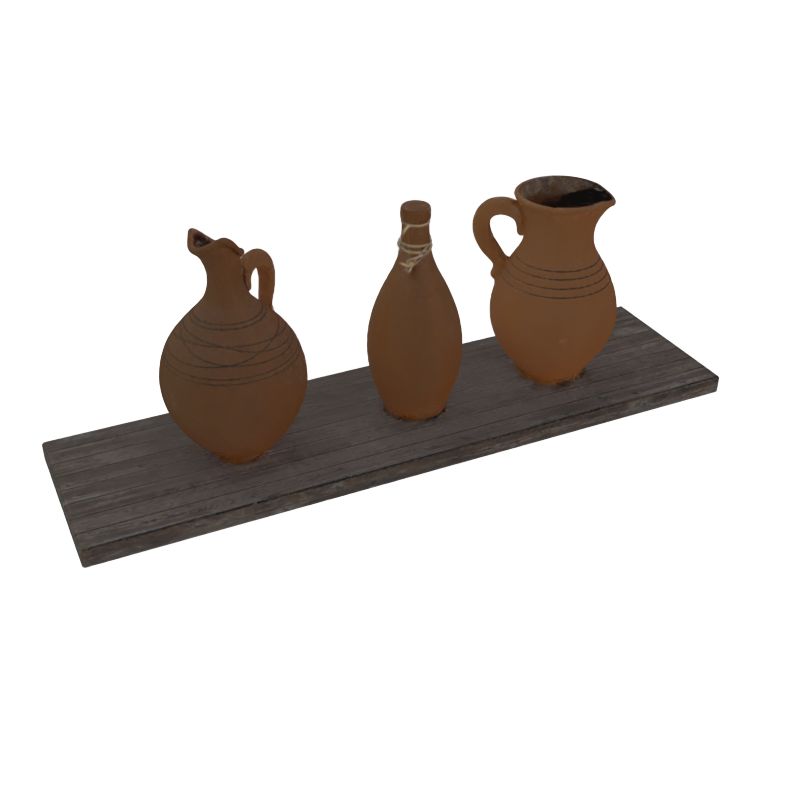}}  &  
                \multicolumn{1}{c}{\includegraphics[width=0.12\linewidth,trim={0cm 8cm 2cm 4cm},clip]{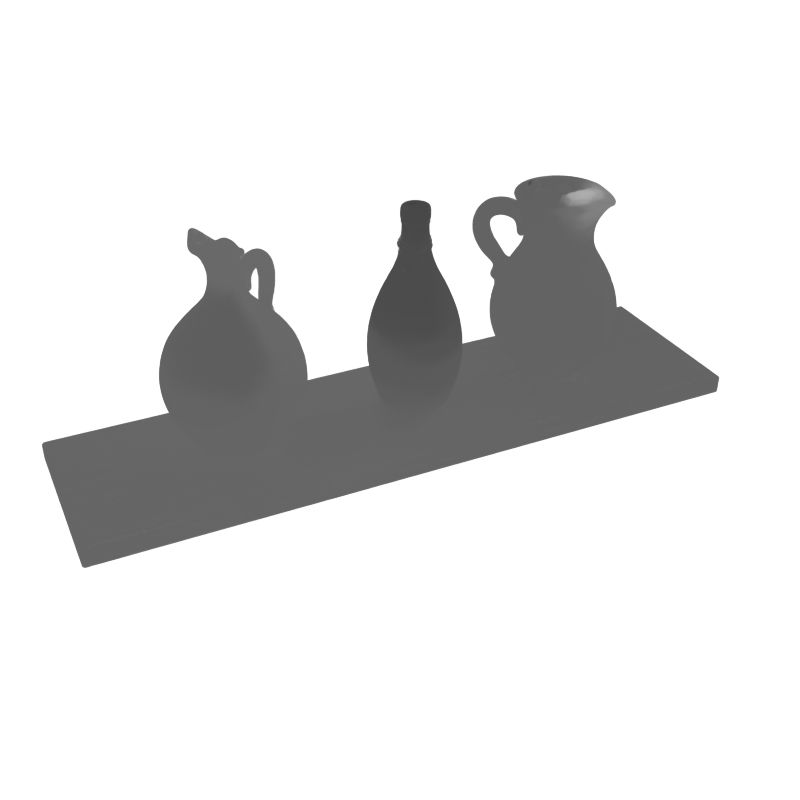}}  &  
                \multicolumn{1}{c}{\includegraphics[width=0.12\linewidth,trim={0.5cm 6cm 0.5cm 4cm},clip]{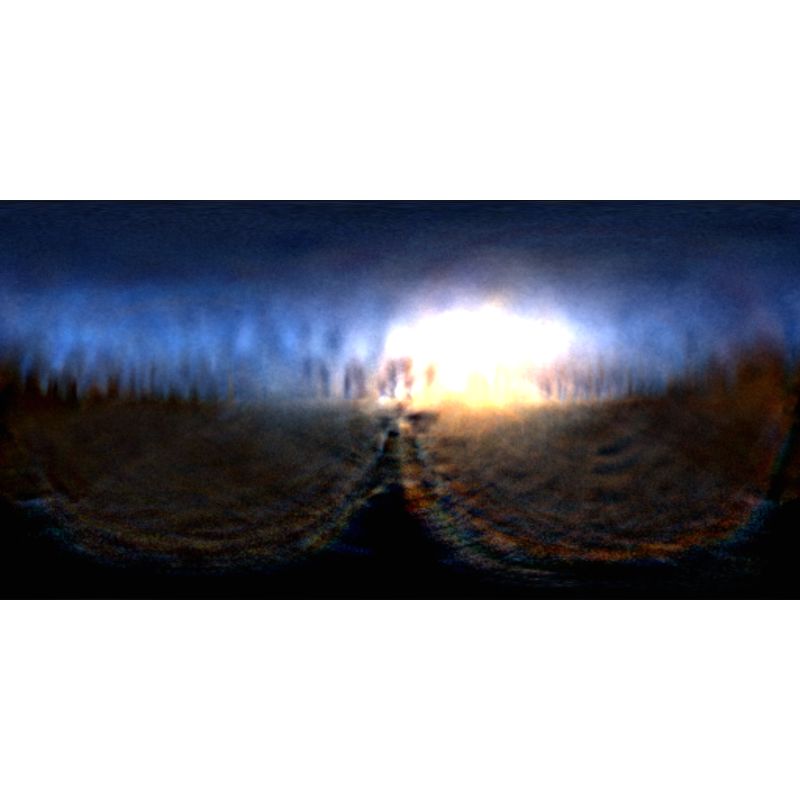}}  & 
                \multicolumn{1}{c}{\includegraphics[width=0.12\linewidth,trim={0cm 3.5cm 0cm 4cm},clip]{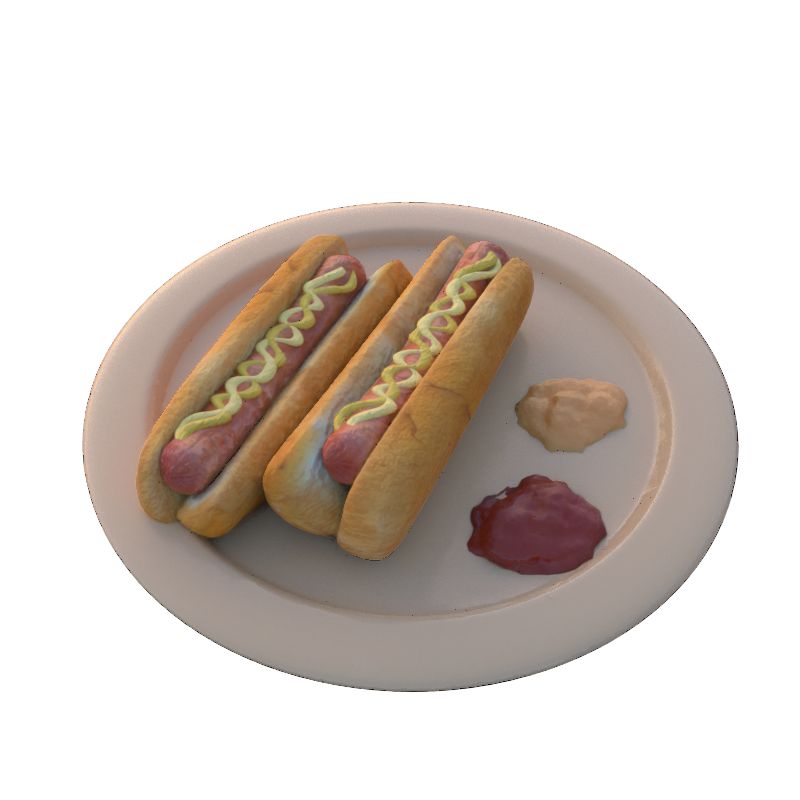}}  &  
                \multicolumn{1}{c}{\includegraphics[width=0.12\linewidth,trim={0cm 3.5cm 0cm 4cm},clip]{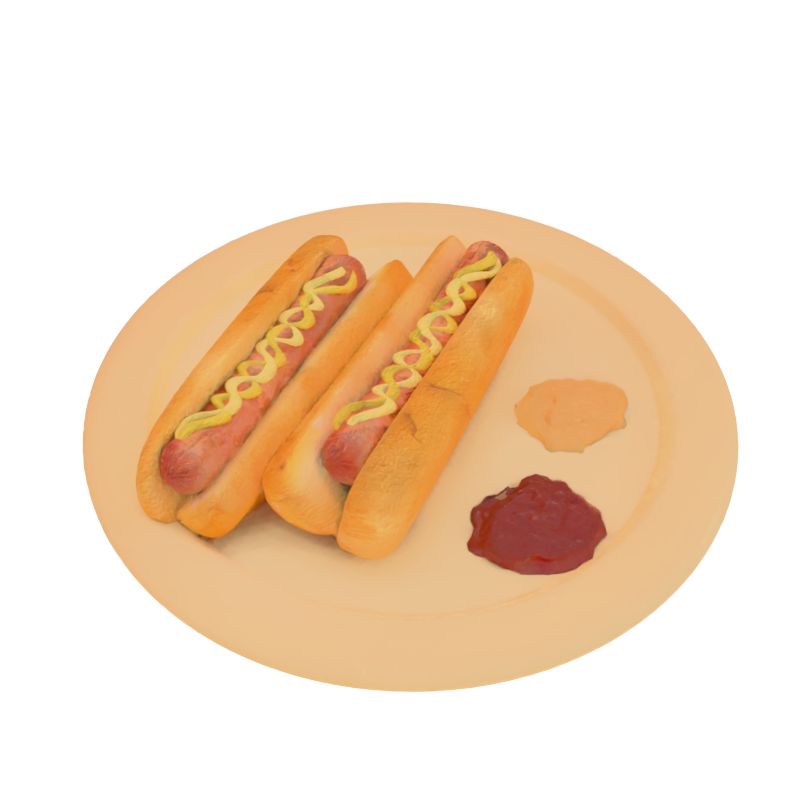}}  &  
                \multicolumn{1}{c}{\includegraphics[width=0.12\linewidth,trim={0cm 3.5cm 0cm 4cm},clip]{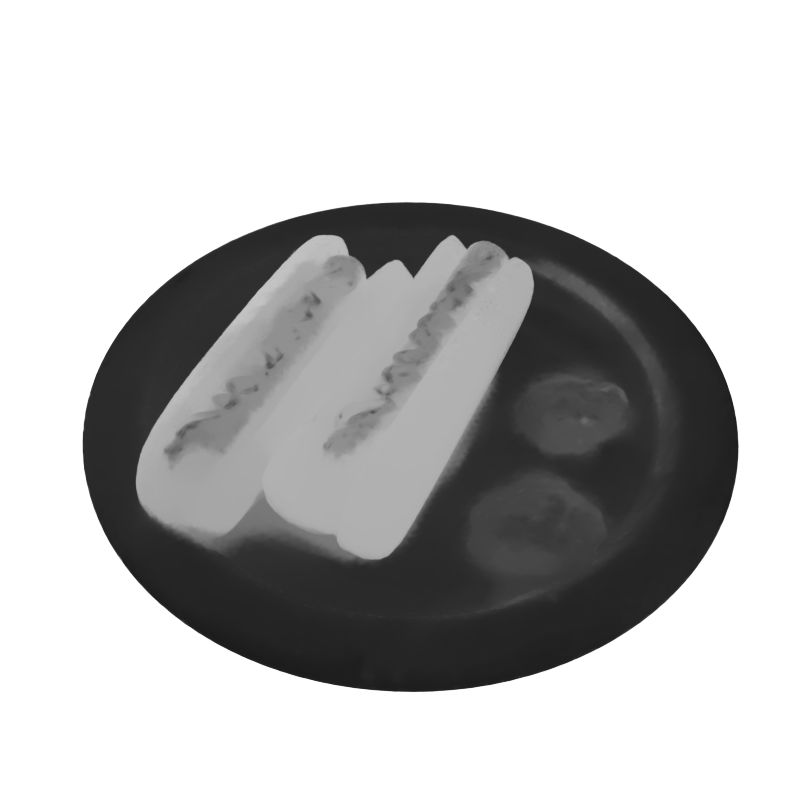}}  &  
                \multicolumn{1}{c}{\includegraphics[width=0.12\linewidth,trim={0.5cm 6cm 0.5cm 4cm},clip]{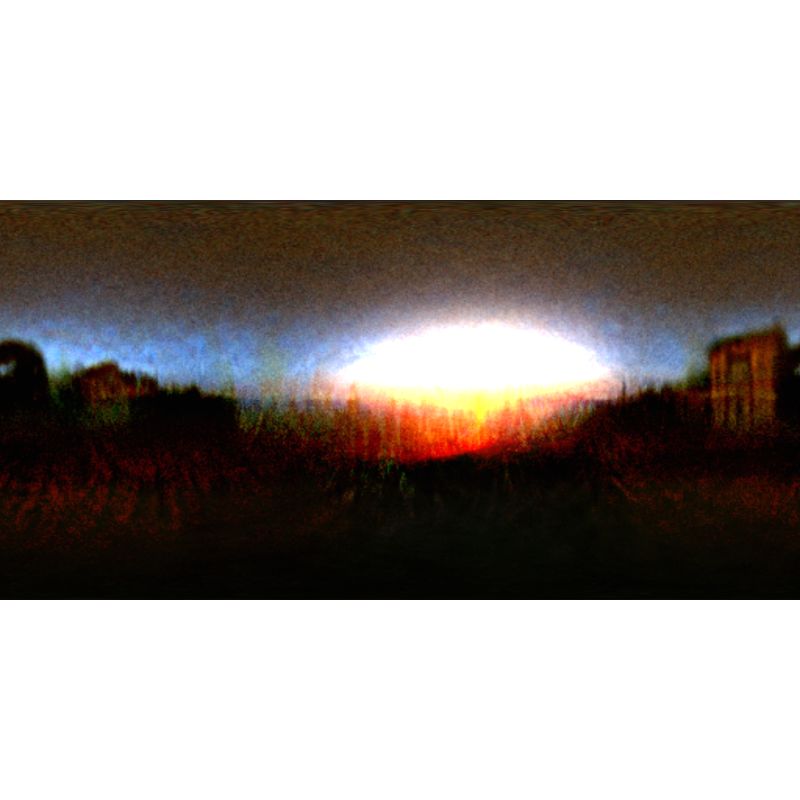}}  \\
                
                \rotatebox{90}{\footnotesize{~{Reference}}} &
                \multicolumn{1}{c}{\includegraphics[width=0.12\linewidth,trim={0cm 8cm 2cm 4cm},clip]{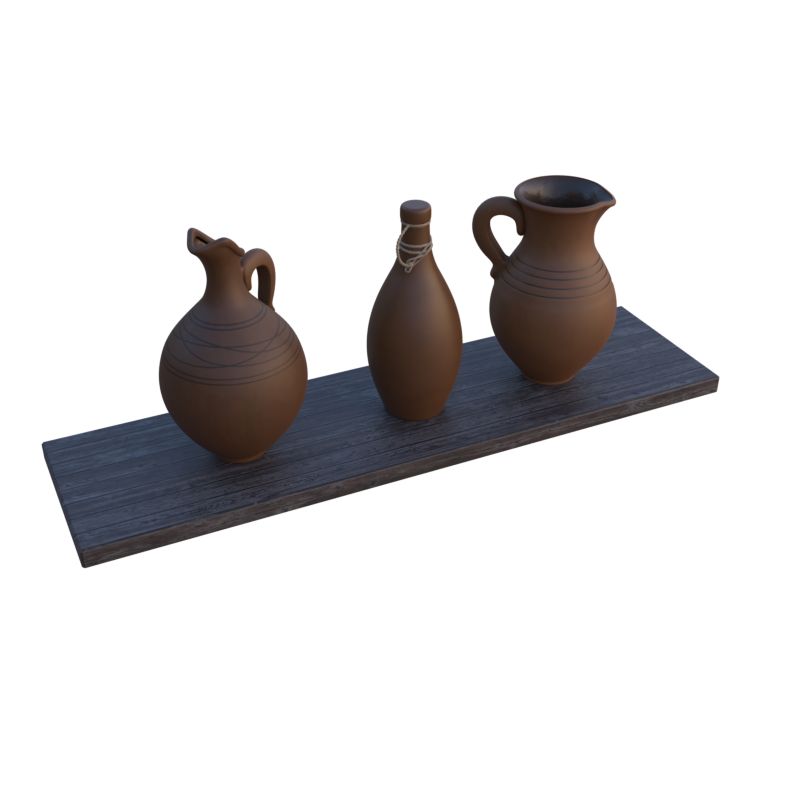}}  &  
                \multicolumn{1}{c}{\includegraphics[width=0.12\linewidth,trim={0cm 8cm 2cm 4cm},clip]{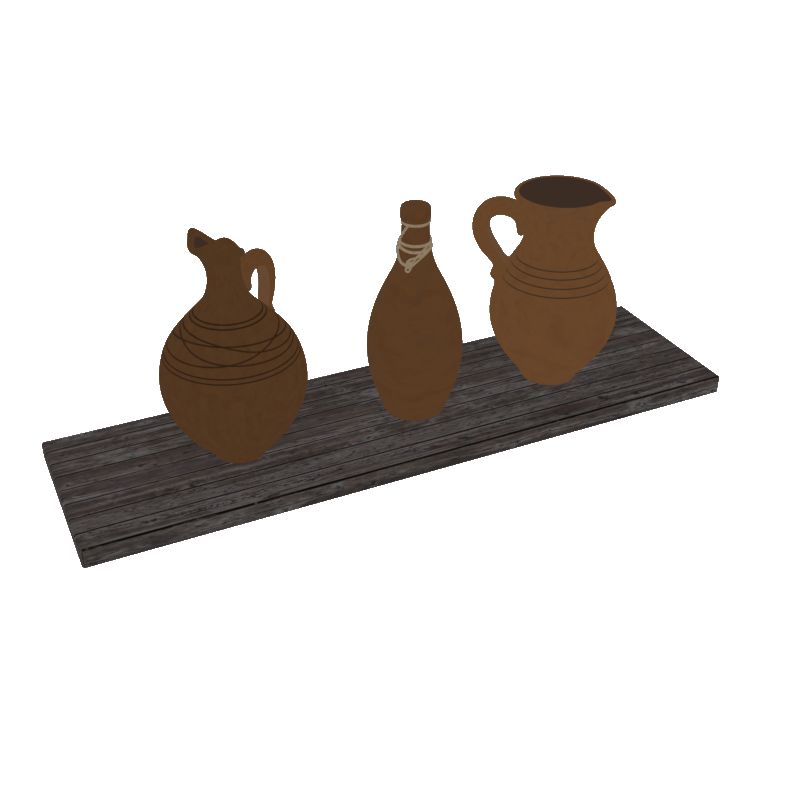}}  &  
                \multicolumn{1}{c}{\includegraphics[width=0.12\linewidth,trim={0cm 8cm 2cm 4cm},clip]{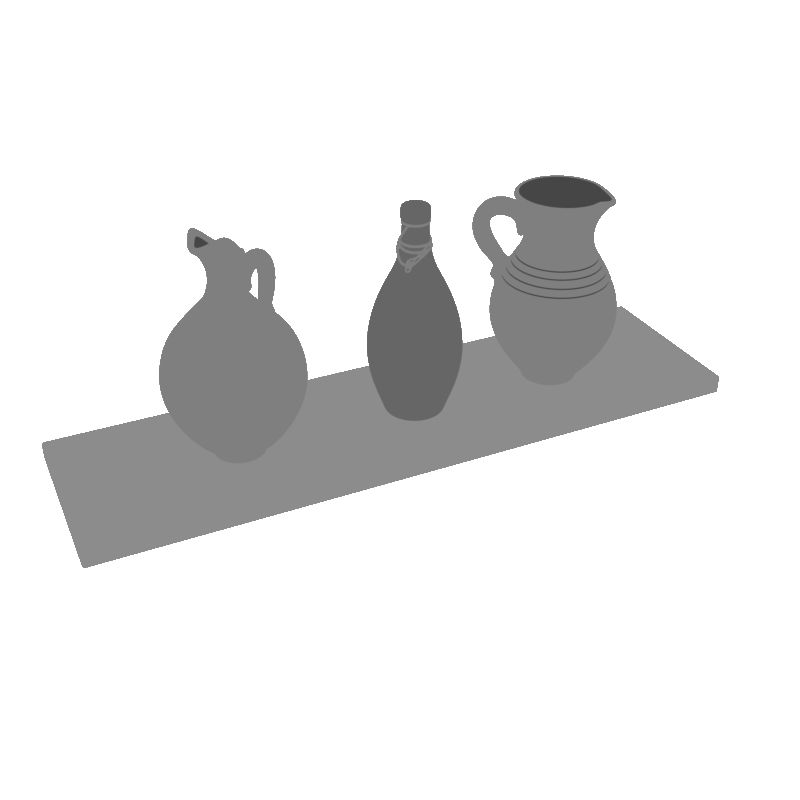}}  &  
                \multicolumn{1}{c}{\includegraphics[width=0.12\linewidth,trim={0.5cm 6cm 0.5cm 4cm},clip]{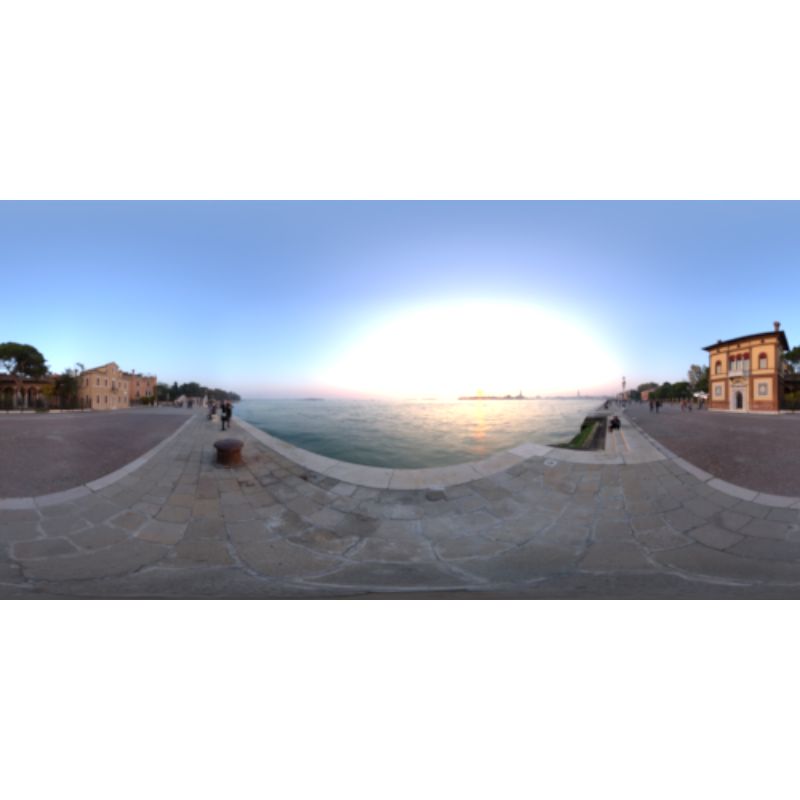}}  &
                \multicolumn{1}{c}{\includegraphics[width=0.12\linewidth,trim={0cm 3.5cm 0cm 4cm},clip]{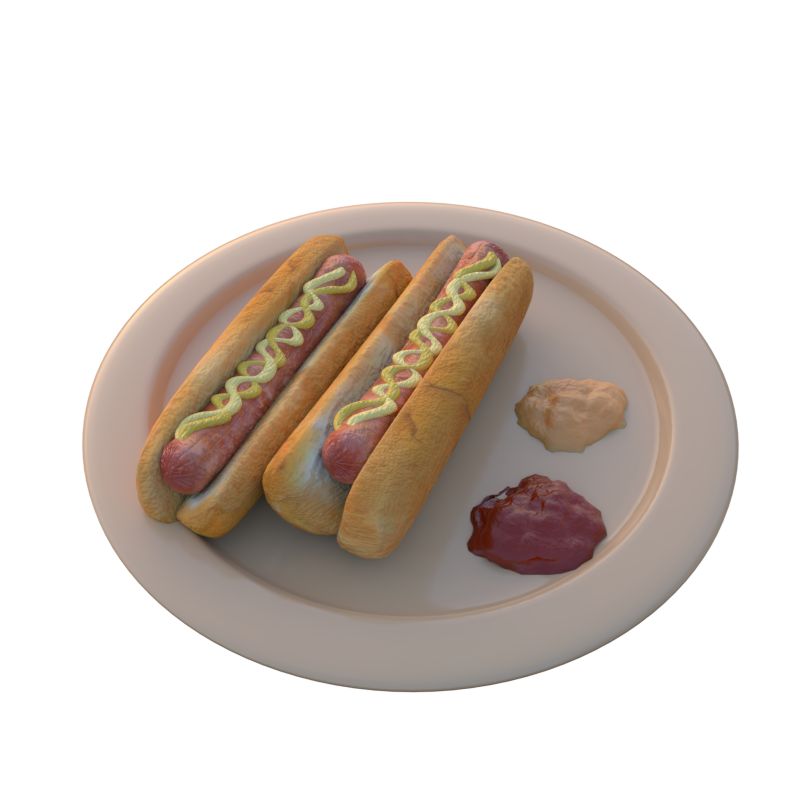}}  &  
                \multicolumn{1}{c}{\includegraphics[width=0.12\linewidth,trim={0cm 3.5cm 0cm 4cm},clip]{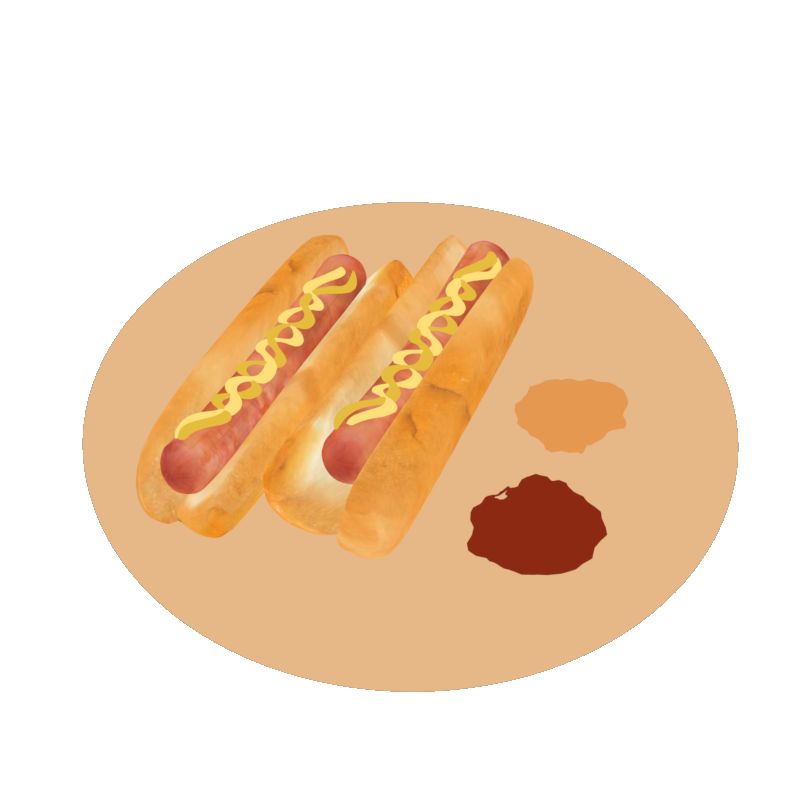}}  &  
                \multicolumn{1}{c}{\includegraphics[width=0.12\linewidth,trim={0cm 3.5cm 0cm 4cm},clip]{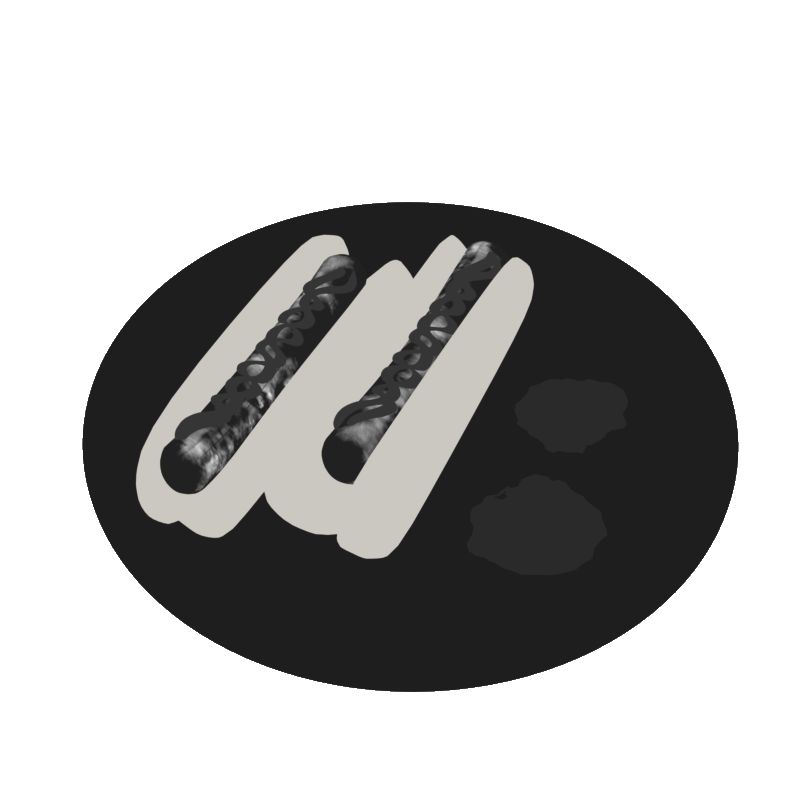}}  &  
                \multicolumn{1}{c}{\includegraphics[width=0.12\linewidth,trim={0.5cm 6cm 0.5cm 4cm},clip]{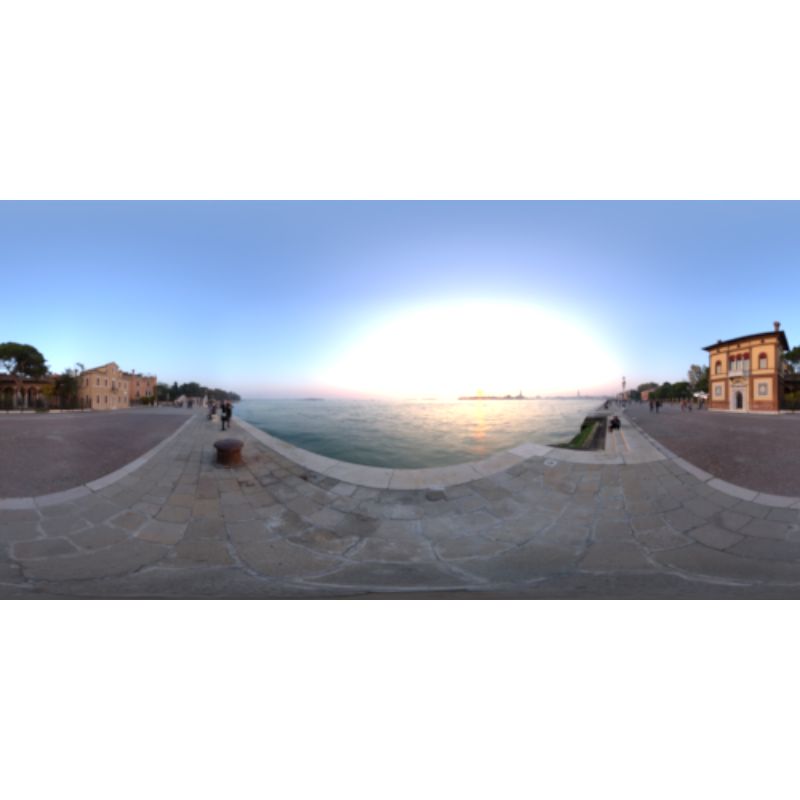}}  \\
            \end{tabular}
        \end{tabular}
    }
    \end{center}
    \vspace{-6mm}
    \caption{ 
        \textbf{Qualitative Results from the Synthetic4Relight dataset.} Compared to the baselines, our method achieves superior inverse rendering performance in albedo, roughness, and environment lighting recovery.
        \label{fig:exp:relight}
    }
    \vspace{-4mm}
\end{figure*}

\vspace{-1mm}
\section{Experiments}
We perform extensive experiments to verify the effectiveness of our inverse rendering method.
We first present material decomposition and relighting results on three different datasets in Sec.~\ref{exp:relighting}. Subsequently, we report normal quality in Sec.~\ref{exp:geo}, real-world results in Sec.~\ref{exp:real}, and ablation studies in Sec.~\ref{exp:abl}.
Our method achieves state-of-the-art inverse rendering performance while demonstrating exceptional efficiency, completing training within 10-15 minutes on a single NVIDIA GTX 4090.

\subsection{Inverse Rendering Performance}
\label{exp:relighting}
\paragraph{Datasets \& Metrics.} We evaluate material decomposition and relighting performance on the Synthetic4Relight dataset~\citep{zhang2022invrender}, the TensoIR Synthetic dataset \cite{jin2023tensoir}, and the Shiny Blender dataset \cite{verbin2022ref}, which cover various challenging scenes with complex non-Lambertian materials, indirect lighting effects, and strong reflection. % Following previous work, we train on 100 input views and evaluate on 200 test views. 
Following~\citep{R3DG2023}, we assess relighting and albedo quality using PSNR, SSIM, and LPIPS metrics, along with roughness MSE, as shown in Table~\ref{table:exp:relight}. For the Shiny Blender dataset consisting of reflective objects, we follow prior works \cite{verbin2022ref,liu2023nero,jiang2023gaussianshader} to evaluate novel view synthesis (NVS) performance.

\vspace{-3mm}
\paragraph{Performance \& Discussion.}
We compare our method with mesh-based (NVdiffrec), implicit-field-based (TensoIR, NeRO), and 3DGS-based ({\yaoyao}, {\gaussianshader}, GS-IR) relightable approaches.
As shown in Table~\ref{table:exp:relight} and Fig.~\ref{fig:exp:relight}, our method achieves state-of-the-art performance in relighting and material assessment, demonstrating superior material and lighting disentanglement, while also delivering exceptional training efficiency.

\subsection{Normal Quality}
\label{exp:geo}
\paragraph{Datasets \& Metrics.} We conduct normal quality comparison on the TensoIR Synthetic dataset \cite{jin2023tensoir} and the Shiny Blender dataset to validate effectiveness of our geometry guidance, covering both diffuse objects and specular objects. Following prior works~\cite{jin2023tensoir,verbin2022ref,zhang2022invrender}, the normal accuracy is evaluated in terms of Mean Angular Error (MAE).

\begin{figure}[H]
\vspace{-6.1mm}
    \hspace*{-0mm}
    \begin{center}
    \resizebox{\linewidth}{!}{
        \setlength{\tabcolsep}{1pt}
        \setlength{\fboxrule}{1pt}
        \begin{tabular}{c}
            \begin{tabular}{cccccc}
                \rotatebox{90}{\scriptsize{~~~~~{R3DG}}} &
                \includegraphics[width=0.18\linewidth,trim={2cm 6cm 2cm 1cm},clip]{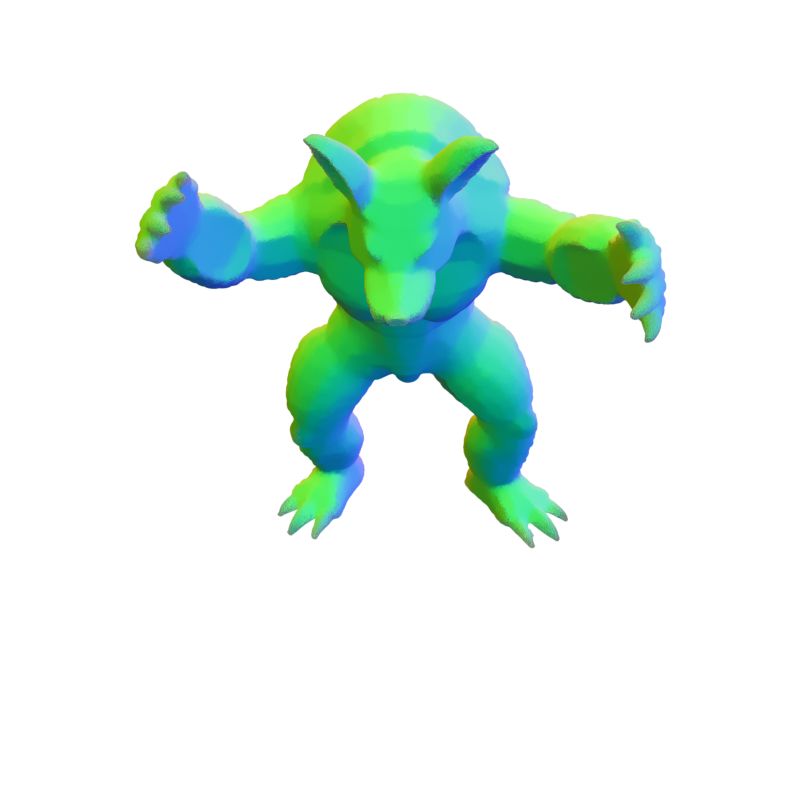}  &
                \includegraphics[width=0.18\linewidth,trim={1.5cm 1cm 0cm 4cm},clip]{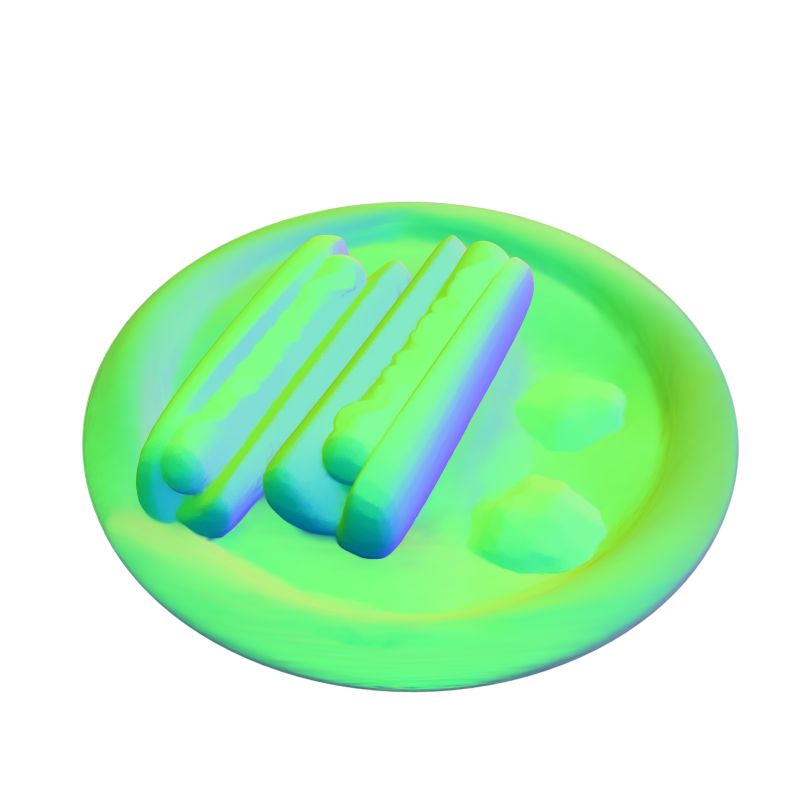}  &
                \includegraphics[width=0.18\linewidth,trim={0cm 2cm 0cm 0cm},clip]{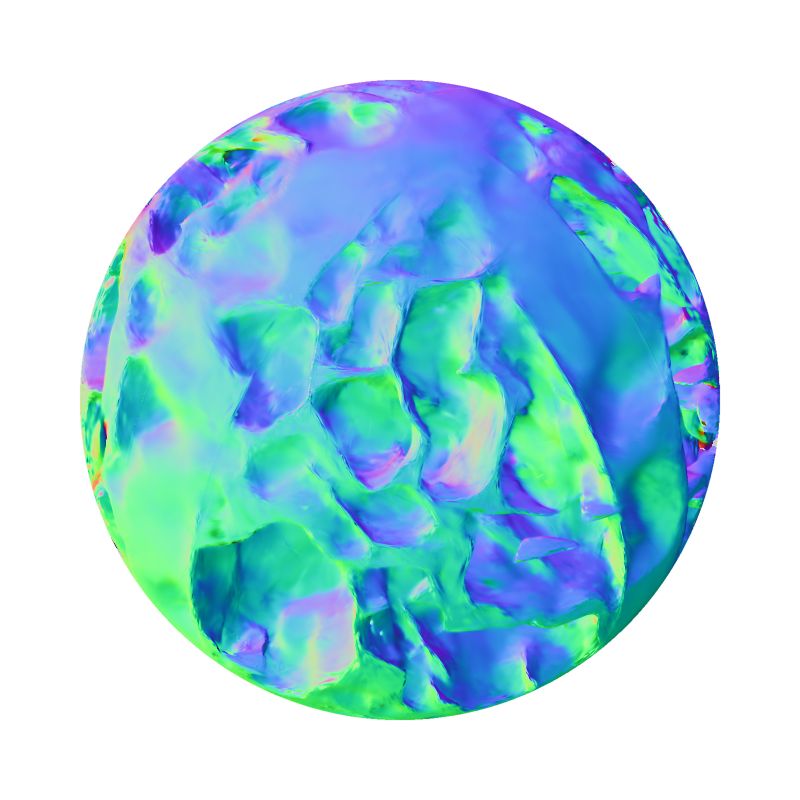}  &
                \includegraphics[width=0.22\linewidth,trim={5cm 9cm 5cm 5cm},clip]{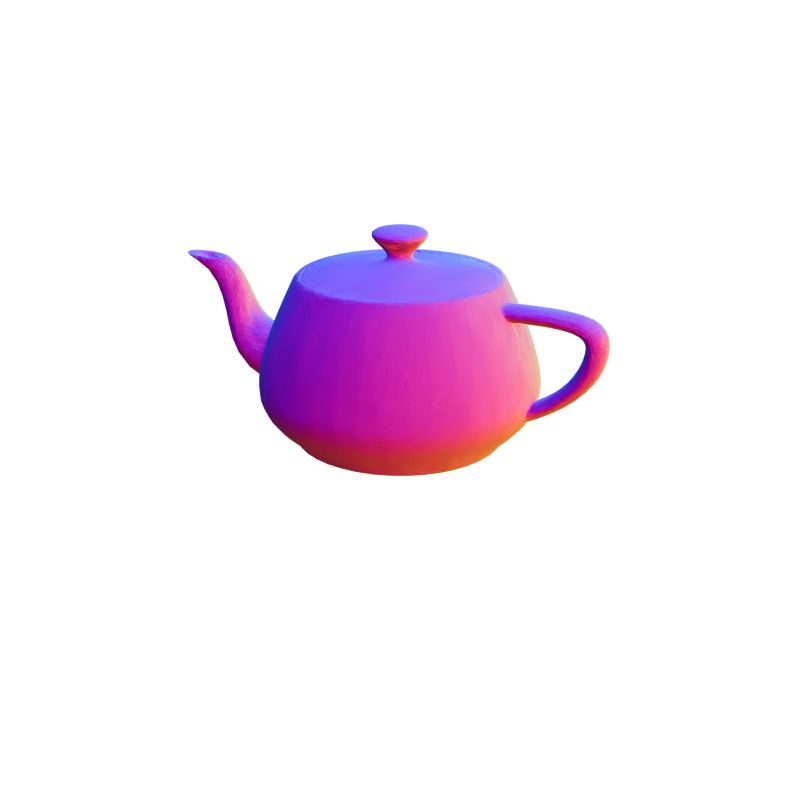}  &
                \includegraphics[width=0.16\linewidth,trim={2cm 0cm 0cm 4cm},clip]{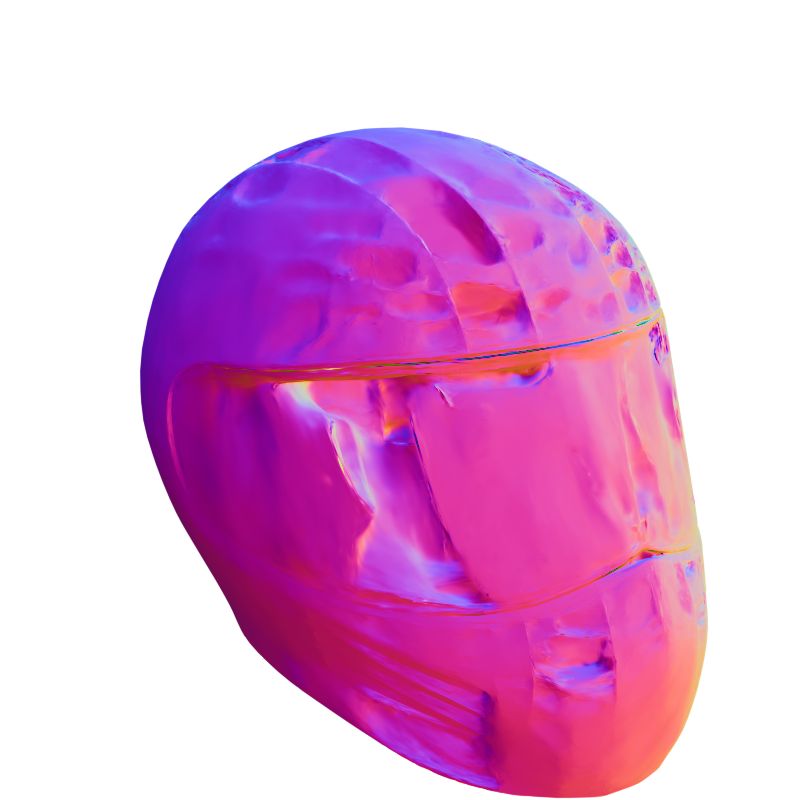}
                \\
                \rotatebox{90}{\scriptsize{~~~{TensoIR}}} &
                \includegraphics[width=0.18\linewidth,trim={2cm 6cm 2cm 1cm},clip]{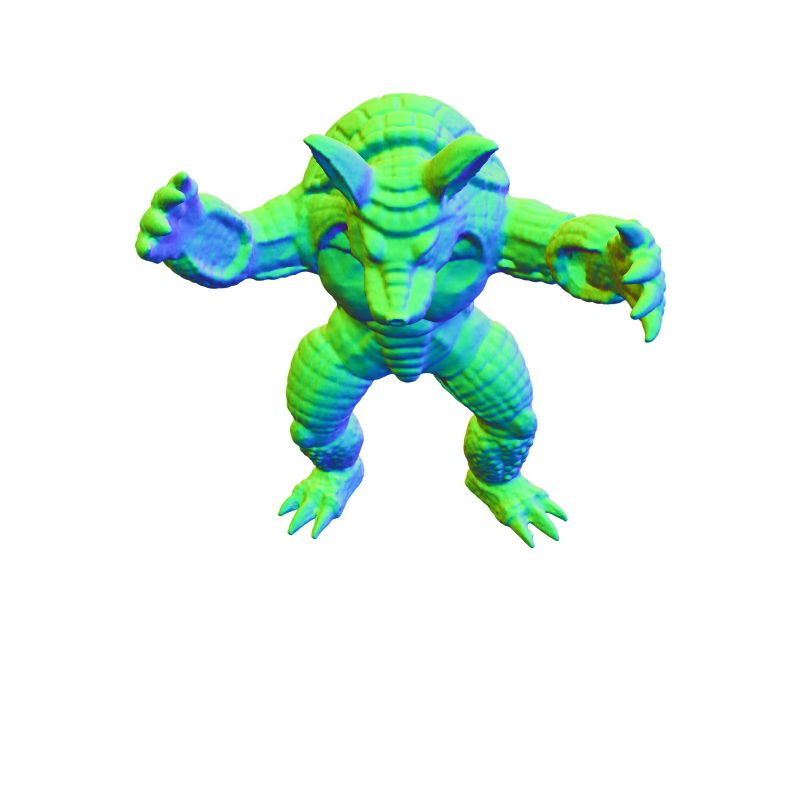}  &
                \includegraphics[width=0.18\linewidth,trim={1.5cm 1cm 0cm 4cm},clip]{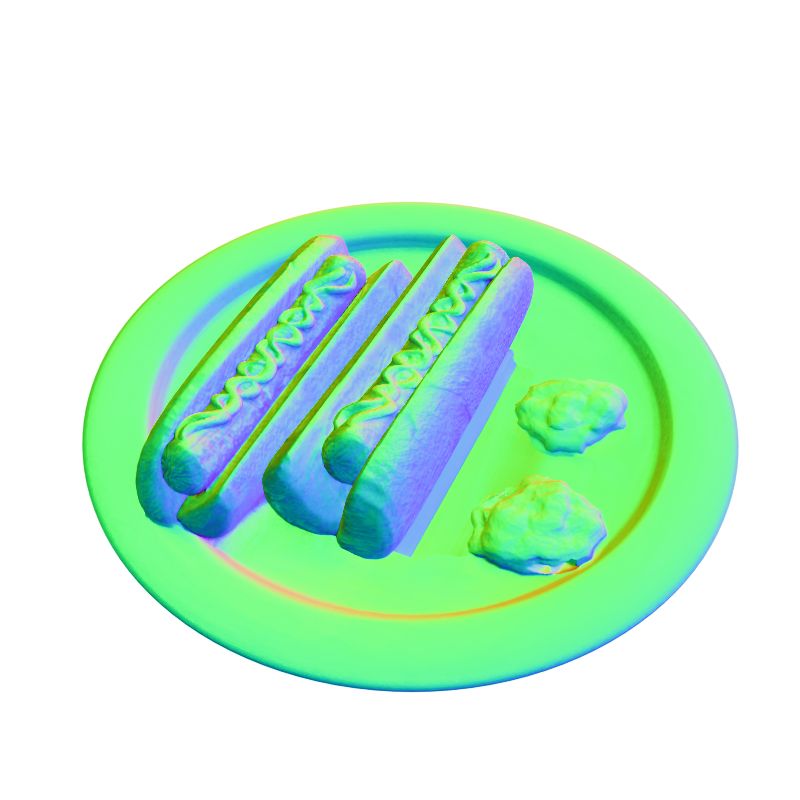}  &
                \includegraphics[width=0.18\linewidth,trim={0cm 2cm 0cm 0cm},clip]{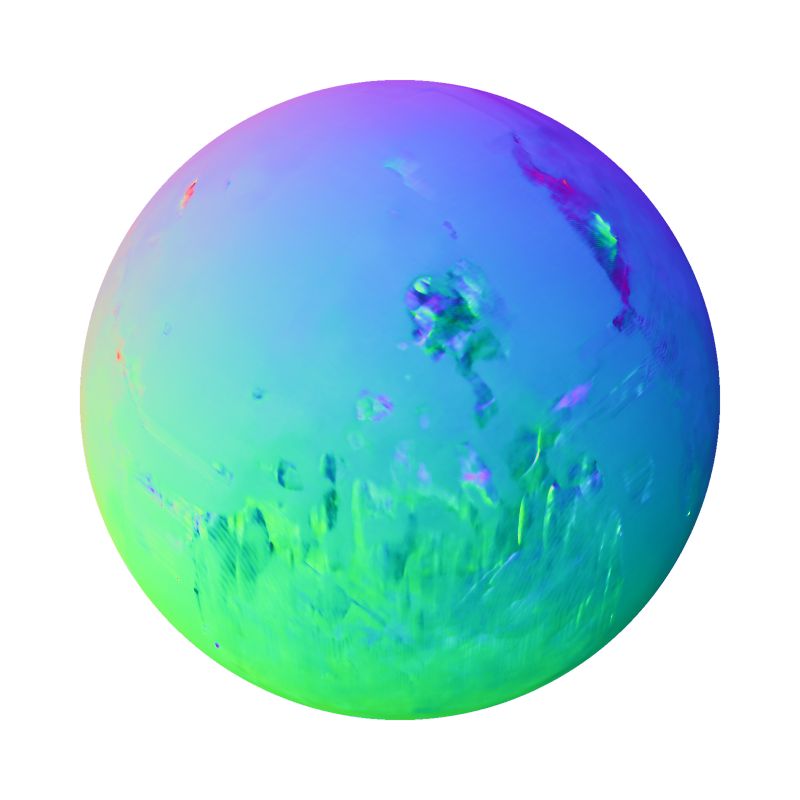}  &
                \includegraphics[width=0.22\linewidth,trim={5cm 9cm 5cm 5cm},clip]{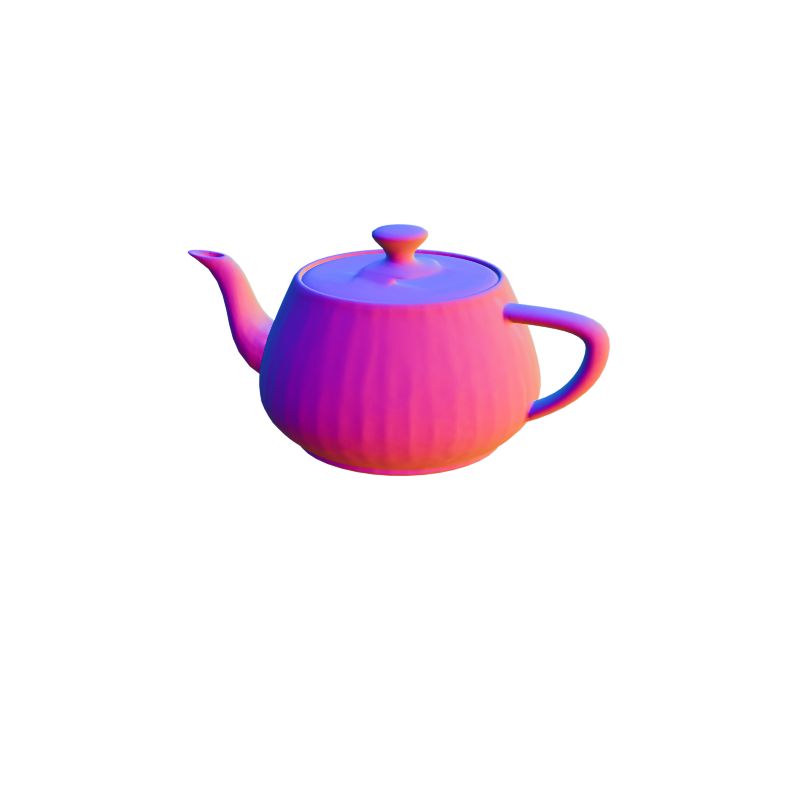}  &
                \includegraphics[width=0.16\linewidth,trim={2cm 0cm 0cm 4cm},clip]{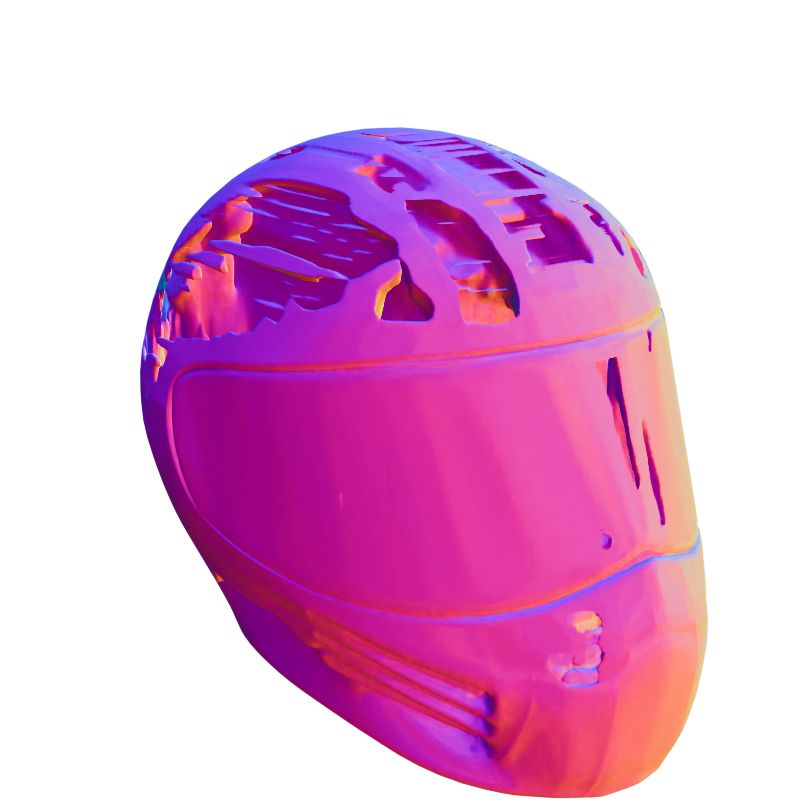}
                \\
                \rotatebox{90}{\scriptsize{~~~~~~{Ours}}} &
                \includegraphics[width=0.18\linewidth,trim={2cm 6cm 2cm 1cm},clip]{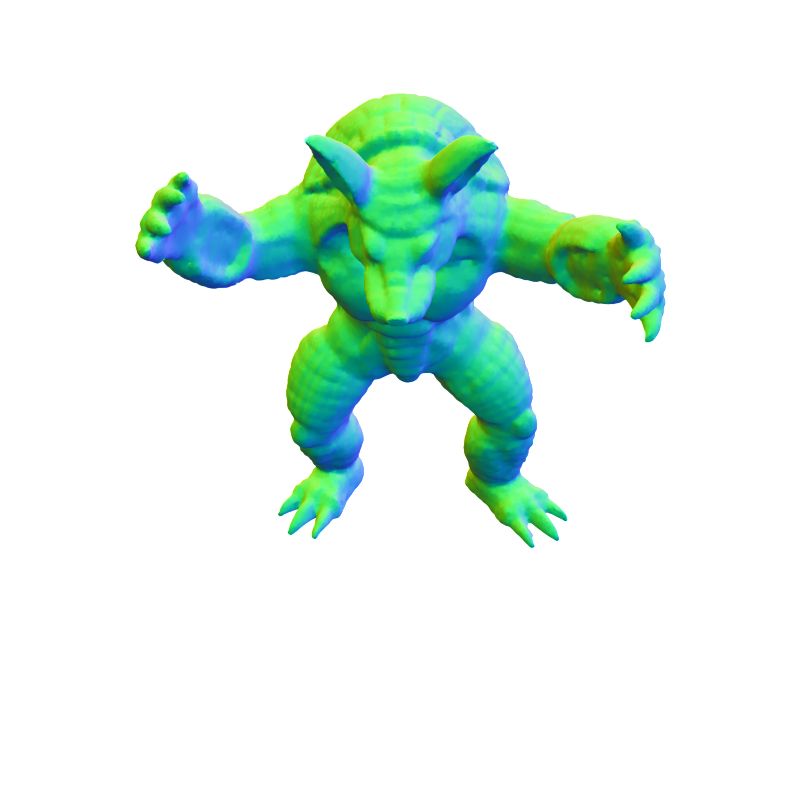}  &
                \includegraphics[width=0.18\linewidth,trim={1.5cm 1cm 0cm 4cm},clip]{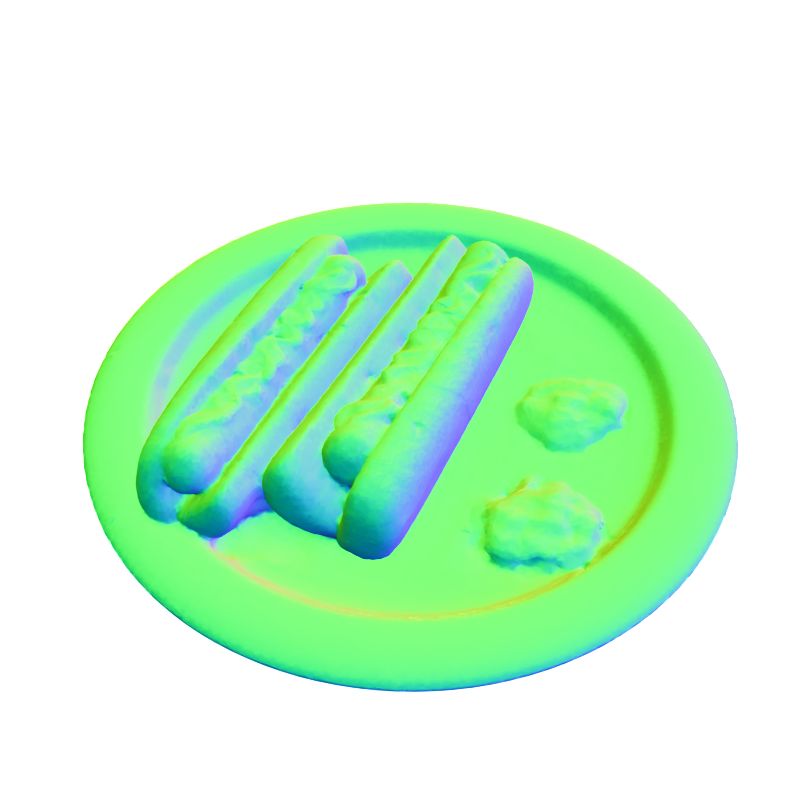}  &
                \includegraphics[width=0.18\linewidth,trim={0cm 2cm 0cm 0cm},clip]{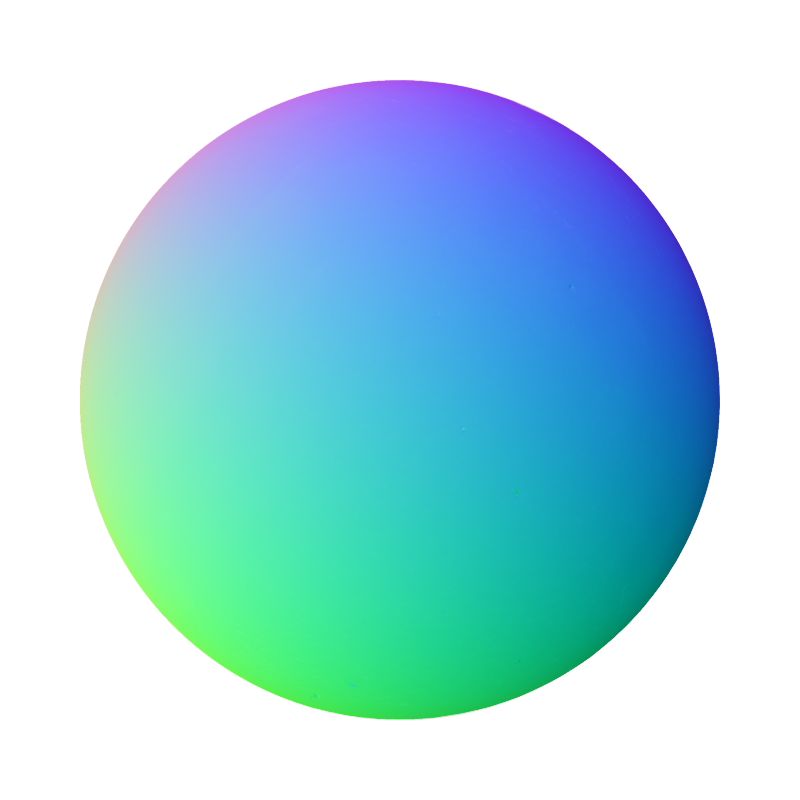}  &
                \includegraphics[width=0.22\linewidth,trim={5cm 9cm 5cm 5cm},clip]{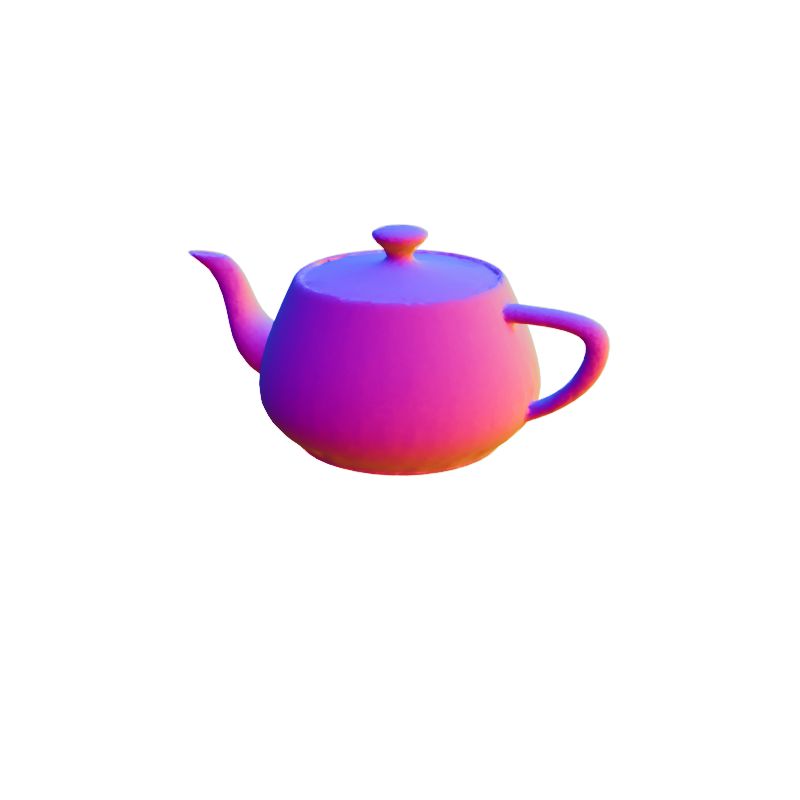}  &
                \includegraphics[width=0.16\linewidth,trim={2cm 0cm 0cm 4cm},clip]{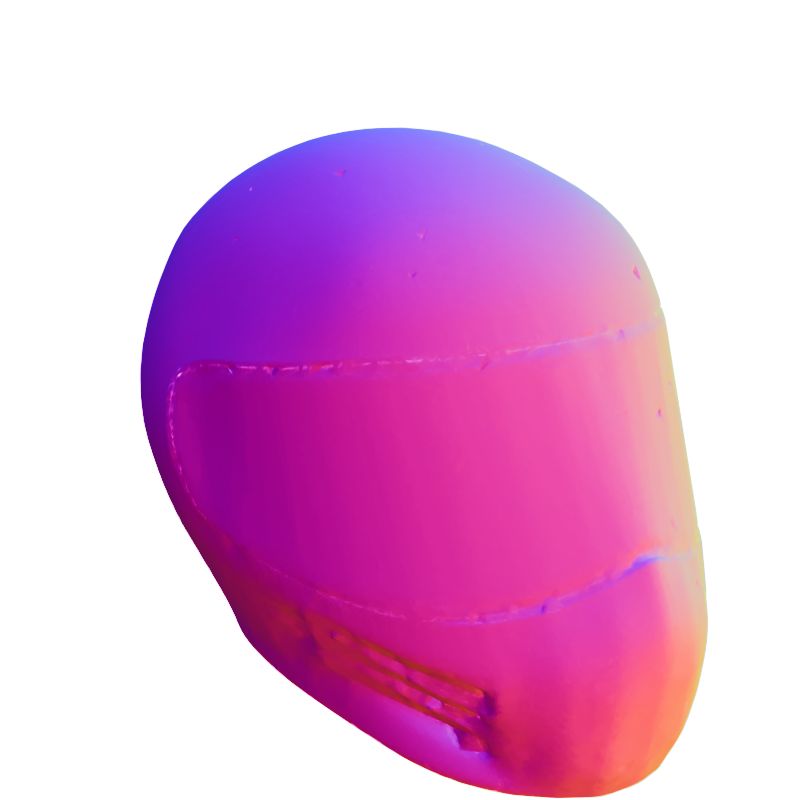}
                \\
                \rotatebox{90}{\scriptsize{~~{GT Normal}}} &
                \includegraphics[width=0.18\linewidth,trim={2cm 6cm 2cm 1cm},clip]{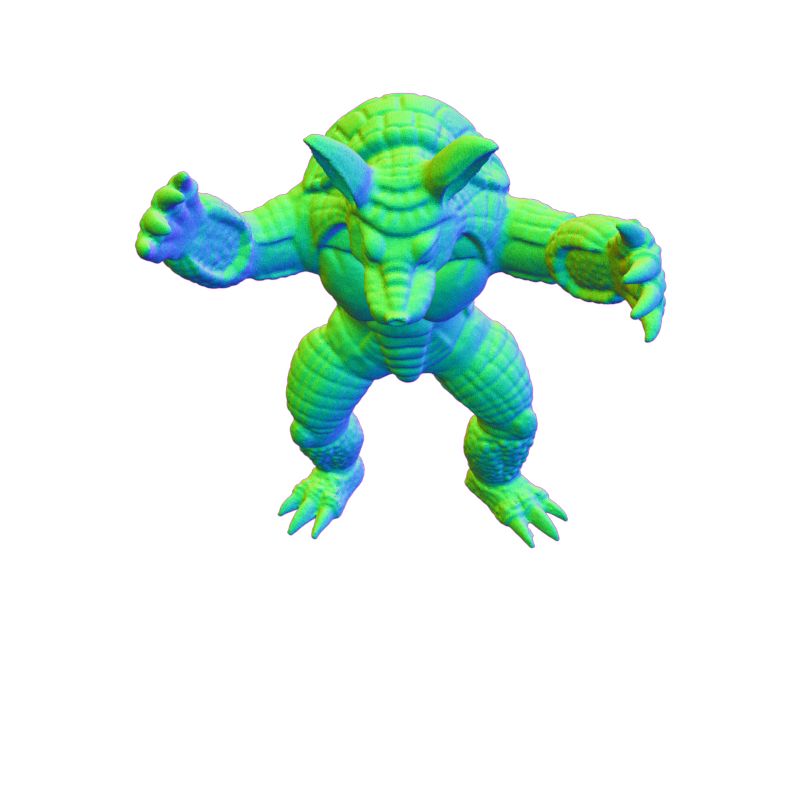}  &
                \includegraphics[width=0.18\linewidth,trim={1.5cm 1cm 0cm 4cm},clip]{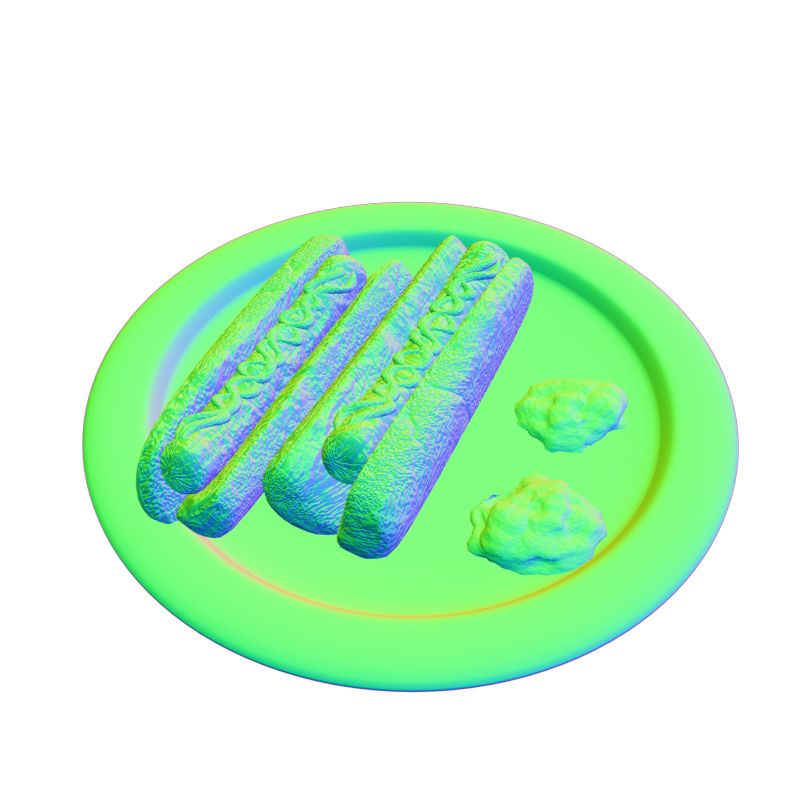}  &
                \includegraphics[width=0.18\linewidth,trim={0cm 2cm 0cm 0cm},clip]{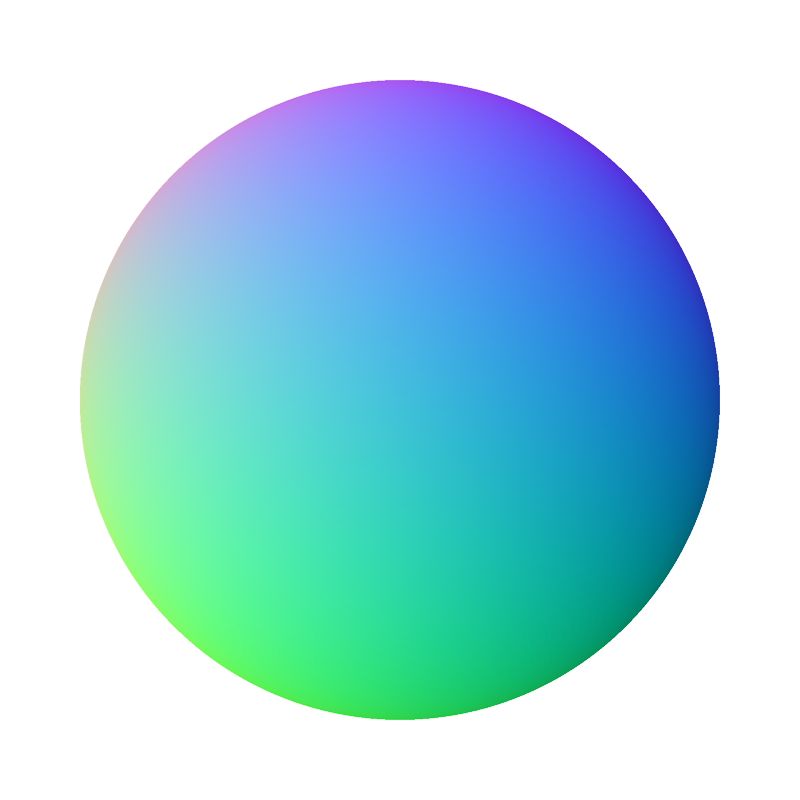}  &
                \includegraphics[width=0.22\linewidth,trim={5cm 9cm 5cm 5cm},clip]{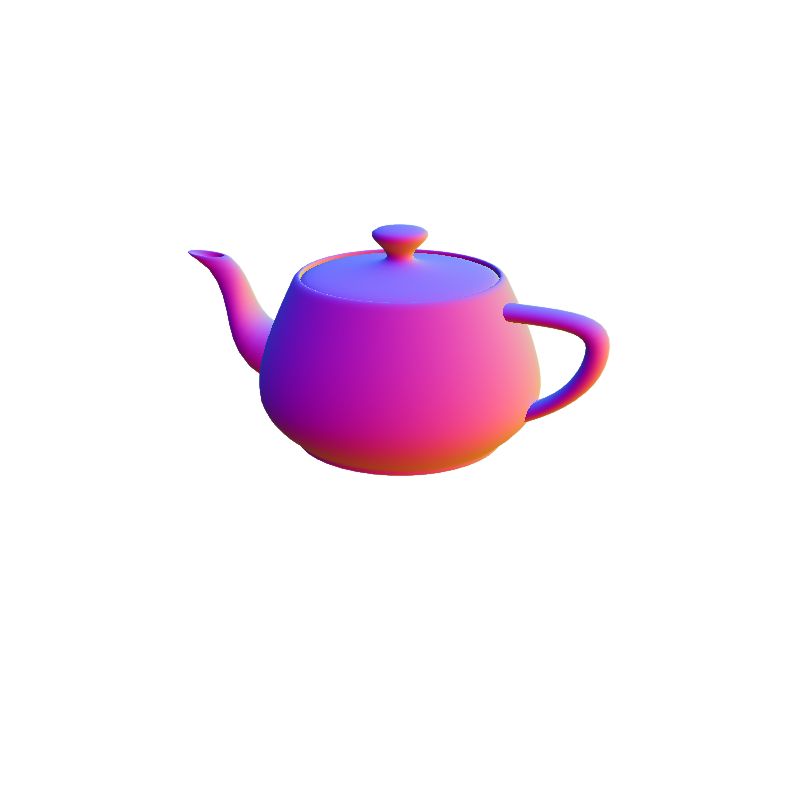}  &
                \includegraphics[width=0.16\linewidth,trim={2cm 0cm 0cm 4cm},clip]{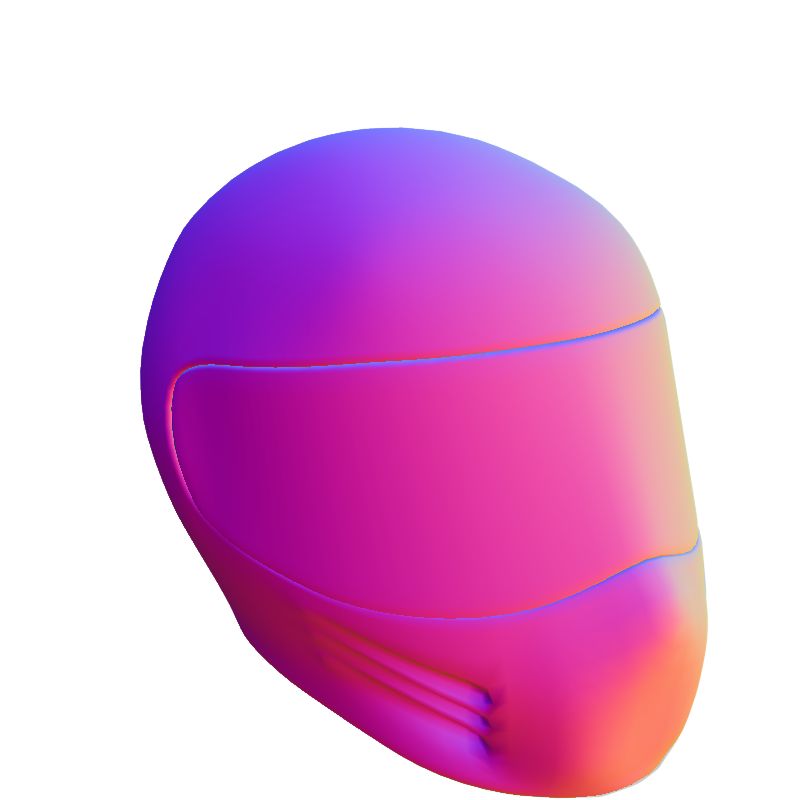}
                \\
                \rotatebox{90}{\scriptsize{~~{GT Render}}} &
                \includegraphics[width=0.18\linewidth,trim={2cm 6cm 2cm 1cm},clip]{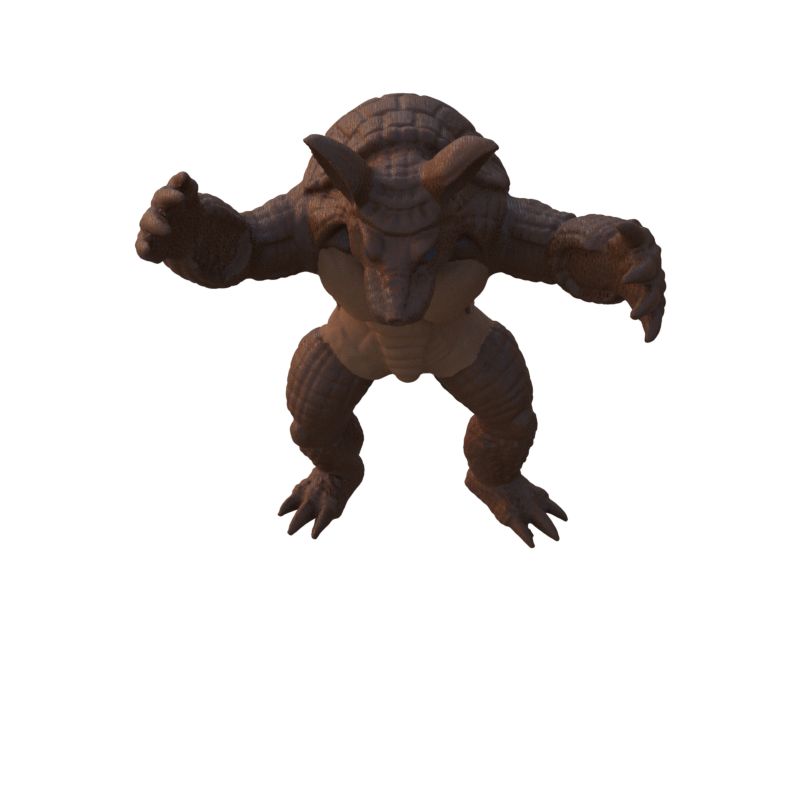}  &
                \includegraphics[width=0.18\linewidth,trim={1.5cm 1cm 0cm 4cm},clip]{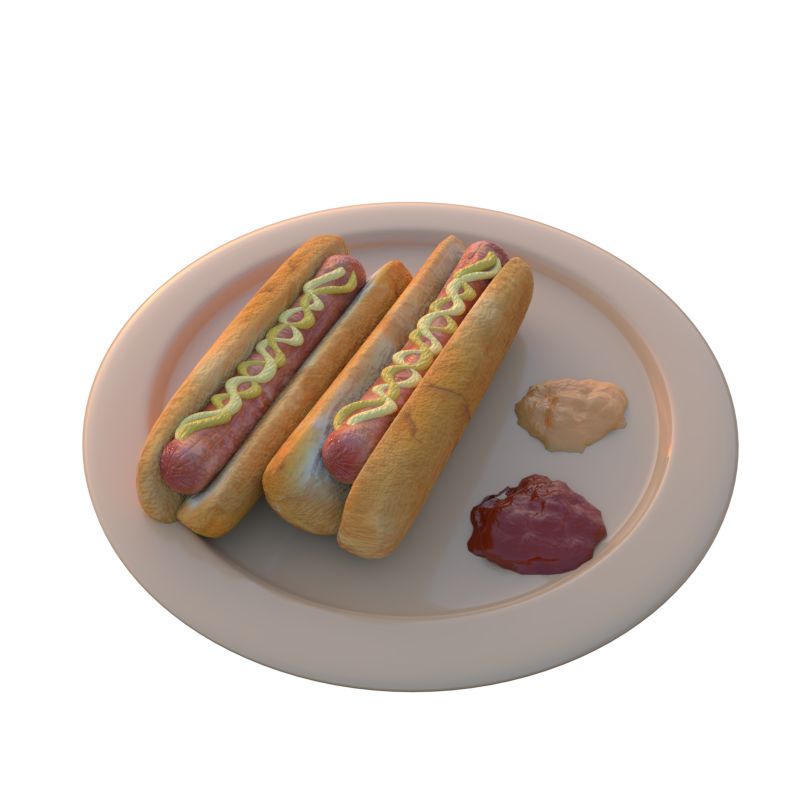}  &
                \includegraphics[width=0.18\linewidth,trim={0cm 2cm 0cm 0cm},clip]{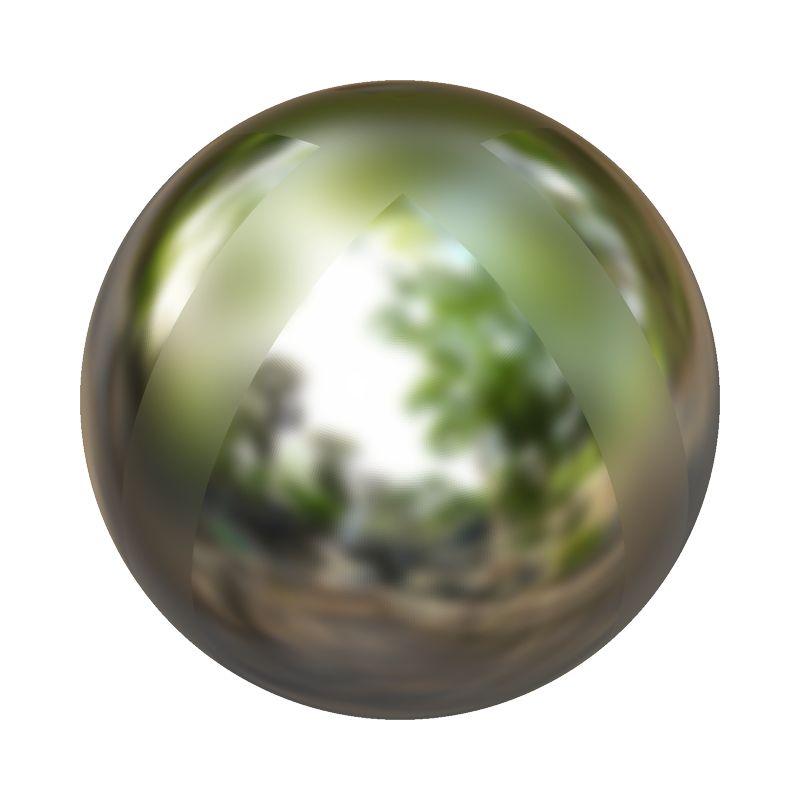}  &
                \includegraphics[width=0.22\linewidth,trim={5cm 9cm 5cm 5cm},clip]{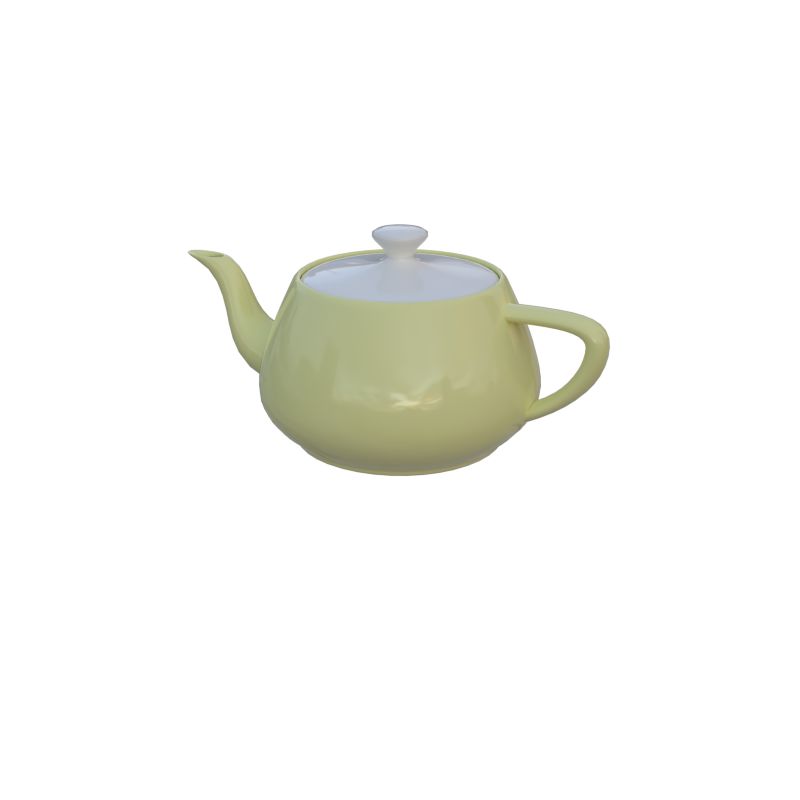}  &
                \includegraphics[width=0.16\linewidth,trim={2cm 0cm 0cm 4cm},clip]{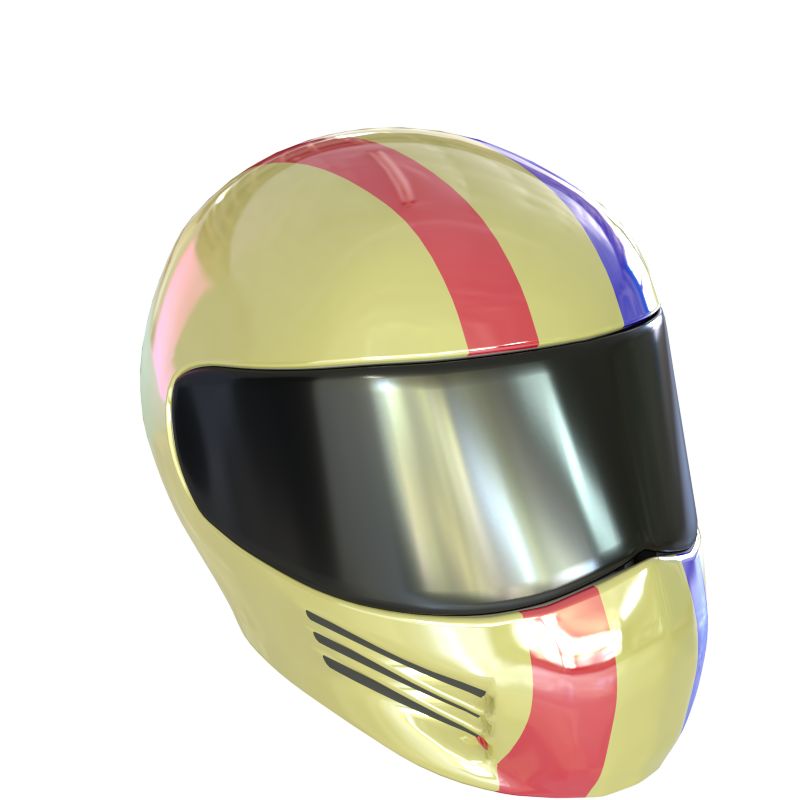}
            \end{tabular}
        \end{tabular}
    }
    \end{center}
    \vspace{-7mm}
    \caption{ 
        \textbf{Qualitative Comparison of Normal Estimation.}
        Our approach delivers excellent normal quality across various cases, including shadow effects, inter-reflection, and specular surfaces.
        \label{fig:exp:normal}
    }
    \vspace{0.6mm}
\end{figure}

\vspace{-1mm}
\paragraph{Performance \& Discussion.} 
As shown in Table~\ref{table:exp:normal} and Fig.~\ref{fig:exp:normal}, our method achieves significantly improved normal quality over 3DGS-based baselines which only deliver approximated normals, demonstrating the effectiveness of our geometry guidance. Compared to approaches based on implicit-field, which are carefully optimized for several hours, our method also achieves competitive normal quality, while delivering exceptional training efficiency.

\begin{table}[H]
\vspace{-9mm}
\begin{center}
    \resizebox{1.0\linewidth}{!}{ % Resizes the table to fit within the text width
    \begin{tabular}{cr|ccc}
    \toprule
    & Method & Training Time & TensoIR Synthetic & Shiny Blender\\
        && minutes $\downarrow$& MAE $\downarrow$ & MAE $\downarrow$ \\
    \midrule
    Mesh-based & NVdiffrec & 72 & 4.97 & 9.38 \\
    & NVdiffrecmc & 82 & 4.81\cellcolor{best3} & 9.76 \\
    \midrule
    Implicit Field & TensoIR & $\sim$270 & 4.10\cellcolor{best2} & 4.42\cellcolor{best3} \\
    & NeRO & $\sim$800 & 5.14 & 2.48\cellcolor{best2} \\
    \midrule
    3DGS-based & GS-IR & 20\cellcolor{best2} & 5.41 & 4.42\cellcolor{best3} \\
    & {\gaussianshader} & 63\cellcolor{best3} & 6.79 & 7.03 \\
     & {\yaoyao} & $\sim$110 & 5.45 & 7.04 \\
     & Ours & 14\cellcolor{best1} & 4.08\cellcolor{best1} & 2.15\cellcolor{best1} \\
    \bottomrule
\end{tabular}
    }
    \end{center}
\vspace{-5mm}
\caption{
\label{table:exp:normal} 
\textbf{Quantitative Results of Normal Estimation.} Our method achieves state-of-the-art normal quality while delivering exceptional training efficiency.
}
\vspace{0mm}
\end{table}

\subsection{Real-World Results}
\label{exp:real}
\begin{figure}[H]
\vspace{-3mm}
    \begin{center}
    \resizebox{\linewidth}{!}{
        \setlength{\tabcolsep}{1pt}
        \setlength{\fboxrule}{1pt}
        \hspace{-4mm}
        \begin{tabular}{c}
            \begin{tabular}{ccccccc}
                \rotatebox{90}{~~~~~~\scriptsize{Car}} &
                \vspace{-0.5mm}
                \includegraphics[width=0.16\linewidth,trim={9cm 7cm 10cm 11cm},clip]{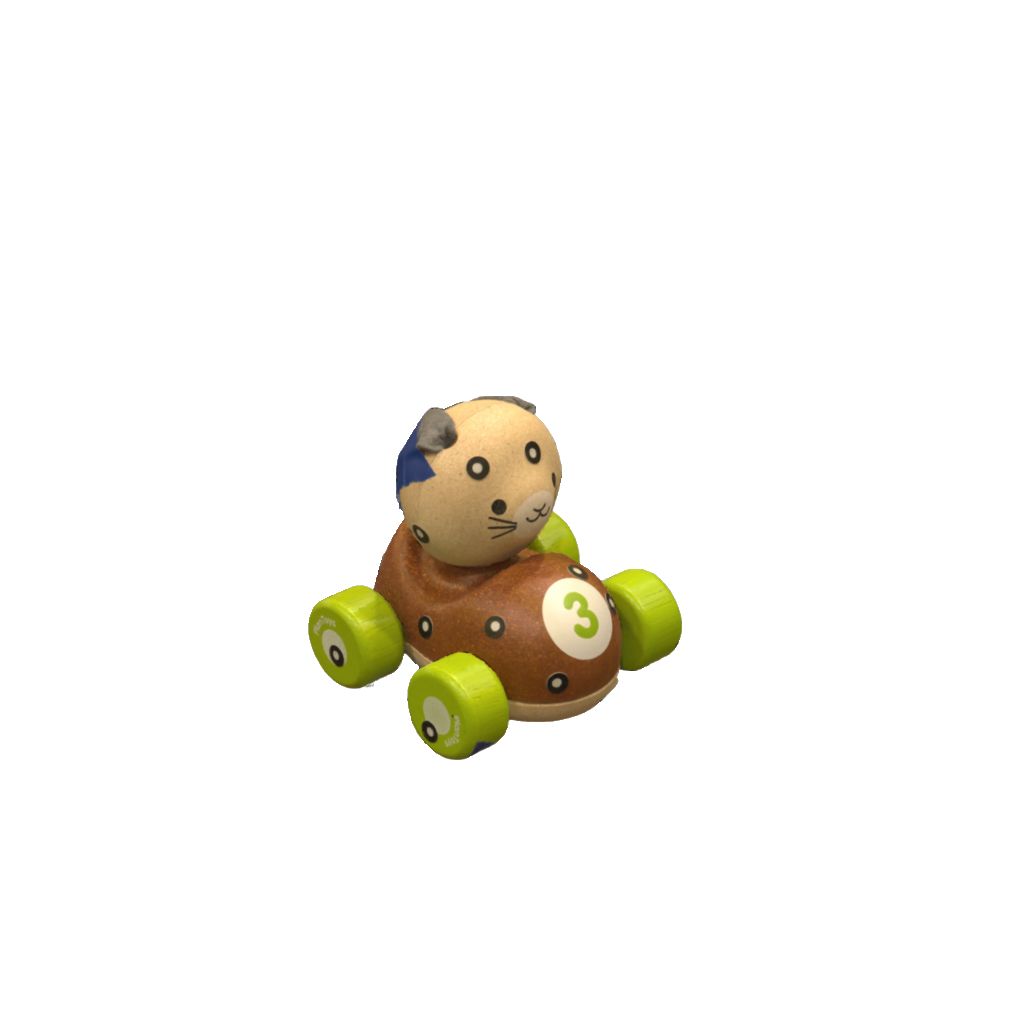}  &
                \includegraphics[width=0.16\linewidth,trim={9cm 7cm 10cm 11cm},clip]{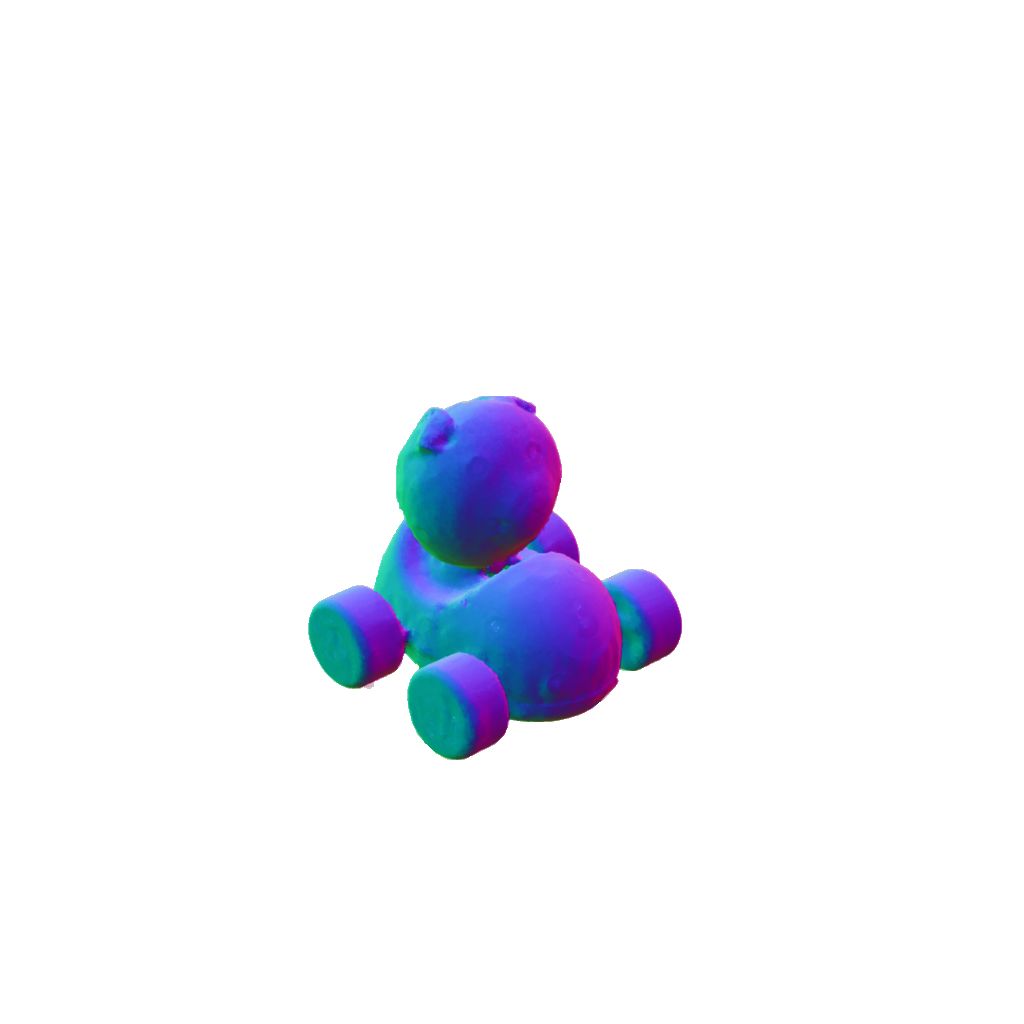}  &
                \includegraphics[width=0.16\linewidth,trim={9cm 7cm 10cm 11cm},clip]{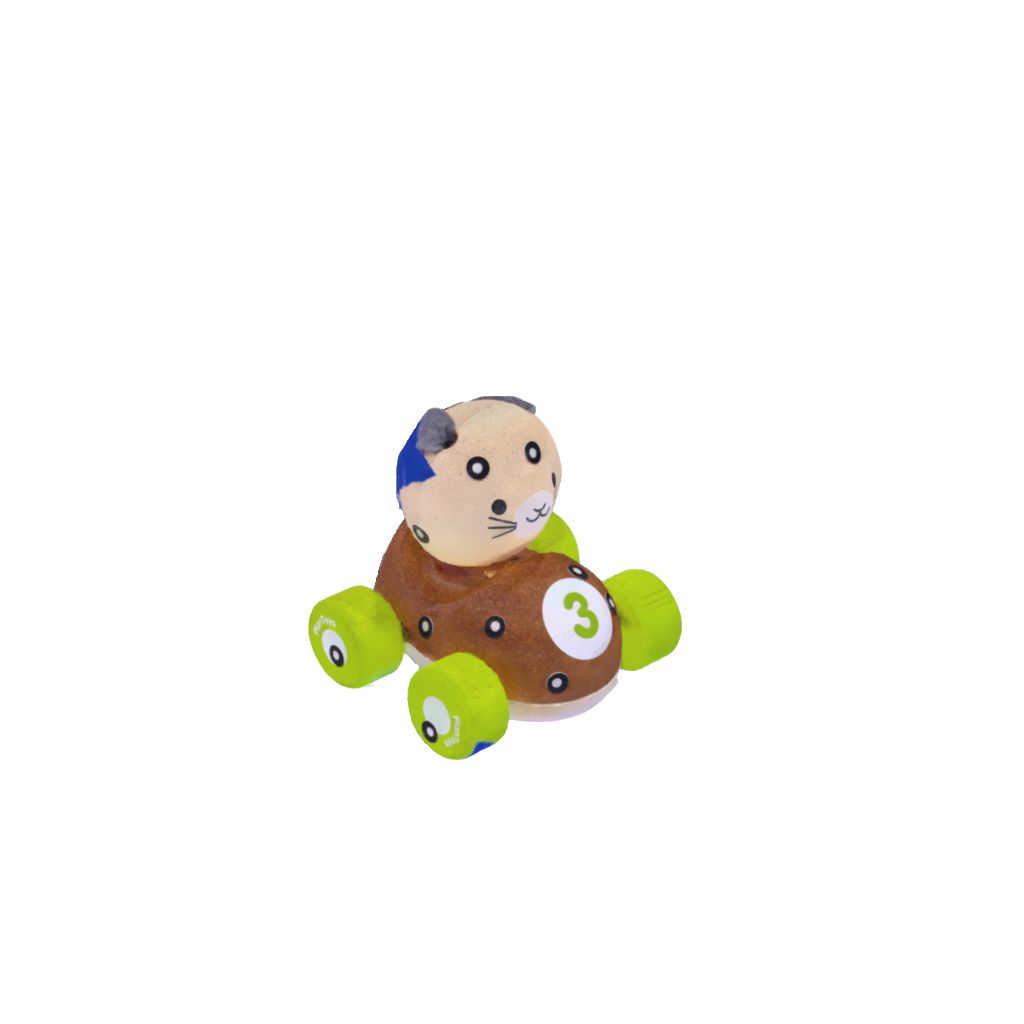}  &
                \includegraphics[width=0.16\linewidth]{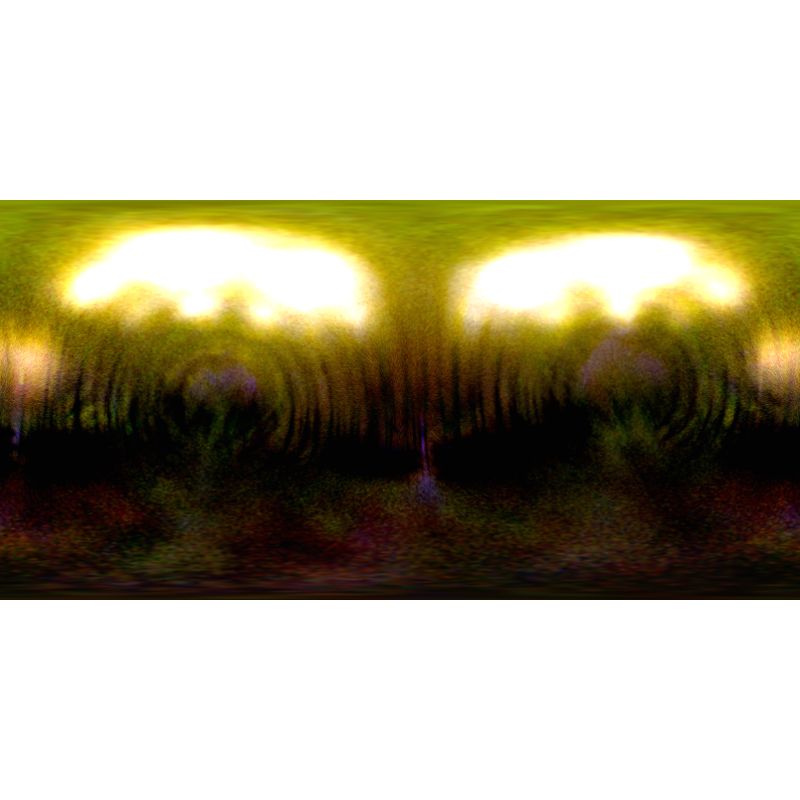}  &
                \includegraphics[width=0.16\linewidth,trim={9cm 7cm 10cm 11cm},clip]{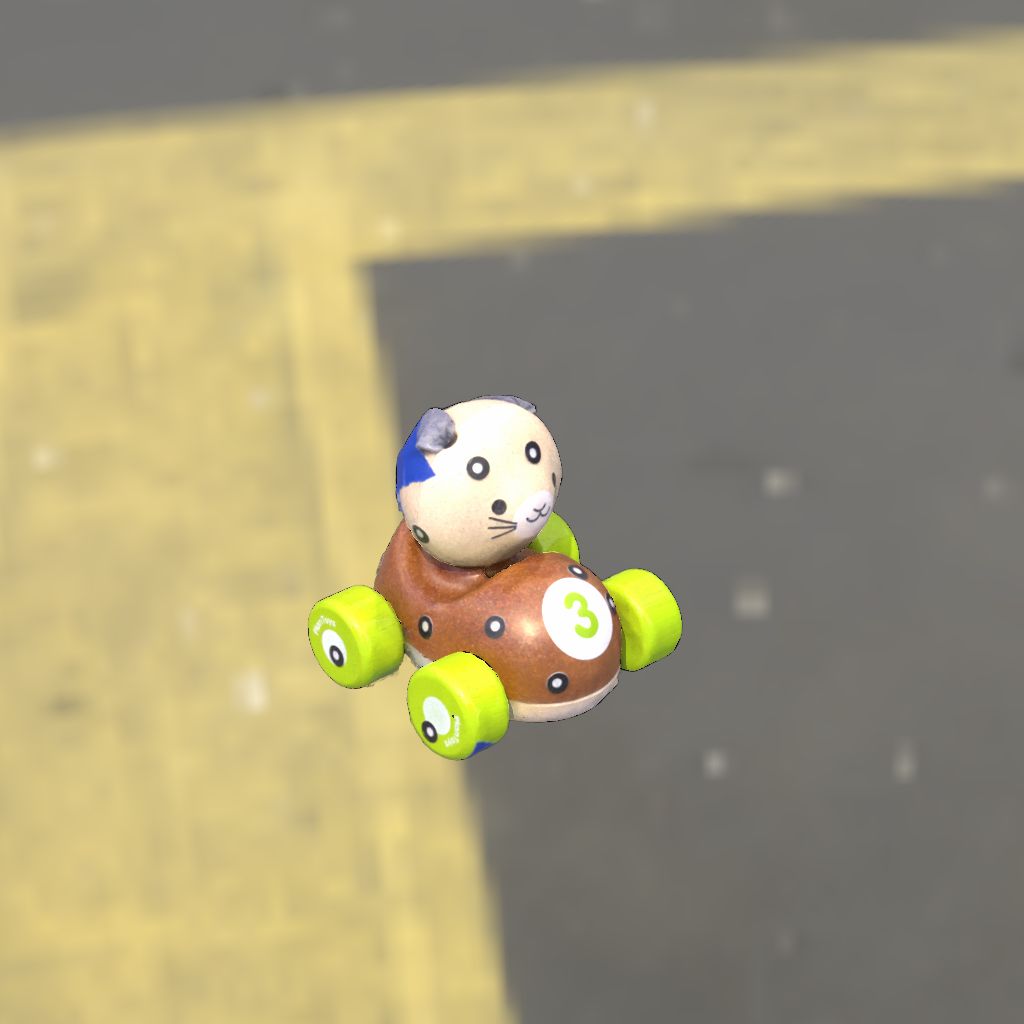}  &
                \includegraphics[width=0.16\linewidth,trim={9cm 7cm 10cm 11cm},clip]{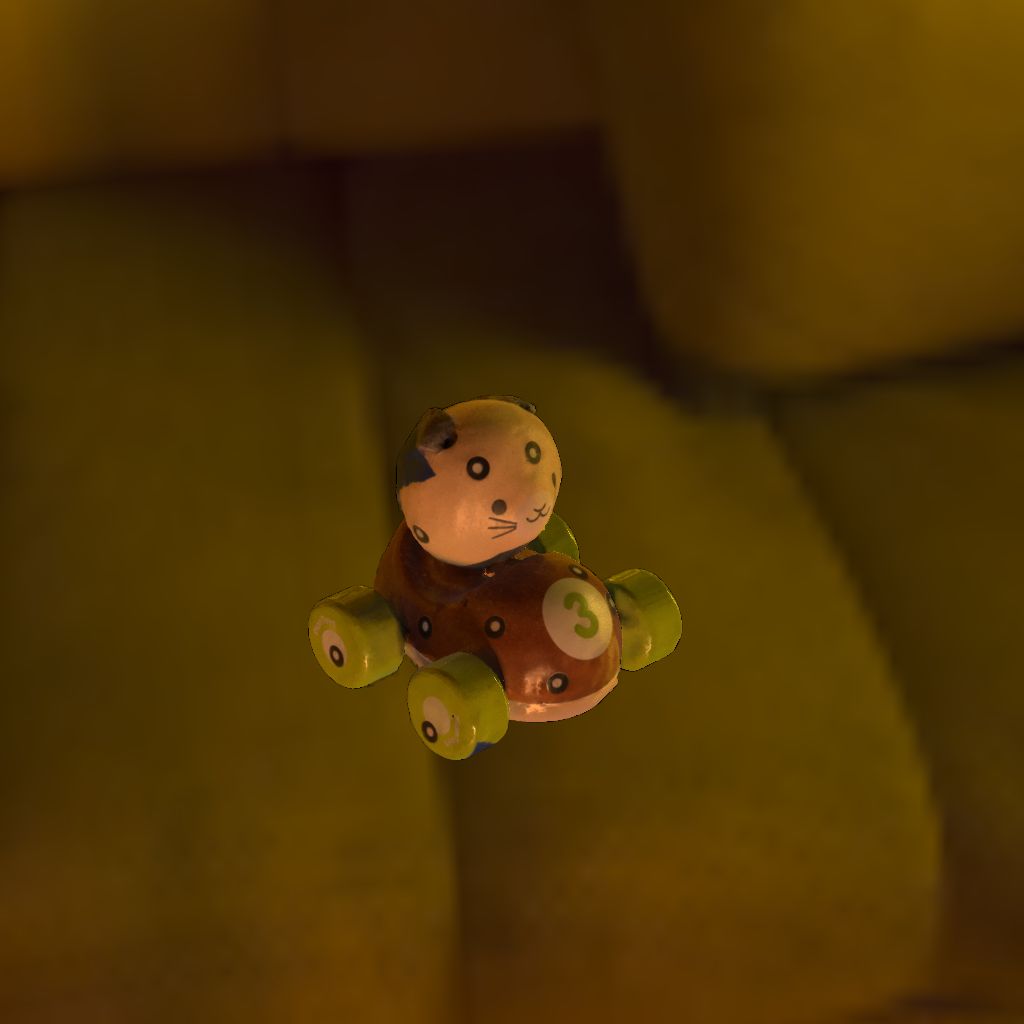}  \\
                \rotatebox{90}{~~~~\scriptsize{Teapot}} &
                \vspace{-0.5mm}
                \includegraphics[width=0.16\linewidth,trim={6cm 8cm 9cm 7cm},clip]{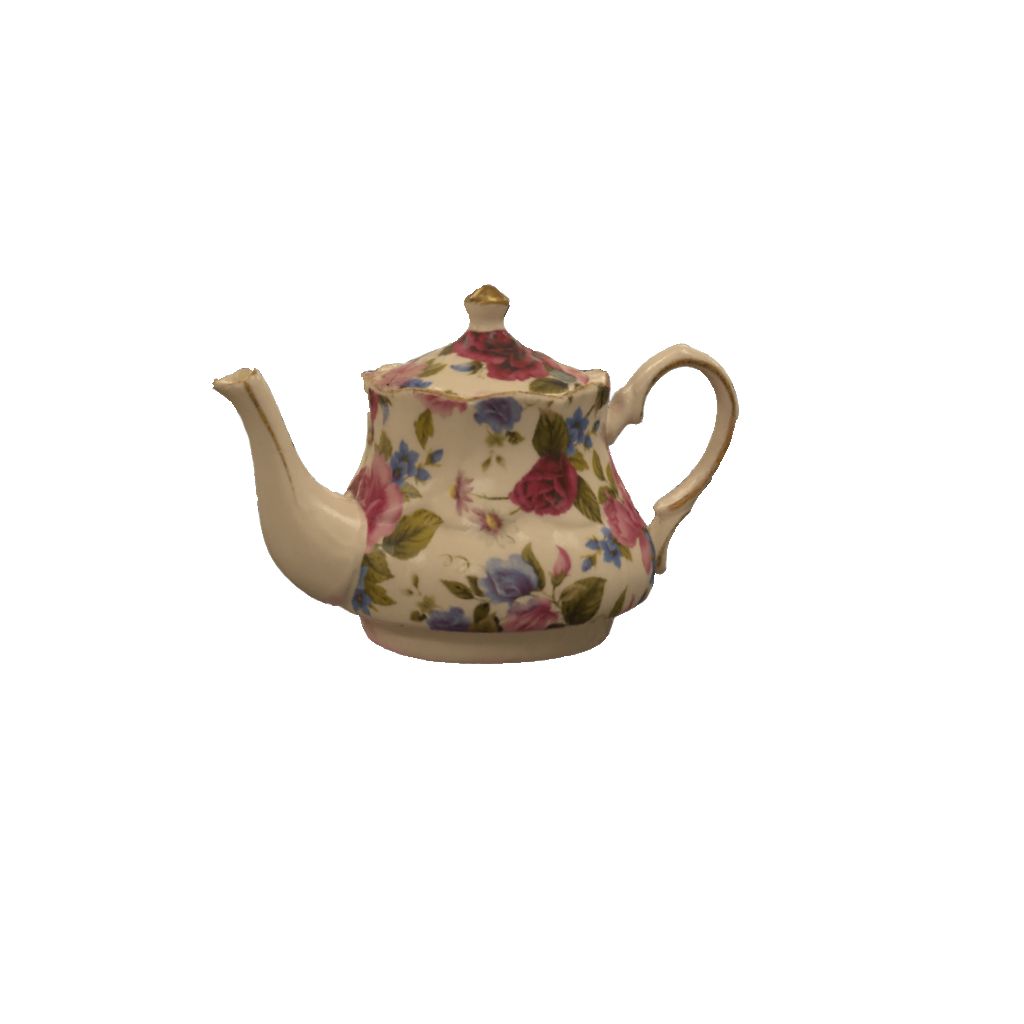}  &
                \includegraphics[width=0.16\linewidth,trim={6cm 8cm 9cm 7cm},clip]{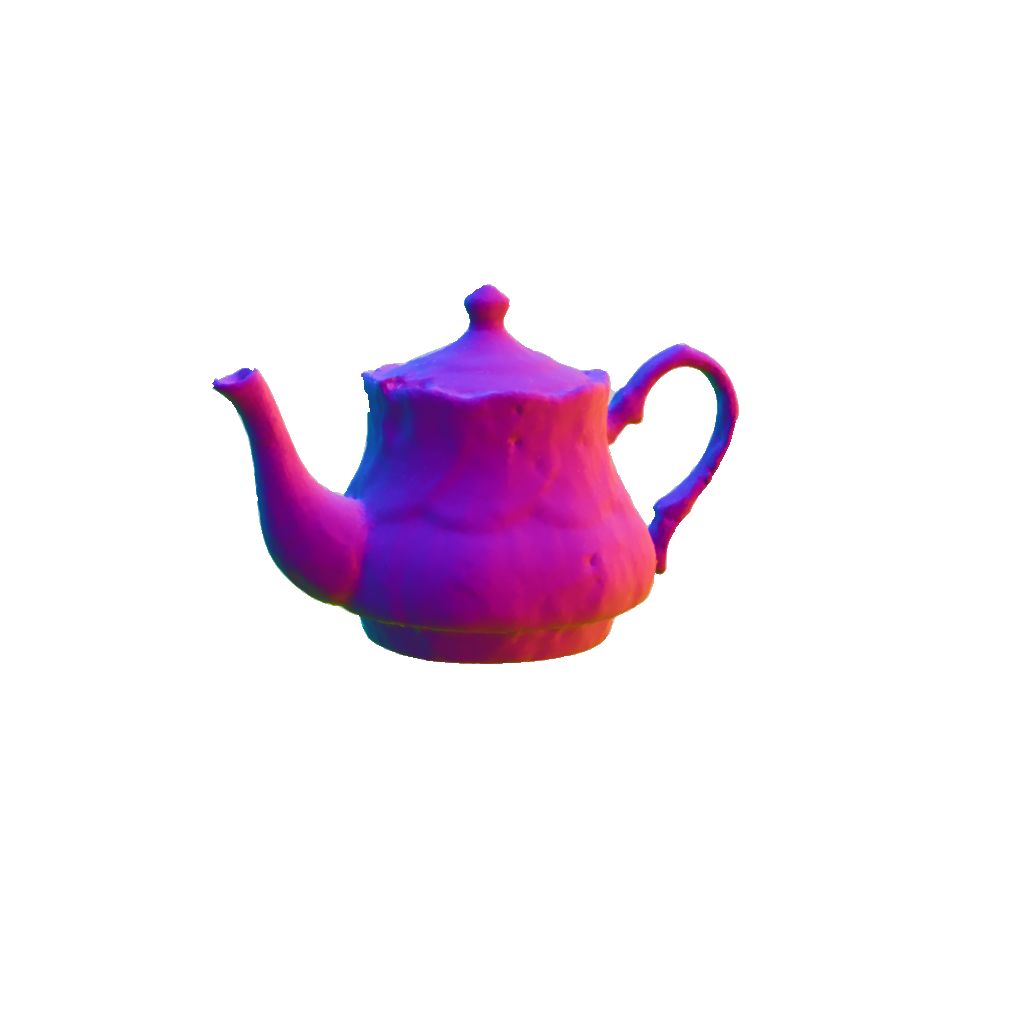}  &
                \includegraphics[width=0.16\linewidth,trim={6cm 8cm 9cm 7cm},clip]{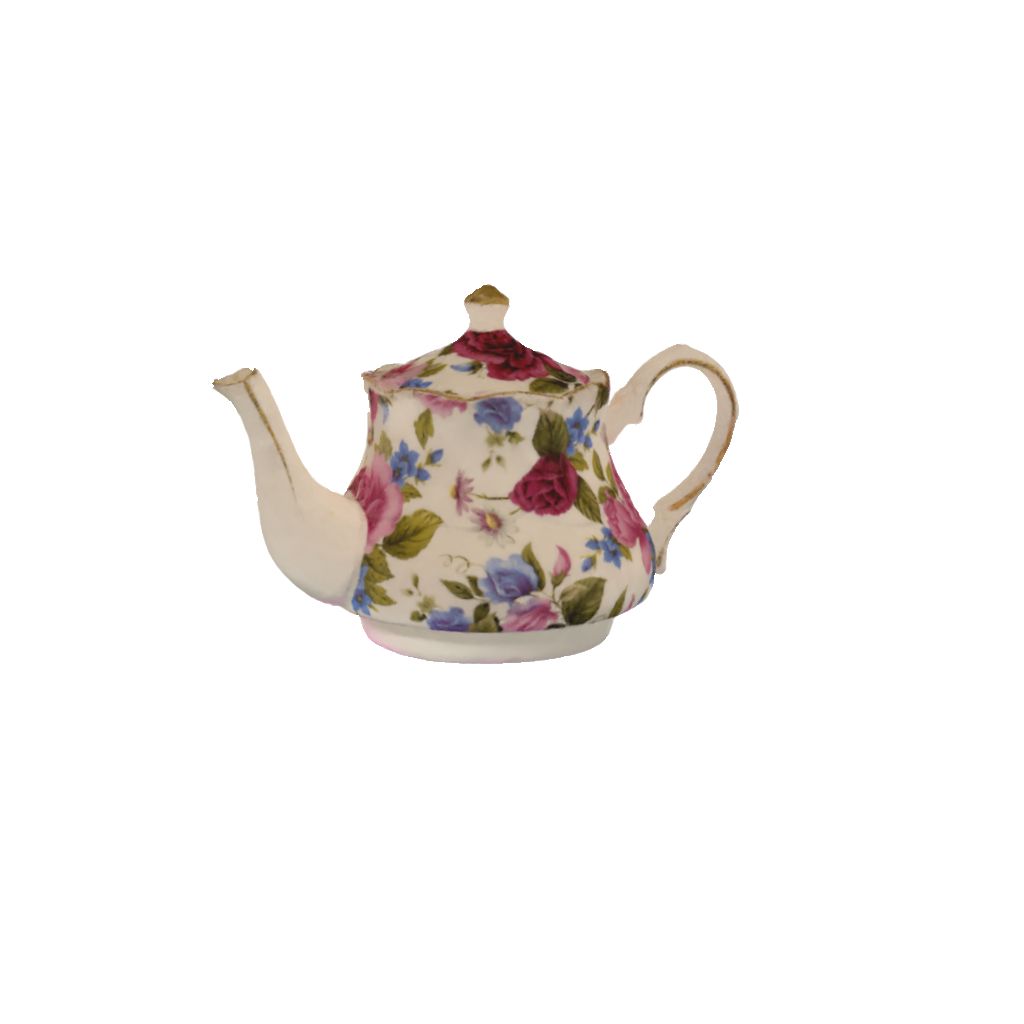}  &
                \includegraphics[width=0.16\linewidth]{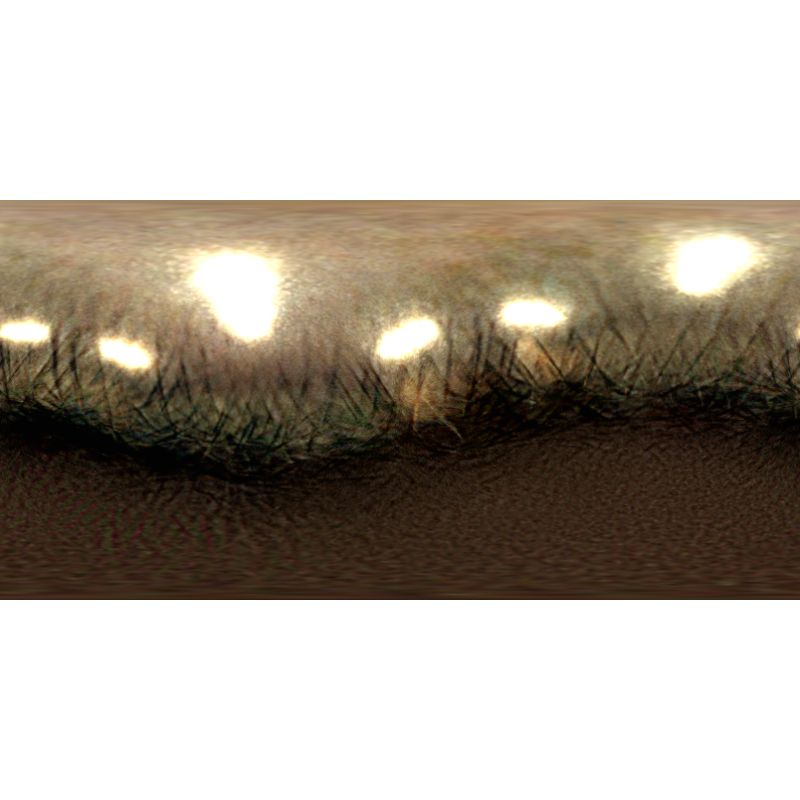}  &
                \includegraphics[width=0.16\linewidth,trim={6cm 8cm 9cm 7cm},clip]{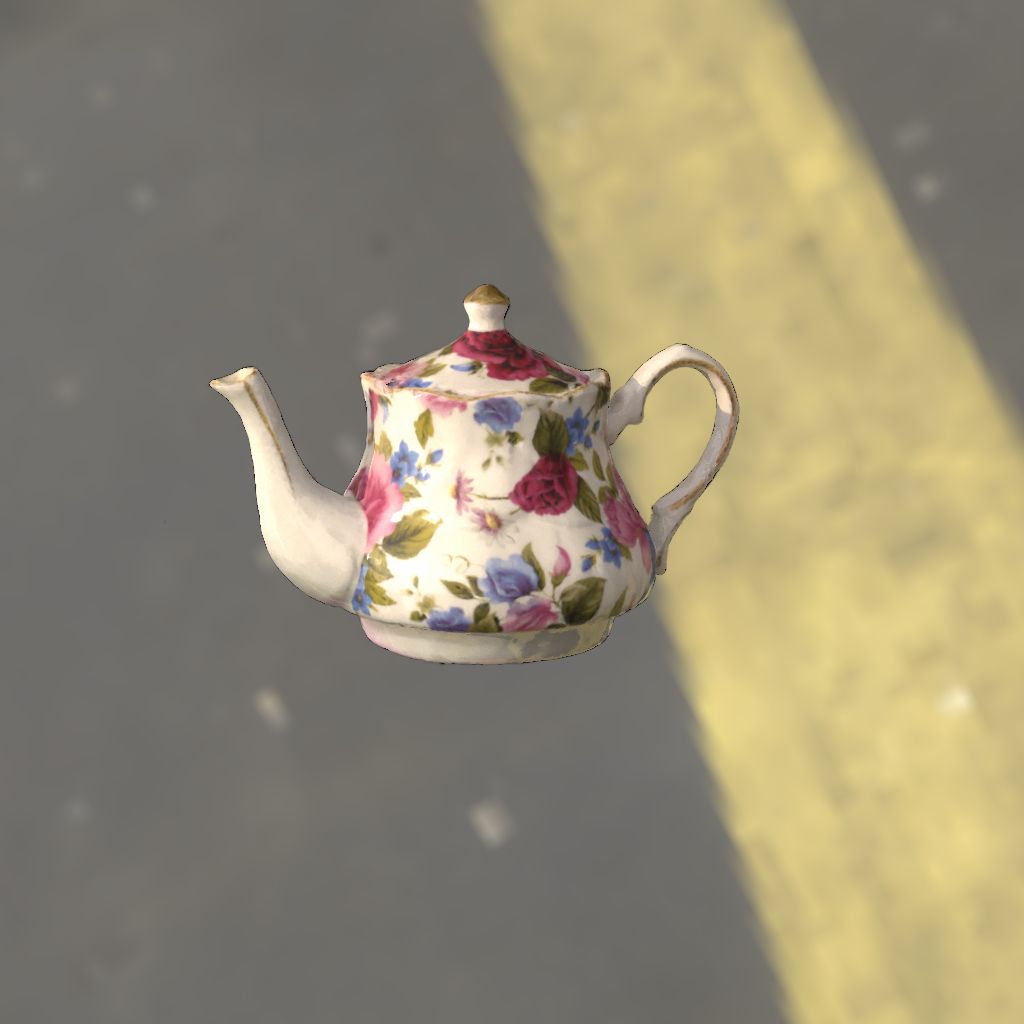}  &
                \includegraphics[width=0.16\linewidth,trim={6cm 8cm 9cm 7cm},clip]{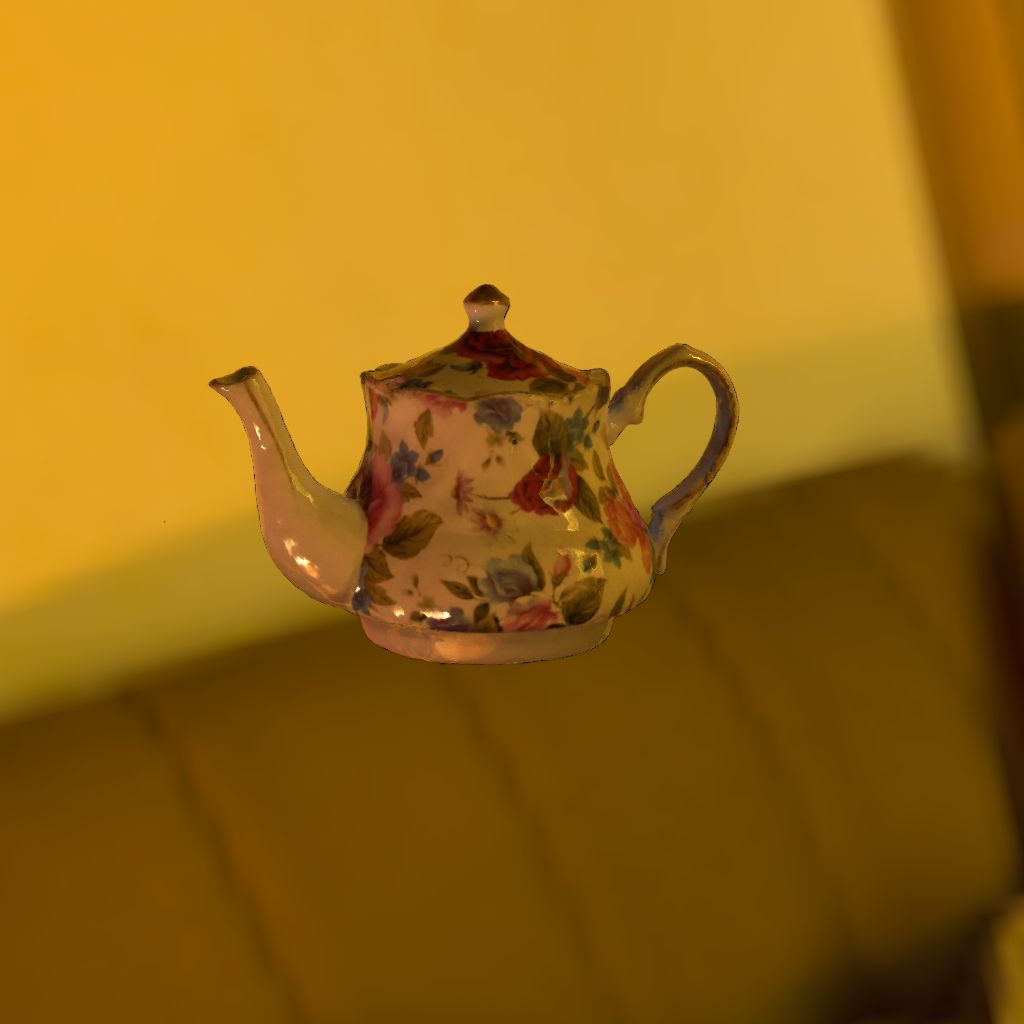}\\
                \rotatebox{90}{~~~~\scriptsize{Gnome}} &
                \vspace{-0.5mm}
                \includegraphics[width=0.16\linewidth,trim={10cm 4cm 4cm 9cm},clip]{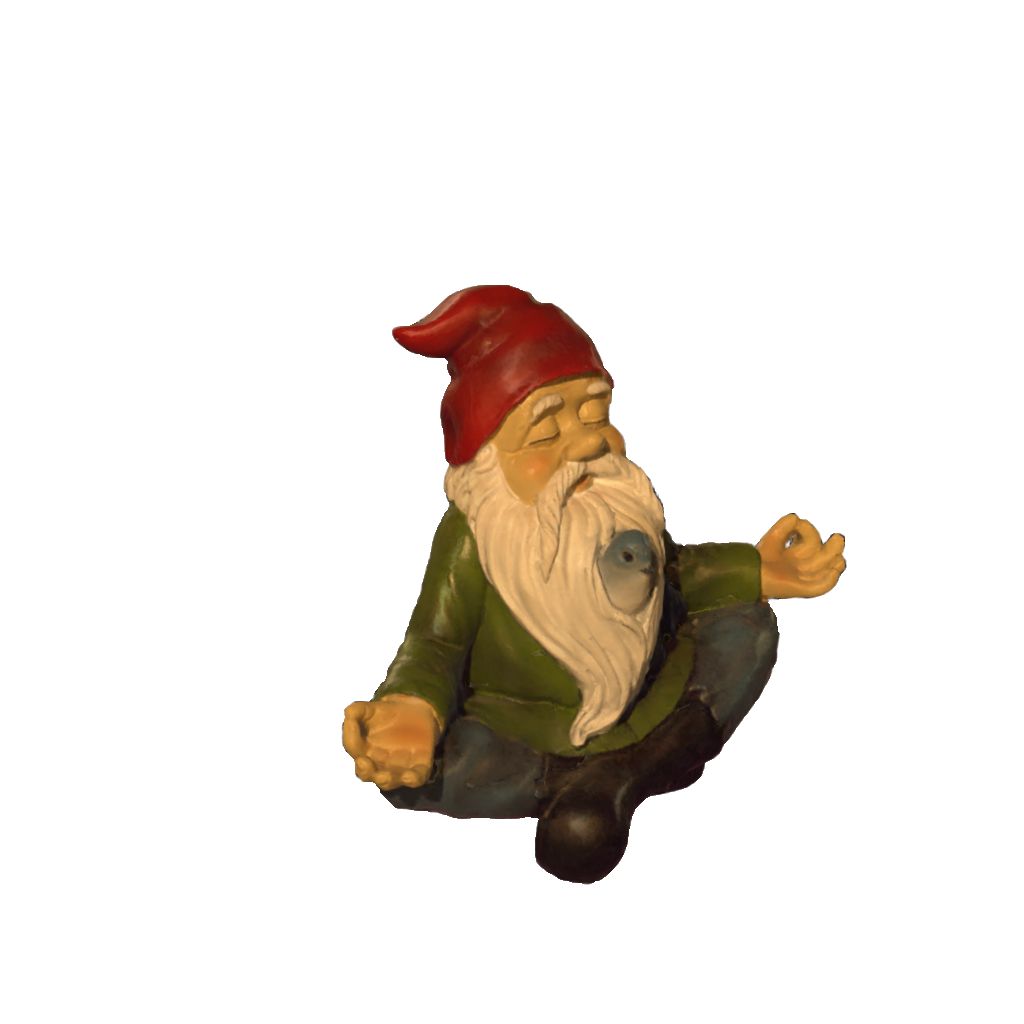}  &
                \includegraphics[width=0.16\linewidth,trim={10cm 4cm 4cm 9cm},clip]{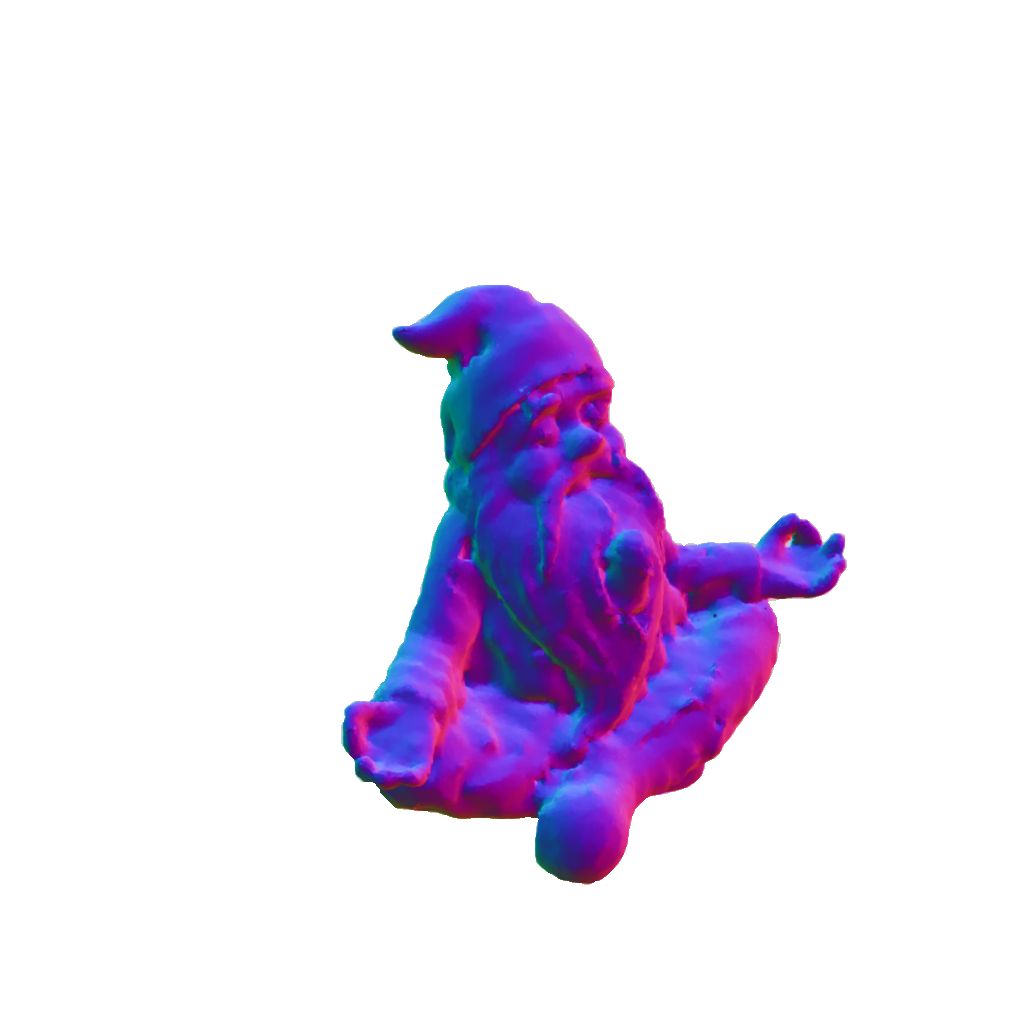}  &
                \includegraphics[width=0.16\linewidth,trim={10cm 4cm 4cm 9cm},clip]{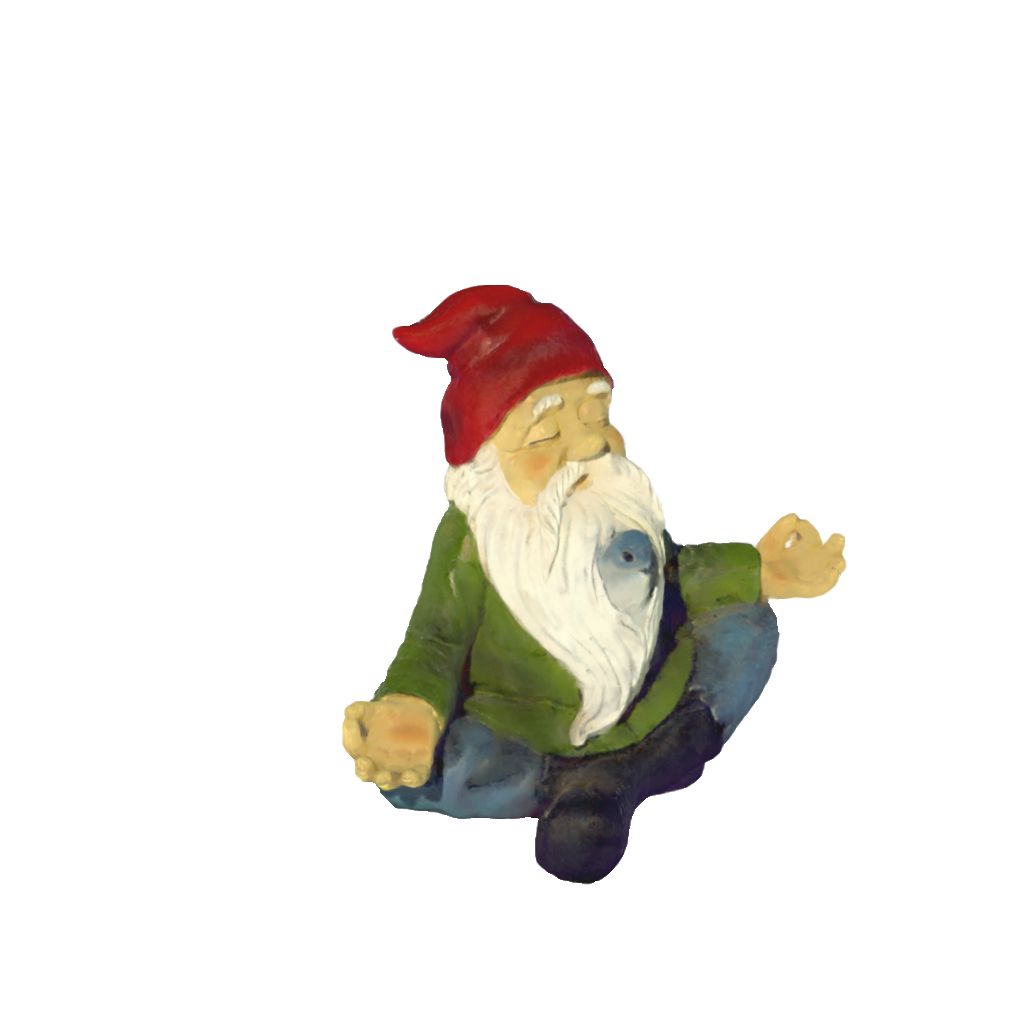}  &
                \includegraphics[width=0.16\linewidth]{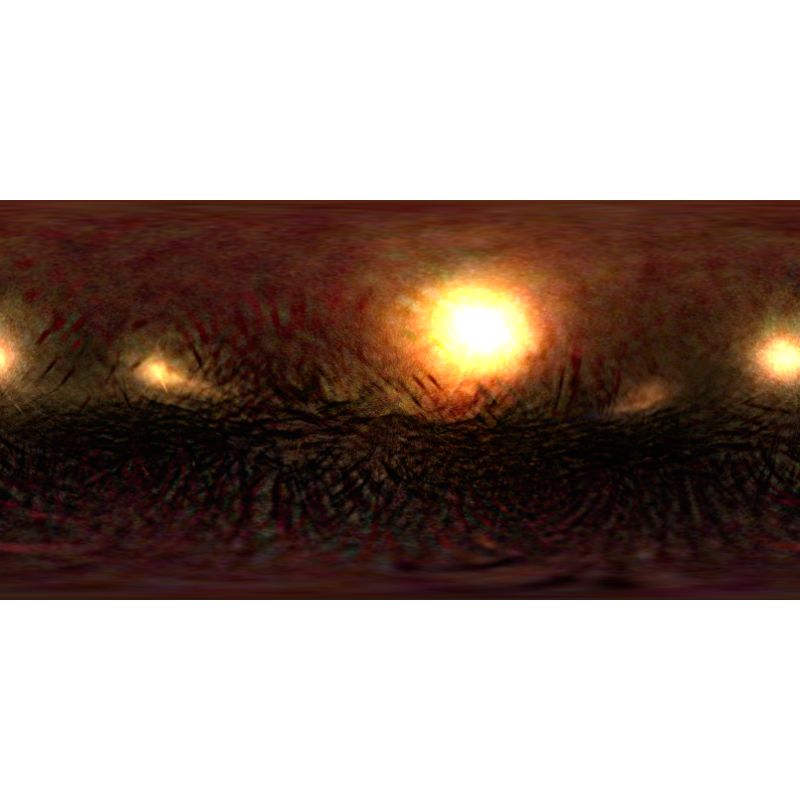}  &
                \includegraphics[width=0.16\linewidth,trim={10cm 4cm 4cm 9cm},clip]{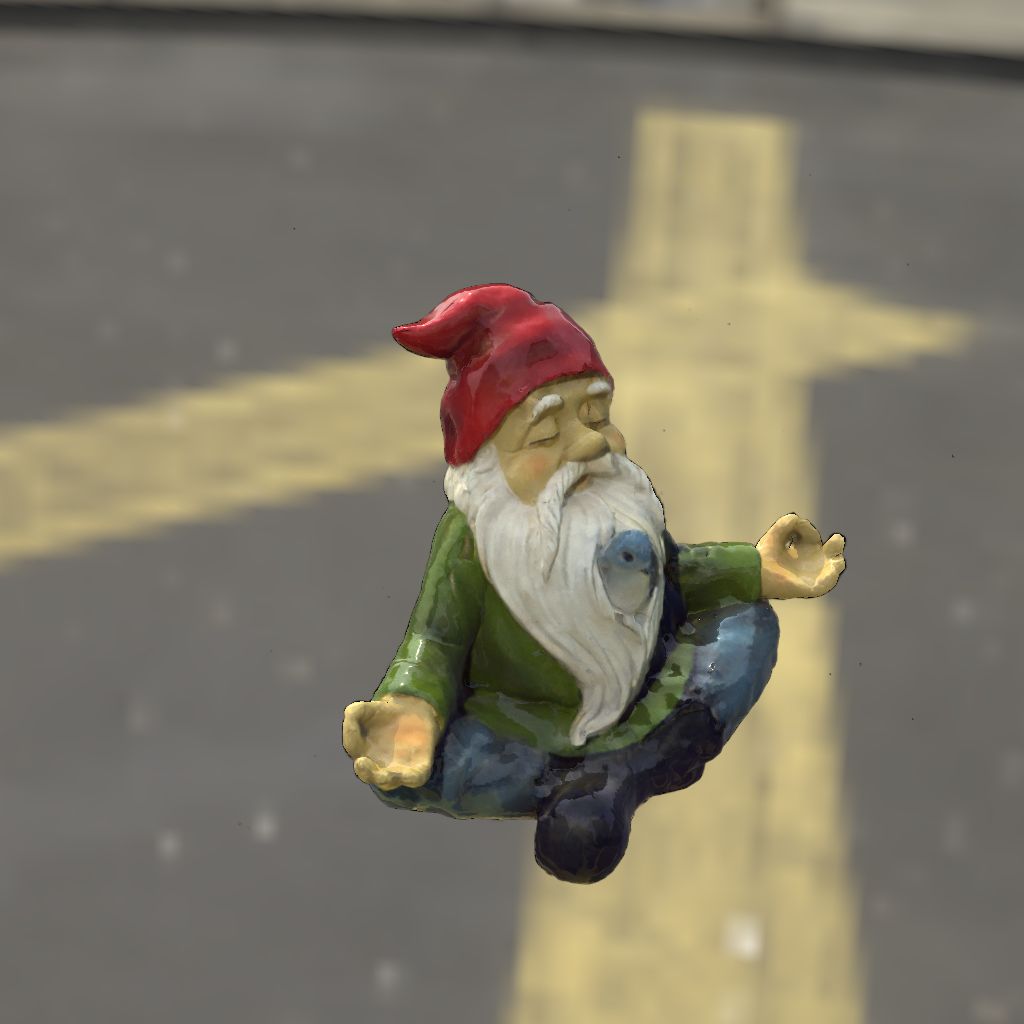}  &
                \includegraphics[width=0.16\linewidth,trim={10cm 4cm 4cm 9cm},clip]{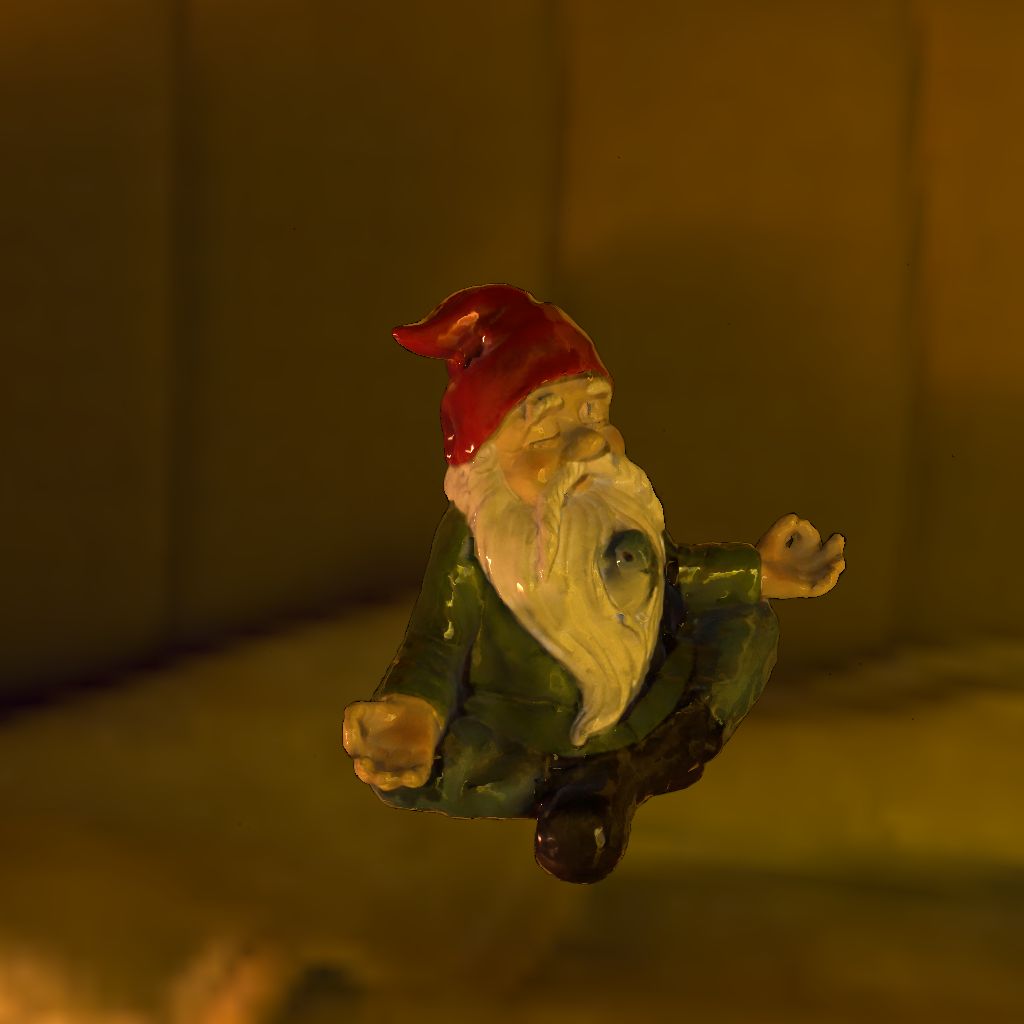}  \\
                & \scriptsize{NVS} & \scriptsize{Normal} & \scriptsize{Albedo} & \scriptsize{Envmap} & \scriptsize{Relit 1} & \scriptsize{Relit 2}
            \end{tabular}
        \end{tabular}
    }
    \end{center}
    \vspace{-5mm}
    \caption{ 
        \textbf{Qualitative Results from the Stanford-ORB Dataset.}
        We provide qualitative results including NVS, surface normal, albedo, environment map, and relighting results.
        \label{fig:exp:real}
    }
    \vspace{-1mm}
\end{figure}

In Fig.~\ref{fig:exp:real}, we provide qualitative results of real-world cases from the Stanford-ORB dataset~\cite{kuang2024stanford}, to further validate the effectiveness of our method. More results and discussions for real-world applications can be found in Appendix~\ref{app:maskanalysis}.

\subsection{Ablation Studies}
\label{exp:abl}

To validate the effectiveness of the aforementioned techniques proposed in this paper, we conduct ablation studies on the TensoIR Synthetic dataset. We compare inverse rendering performance in terms of NVS, relighting, albedo estimation, and normal accuracy, as shown in Table~\ref{table:exp:ablation}.

\begin{table}[H]
\vspace{-1mm}
\begin{center}
    \resizebox{1.0\linewidth}{!}{ % Resizes the table to fit within the text width
    \begin{tabular}{l|cccc}
    \toprule
    Description & NVS & Relighting & Albedo & Normal\\
    & PSNR $\uparrow$& PSNR $\uparrow$& PSNR $\uparrow$& MAE $\downarrow$ \\
    \midrule
    w/o shape alignment & 35.95 & 26.39 & 26.72 & 8.29 \\ 
    w/o appearance refinement & 35.07 & 27.95 & 29.31 & 4.42 \\ 
    w/o occlusion modeling & 35.87 & 27.36 & 27.80 & 6.17 \\ 
    w/o indirect light & 36.01 & 28.92 & 29.18 & 4.97 \\ 
    \textbf{full modeling} & \textbf{36.45} & \textbf{29.95} & \textbf{29.41} & \textbf{4.08} \\ 
    \bottomrule
\end{tabular}
    }
    \end{center}
\vspace{-5mm}
\caption{
\label{table:exp:ablation} 
\textbf{Ablation Studies.} We conduct extensive quantitative comparisons to validate the effectiveness of {\name}.
}
\vspace{-4mm}
\end{table}

\vspace{-3mm}
\paragraph{Shape Alignment.} Firstly, we conduct an ablation study on our {\meshsampler} by no longer generating shape-consistent Gaussian points from the barycentric pattern. Instead, we randomly sample $K$ Gaussian points on each face and allow them to automatically learn scale and rotation attributes. The quantitative results demonstrate that the shape alignment between Gaussian points and the underlying mesh guidance is crucial to {\name}, as both normal modeling and light transport modeling depend on it.

\vspace{-3mm}
\paragraph{Appearance Refinement.} Next, as confirmed by the quantitative results, appearance refinement effectively enhances rendering quality but has minimal impact on decomposition results and normal quality. This is because the appearance refinement is applied at the end of the training stage, once material and geometry recovery has already converged.

\begin{figure}[H]
\vspace{0mm}
    \begin{center}
    \resizebox{\linewidth}{!}{
        \setlength{\tabcolsep}{1pt}
        \setlength{\fboxrule}{1pt}
        \hspace{-4mm}
        \begin{tabular}{c}
            \begin{tabular}{cccccc}
                \rotatebox{90}{~~\scriptsize{w/o Occ.}} &
                \vspace{0mm}
                \includegraphics[width=0.19\linewidth,trim={1.5cm 1.5cm 0.5cm 5cm},clip]{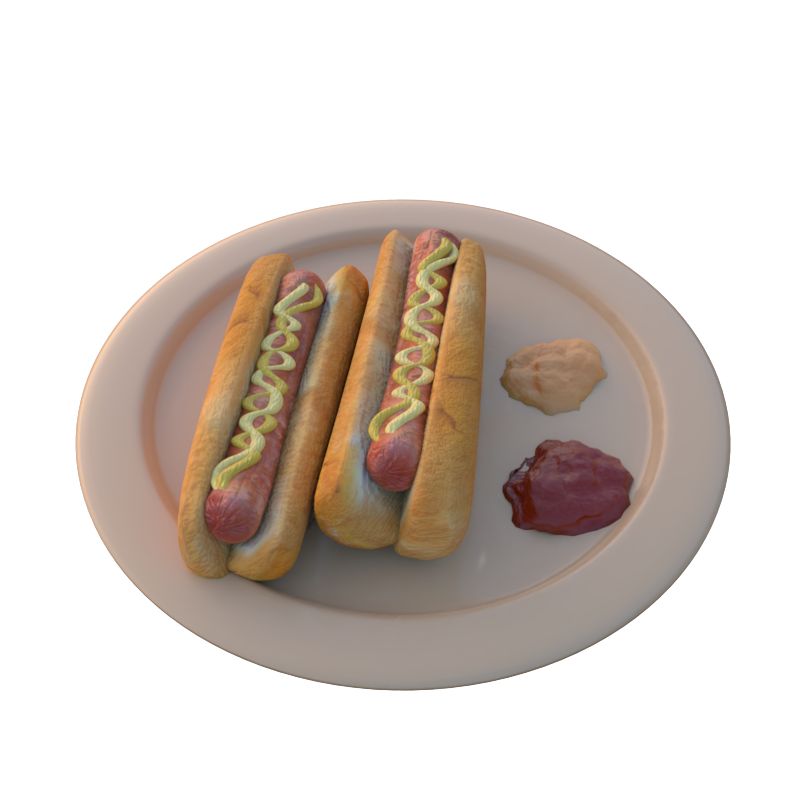}  &
                \includegraphics[width=0.19\linewidth,trim={1.5cm 1.5cm 0.5cm 5cm},clip]{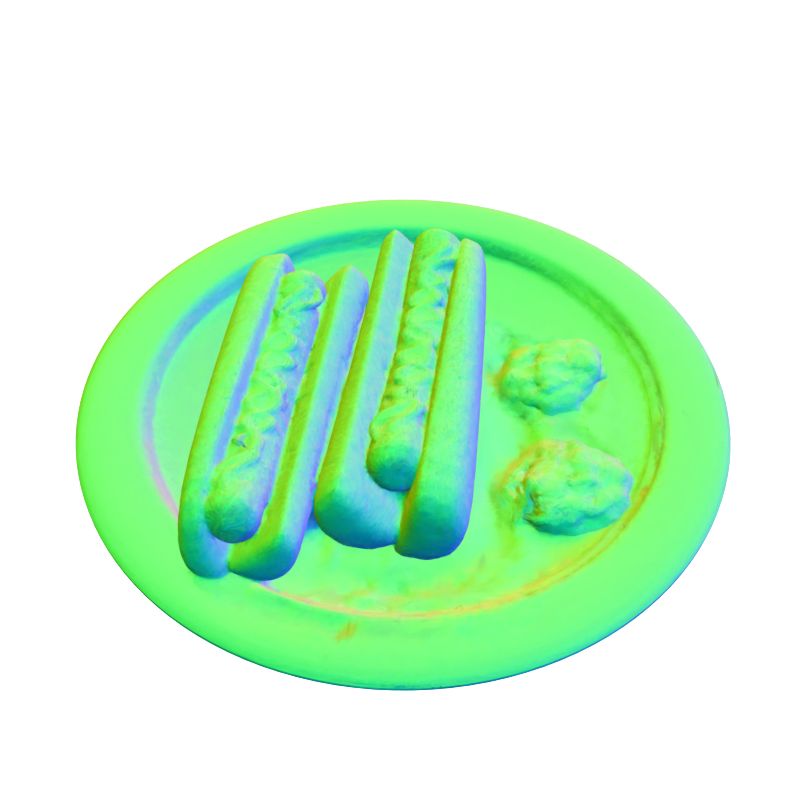}  &
                \includegraphics[width=0.19\linewidth,trim={1.5cm 1.5cm 0.5cm 5cm},clip]{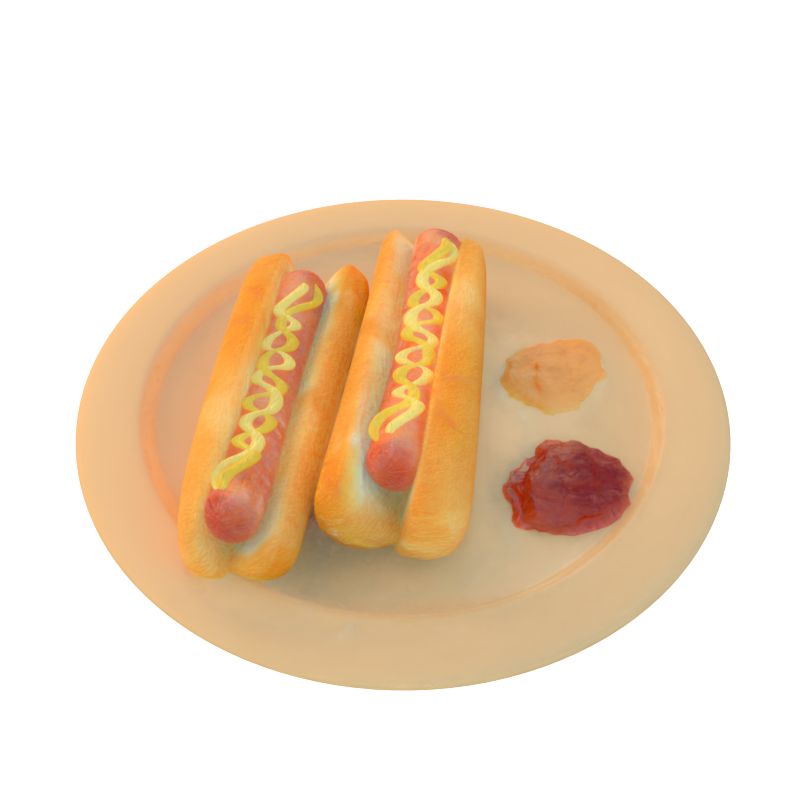}  &
                \includegraphics[width=0.19\linewidth,trim={1.5cm 1.5cm 0.5cm 5cm},clip]{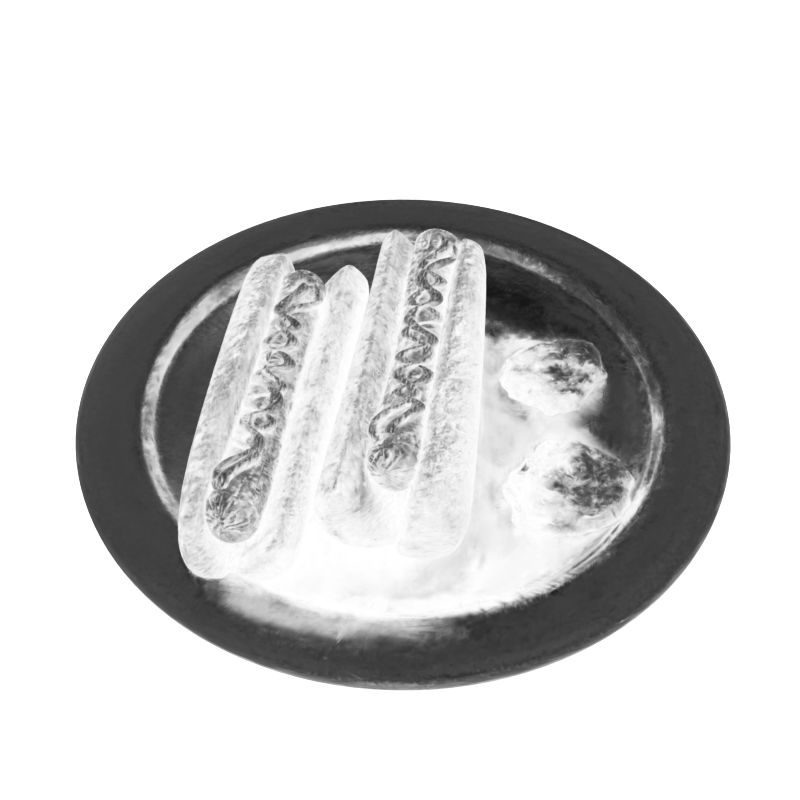}  &
                \includegraphics[width=0.19\linewidth,trim={1.5cm 1.5cm 0.5cm 5cm},clip]{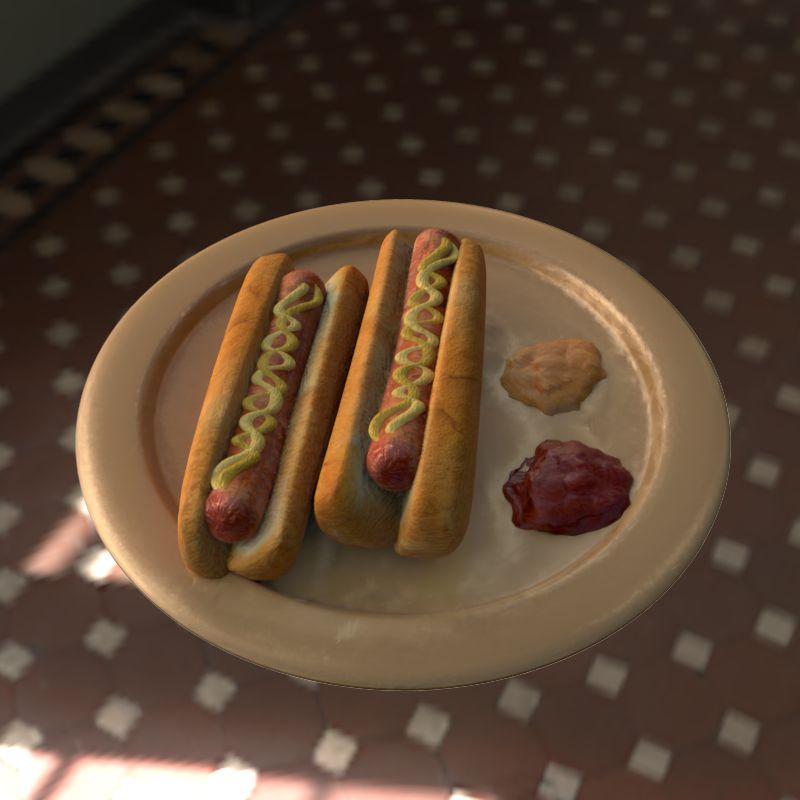}  \\
                \rotatebox{90}{~~\scriptsize{w/ Occ.}} &
                \vspace{-1mm}
                \includegraphics[width=0.19\linewidth,trim={1.5cm 1.5cm 0.5cm 5cm},clip]{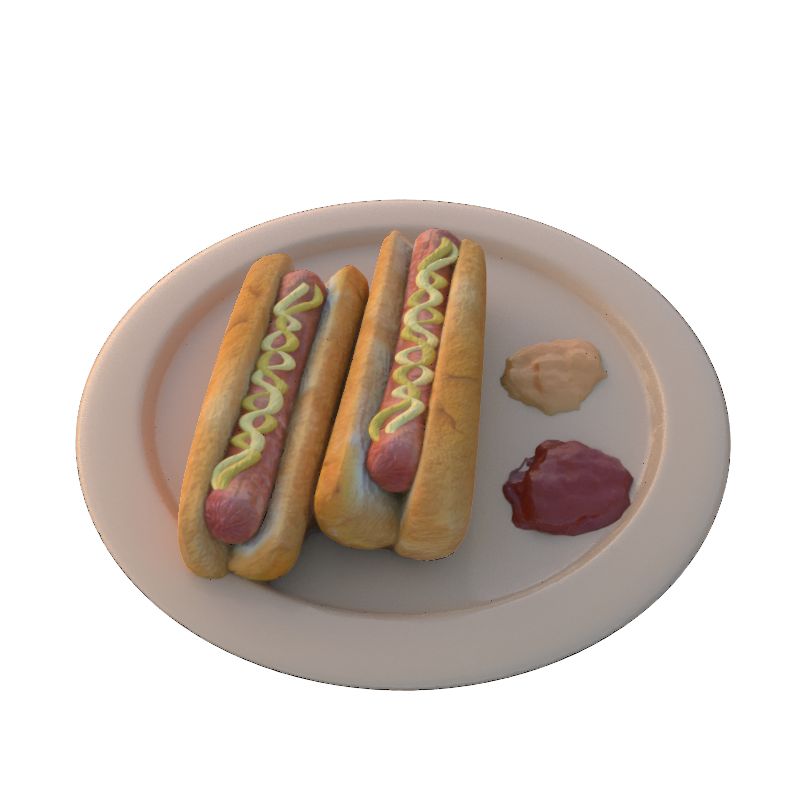}  &
                \includegraphics[width=0.19\linewidth,trim={1.5cm 1.5cm 0.5cm 5cm},clip]{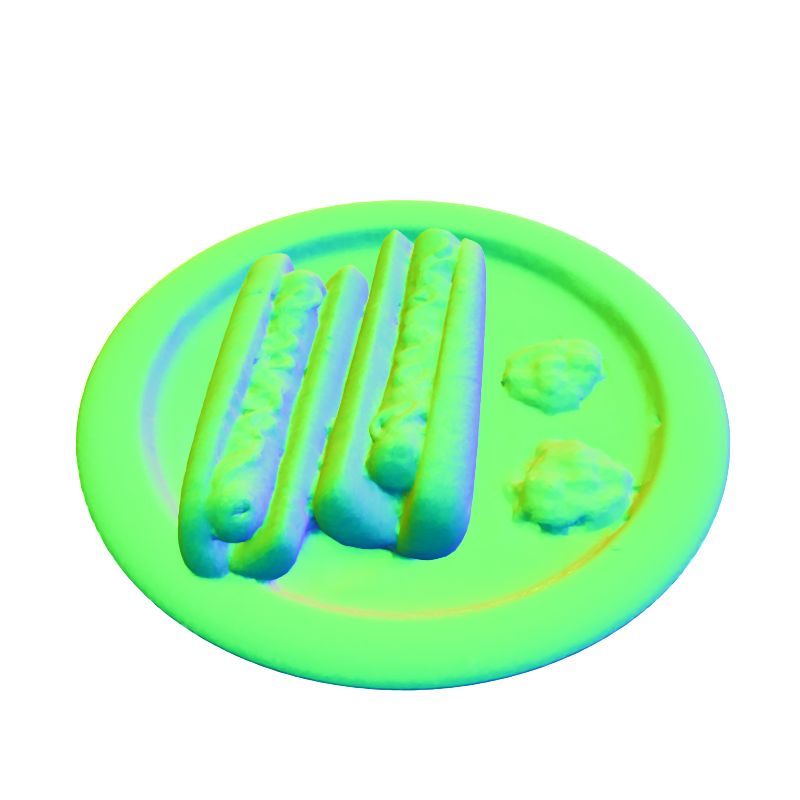}  &
                \includegraphics[width=0.19\linewidth,trim={1.5cm 1.5cm 0.5cm 5cm},clip]{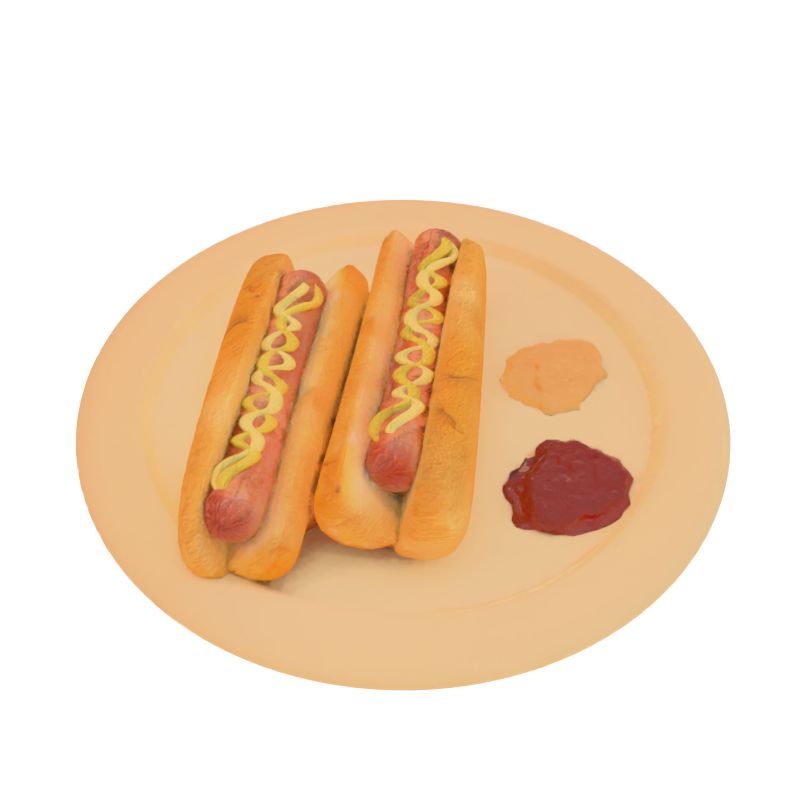}  &
                \includegraphics[width=0.19\linewidth,trim={1.5cm 1.5cm 0.5cm 5cm},clip]{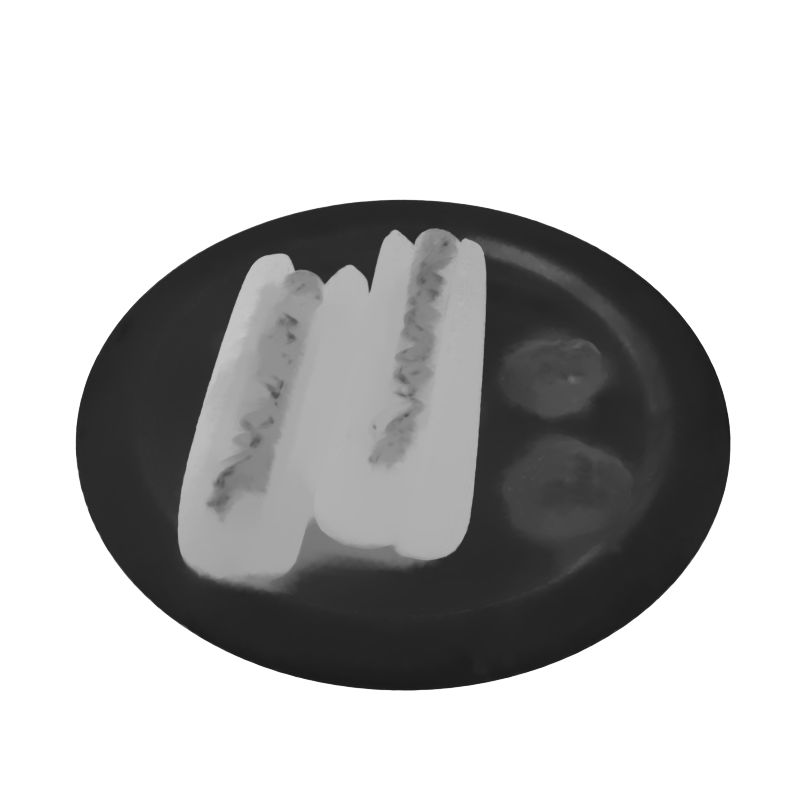}  &
                \includegraphics[width=0.19\linewidth,trim={1.5cm 1.5cm 0.5cm 5cm},clip]{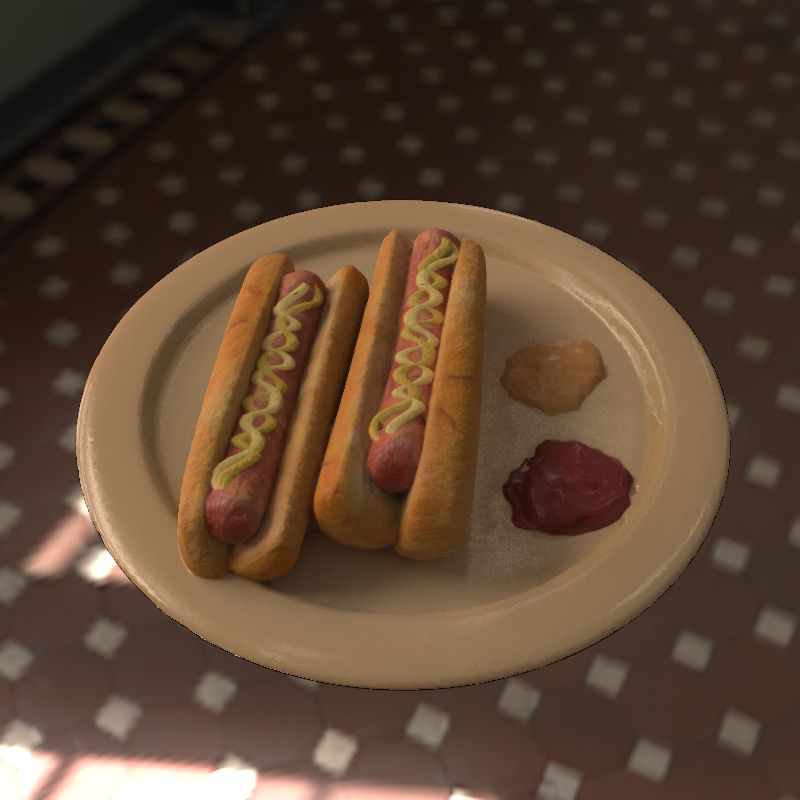} \\
                & \scriptsize{NVS} & \scriptsize{Normal} & \scriptsize{Albedo} & \scriptsize{Roughness} & \scriptsize{Relit}~
            \end{tabular}
        \end{tabular}
    }
    \end{center}
    \vspace{-4mm}
    \caption{ 
        \textbf{Qualitative Ablation Studies on Self-Occlusion Modeling.}
        Despite comparable NVS quality, the absence of occlusion modeling leads to significantly degraded performance, in terms of normal accuracy, material estimation, and relighting quality.
        \label{fig:exp:abl}
    }
    \vspace{1mm}
\end{figure}

\vspace{-3mm}
\paragraph{Self-Occlusion Modeling.} Lastly, we validate the effectiveness of self-occlusion modeling. By setting $\Ogs=0$, we evaluate the performance of {\name} without occlusion modeling. Similarly, by setting $\Lind=0$, we assess the performance of {\name} without indirect light. As illustrated in Table~\ref{table:exp:ablation} and Fig.~\ref{fig:exp:abl}, the occlusion modeling has little impact on NVS performance but can be crucial for accurate decomposition and relighting.

\section{Discussion}
\label{sec:conclusion}

\paragraph{Limitation.}
While {\name} demonstrates state-of-the-art performance in NVS and relighting tasks, it still faces several challenges that motivate further research. First, its geometry guidance is derived from the isosurface. Although this significantly improves the geometry performance of 3DGS, it also requires masks during training. Moreover, constrained by the grid resolution of the isosurface, {\name} struggles with thin structures and surfaces with complicated geometry. A promising direction for future work would be to explore how to apply adaptive resolution to accommodate detailed geometry, enabling its extension to scene-level tasks. More comprehensive discussions are provided in Appendix~\ref{app:maskanalysis} \& \ref{app:resolution}.

\vspace{-3mm}
\paragraph{Conclusion.}
We propose {\name}, a novel hybrid representation that enhances 3DGS with explicit geometric guidance and differentiable PBR equations. {\name} achieves superior efficiency and state-of-the-art inverse rendering performance across various datasets, which validates the effectiveness of its geometry guidance. Compared to prior 3DGS-based inverse rendering approaches, {\name} significantly improves the normal quality and the accuracy of light transport modeling, delivering better decomposition results. Moreover, compared to state-of-the-art implicit-field-based approaches which require several hours to optimize, our method achieves competitive performance while converging within 10-15 minutes. We will release all the code to facilitate related research.

\clearpage

\section*{Acknowledgement}

This work was supported in part by the National Key R\&D Program of China under Grants 2022ZD0160801. This work was also supported by the projects of Beijing Science and Technology Program (Z231100007423011) and National Natural Science Foundation of China (No. 62376012), which is a research achievement of State Key Laboratory of Multimedia Information Processing and Key Laboratory of Science, Technology and Standard in Press Industry (Key Laboratory of Intelligent Press Media Technology).

~

{
    \small
    \bibliographystyle{ieeenat_fullname}
    \bibliography{references}
}

\onecolumn
\appendix
\section{Appendix Overview}

In the appendix, we provide a comprehensive explanation about the details of our work, including detailed implementations and limitations of our method, as well as supplementary results from both quantitative and qualitative experiments.

The appendix begins with a detailed explanation of the {\meshsampler} in Appendix~\ref{app:meshsampler}, followed by an overview of the implementation details for our loss functions in Appendix~\ref{app:loss}. In Appendix~\ref{app:appref}, we provide details of the appearance refinement technique. Furthermore, to explore the limitations of our method, we include a discussion on the input mask required during training in Appendix~\ref{app:maskanalysis} and the impact of isosurface resolution in Appendix~\ref{app:resolution}. Finally, we demonstrate how {name} can be enhanced with an initial mesh in Appendix~\ref{app:imesh} and incorporated with path tracing in Appendix~\ref{app:pt}.

\section{Explanation of {\meshsampler}}
\label{app:meshsampler}

We first describe the details of our {\meshsampler}. As discussed in Sec.~\ref{sec:method_mesh2gaussian}, the {\meshsampler} takes a triangle mesh as input and generates a set of Gaussian points corresponding to the shape of the underlying mesh. The core idea of {\meshsampler} is to maintain shape consistency between the Gaussian points and the mesh guidance.

\subsection{Overview}

The straightforward implementation of our {\meshsampler} involves sampling several Gaussian points on the mesh surface, which are then optimized in terms of scale and rotation to minimize the depth map difference between the Gaussian points and the target mesh. However, this approach introduces an additional optimization step that must be re-executed each time the mesh is modified, resulting in reduced optimization efficiency and an unstable training process.

Instead of maintaining shape consistency in real-time, we propose utilizing a predefined heuristic function $\meshSamplerFunc$ to achieve an approximate alignment. As described in Eq.~\ref{eq:meshsampler} (Sec.~\ref{sec:method_mesh2gaussian}), the {\meshsampler} $\meshSamplerFunc$ takes arbitrary triangle meshes as input and outputs Gaussian point attributes, including position $\centerPos$, scale $\threeDScaling$, rotation $\threeDRotation$, and normal $\normal$. This process acts as a generalized adapter between the input meshes and the corresponding Gaussian points. We provide a detailed implementation in Sec.~\ref{app:meshsampler:eq1}, followed by an analysis of shape consistency between the underlying mesh and our generated Gaussian points in Sec.~\ref{app:meshsampler:analysis}. Then, we explain the OOM issue at the beginning of the training stage and propose the vertex-sample stage to address it in Sec.~\ref{app:meshsampler:warmup}.

\subsection{Implementation of Eq.~1}
\label{app:meshsampler:eq1}

Specifically, given the triangle mesh, each triangle face $F_i$ comprises three vertices $\mathbf{P}_i=(\mathbf{p}_{i1}, \mathbf{p}_{i2}, \mathbf{p}_{i3})$ with their vertex normals $\mathbf{N}_i=(\mathbf{n}_{i1}, \mathbf{n}_{i2}, \mathbf{n}_{i3})$. We symmetrically sample 6 points on $F_i$ with barycentric coordinates:
\begin{equation}
\begin{aligned}
\mathbf{b}_1&=(u,u,1-2u)\eqcomma&\mathbf{b}_4&=(v,v,1-2v)\eqcomma\\
\mathbf{b}_2&=(u,1-2u,u)\eqcomma&\mathbf{b}_5&=(v,1-2v,v)\eqcomma\\
\mathbf{b}_3&=(1-2u,u,u)\eqcomma&\mathbf{b}_6&=(1-2v,v,v)\eqstop
\end{aligned}
\end{equation}
And we can obtain 6 midpoints $m_{jk}$:
\begin{equation}
\begin{aligned}
\mathbf{m}_{12}&=\frac{\mathbf{b}_1+\mathbf{b}_2}2=\left(u,\frac{1-u}2,\frac{1-u}2\right)\eqcomma&
\mathbf{m}_{45}&=\frac{\mathbf{b}_4+\mathbf{b}_5}2=\left(v,\frac{1-v}2,\frac{1-v}2\right)\eqcomma\\
\mathbf{m}_{23}&=\frac{\mathbf{b}_2+\mathbf{b}_3}2=\left(\frac{1-u}2,\frac{1-u}2,u\right)\eqcomma&
\mathbf{m}_{56}&=\frac{\mathbf{b}_5+\mathbf{b}_6}2=\left(\frac{1-v}2,\frac{1-v}2,v\right)\eqcomma\\
\mathbf{m}_{31}&=\frac{\mathbf{b}_3+\mathbf{b}_1}2=\left(\frac{1-u}2,u,\frac{1-u}2\right)\eqcomma&
\mathbf{m}_{64}&=\frac{\mathbf{b}_6+\mathbf{b}_4}2=\left(\frac{1-v}2,v,\frac{1-v}2\right)\eqstop
\end{aligned}
\end{equation}
Given an attribute $\mathbf{A}_i = (\mathbf{a}_{i1}, \mathbf{a}_{i2}, \mathbf{a}_{i3})$ defined at the triangle vertices and a barycentric coordinate $(q_1,q_2,q_3)$, we represent the barycentric interpolation as:
\begin{equation}
(q_1,q_2,q_3)\odot \mathbf{A}_i =q_1\mathbf{a}_{i1}+q_2\mathbf{a}_{i2}+q_3\mathbf{a}_{i3}\eqstop
\end{equation}
Then, for each midpoint $m_{jk}$, we sample a Gaussian point as:
\begin{equation}
\label{eq:app:meshsample}
\begin{aligned}
\centerPos&=\mathbf{m}_{jk}\odot \mathbf{P}_i\eqcomma&
\normal&=\mathbf{m}_{jk}\odot \mathbf{N}_i\eqcomma\\
\threeDScaling_x&=\alpha_{jk}\|\mathbf{b}_k\odot \mathbf{P}_i-\mathbf{m}_{jk}\odot \mathbf{P}_i\|_2\eqcomma&
\threeDRotation_x&=\frac{\mathbf{b}_{k}\odot \mathbf{P}_i-\mathbf{m}_{jk}\odot \mathbf{P}_i}{\|\mathbf{b}_{k}\odot \mathbf{P}_i-\mathbf{m}_{jk}\odot \mathbf{P}_i\|_2}\eqcomma\\
\threeDScaling_y&=\frac{\text{Area}(F_i)}{\beta_{jk}\|\mathbf{b}_k\odot \mathbf{P}_i-\mathbf{m}_{jk}\odot \mathbf{P}_i\|_2}\eqcomma&
\threeDRotation_y&=\normal\times\threeDRotation_x\eqcomma\\
\threeDScaling_z&=\delta_{jk}\eqcomma&
\threeDRotation_z&=\normal\eqstop
\end{aligned}
\end{equation}
Here, Eq.~\ref{eq:app:meshsample} provide the formulation of our heuristic
function $\meshSamplerFunc$, with $u,v,\alpha_{jk},\beta_{jk},\delta_{jk}$ as hyperparameters. To achieve the generalized geometric alignment, we practically set these parameters as follows:
\begin{equation}
\hspace{-2.3cm}
\begin{aligned}
u&=0.07\eqcomma\\
v&=0.22\eqcomma\\
\alpha_{12}=\alpha_{23}=\alpha_{31}&=0.80\eqcomma\\
\alpha_{45}=\alpha_{56}=\alpha_{64}&=2.08\eqcomma\\
\beta_{12}=\beta_{23}=\beta_{31}&=15.0\eqcomma\\
\beta_{45}=\beta_{56}=\beta_{64}&=13.0\eqcomma\\
\delta_{12}=\delta_{23}=\delta_{31}=\delta_{45}=\delta_{56}=\delta_{64}&=4.5\times10^{-5}\eqstop\\
\end{aligned}
\end{equation}

\subsection{Analysis of Shape Consistency}
\label{app:meshsampler:analysis}

\paragraph{Metrics}

We begin by providing a formal definition of shape consistency, as 3DGS points lack inherent geometry boundaries. In the context of inverse rendering, geometry quality is critical for accurate light transport modeling. Therefore, we measure shape consistency based on light transport accuracy, specifically in terms of errors in light reflection directions and light transfer distances. For computational convenience, we assess these errors using a ray-tracing approach. Specifically, given a reference mesh model and a set of viewpoints, we compute the intersections between per-pixel camera rays and the underlying mesh surface in screen space. By measuring the Mean Angular Error (MAE) of the reflected ray directions and the distance between intersection points and the camera position, we evaluate shape consistency, as shown in Table~\ref{app:shape:consistency}.

\begin{table}[h]
    \centering
    \resizebox{0.7\textwidth}{!}{
        \begin{tabular}{ccccc}
            \toprule
            \textbf{Metrics} & Air balloons & Chair & Hotdog & Jugs\\
            \midrule
            Reflecting Directions (MAE~$\downarrow$) & 1.23 & 0.81 & 0.74 & 0.89\\
            Transfer Distance (L1~$\downarrow$)& 0.016& 0.010 & 0.012 & 0.013\\
            \bottomrule
        \end{tabular}
    }
    \caption{We measure shape consistency on the Synthetic4Relight dataset, including reflection directions (in degrees) and transfer distances (relative to the size of the scene's bounding box).}
    \label{app:shape:consistency}
\end{table}

\paragraph{Discussion}

Numerous works have attempted to ground 3DGS points to the triangle mesh surface. While many focus on making Gaussian points deformable~\cite{wac2024gamesmeshbasedadaptingmodification, MeshGaussian2024} or enhancing rendering quality~\cite{liu2025meshaligned, meshgs}, only a few have addressed the enhancement of geometry quality in 3DGS through mesh-Gaussian binding.

MeshSplats~\cite{tobiasz2025meshsplatsmeshbasedrenderinggaussian} converts Gaussian points into triangle slices, enabling ray tracing techniques for advanced lighting effects such as shadows and reflections. However, this method operates on pretrained Gaussian points, aligning them with the mesh in a post-hoc manner. This approach is not compatible with inverse rendering methods, which require a fully differentiable pipeline to propagate gradients from photometric loss functions to the underlying geometry. Another method~\cite{lin2024directlearningmeshappearance} proposes a simple technique for differentiable mesh-Gaussian alignment, enabling end-to-end training. However, this technique only considers mesh-Gaussian alignment along the normal direction, leaving tangent-space alignment uncontrolled—similar to the ablated Gaussian sampling approach discussed in Sec.~\ref{exp:abl}. As demonstrated in Table~\ref{table:exp:ablation} of Sec.~\ref{exp:abl}, simply applying this method to inverse rendering tasks results in shape disparity between the 3DGS and the mesh, leading to inaccurate light transport modeling and degraded decomposition results.

In contrast, we have carefully designed our {\meshsampler} to ensure shape consistency between the Gaussian points and the mesh guidance. The inherently shape-consistent nature of {\meshsampler} allows for end-to-end optimization of geometry guidance during training, enabling precise normal estimation and accurate light transport modeling for superior inverse rendering performance.

\subsection{Vertex-Sampling Stage}
\label{app:meshsampler:warmup}

\begin{wrapfigure}[9]{r}{0.2\textwidth}
    \centering
    \vspace{-4mm}
    \includegraphics[width=0.9\linewidth]{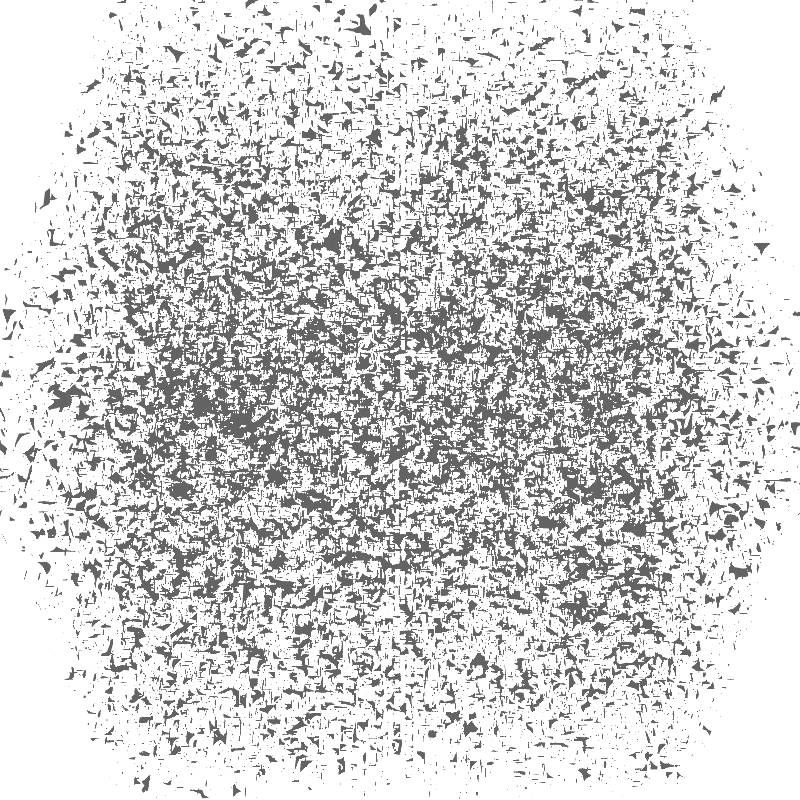}
    \vspace{-2mm}
    \caption{ \label{fig:supp:flexi} Initial mesh slices of FlexiCubes. }
    \vspace{-3mm}
\end{wrapfigure}

As outlined in Sec.~\ref{app:meshsampler:eq1}, we typically sample six Gaussian points from each triangle surface. However, at the start of the training stage, the isovalues of FlexiCubes are randomly initialized, leading to an excessive number of triangle slices, as shown in Fig.~\ref{fig:supp:flexi}. Directly sampling six Gaussian points per face in this context can incur substantial memory costs and reduce training efficiency.

To address this issue, we implement an vertex-sampling stage for {\meshsampler} at the beginning of training (covering the first 5\%). During this stage, {\meshsampler} outputs a significantly reduced number of Gaussian points by performing vertex sampling. Specifically, given a mesh vertex $\mathbf{v}$ with its normal $\mathbf{n_v}$, {\meshsampler} samples a single Gaussian point with $\centerPos=\mathbf{v},\normal=\mathbf{n_v}$, and:
\begin{equation}
    \begin{aligned}
    \threeDScaling_x=\threeDScaling_y=\sqrt{\frac{k}{3}\sum_{F_i\in\text{Star}(\mathbf{v})}\text{Area}(F_i)}\eqcomma\,
    \threeDScaling_z=\delta_{jk}\eqcomma\\
    \threeDRotation_x=(0,0,1)\times\mathbf{n_v}\eqcomma\,
    \threeDRotation_y=\mathbf{n_v}\times\threeDRotation_x\eqcomma\,
    \threeDRotation_z=\mathbf{n_v}\eqstop\\
    \end{aligned}
\end{equation}
Once the vertex-sampling stage concludes, {\meshsampler} switches to face-sampling, as described in Sec.~\ref{app:meshsampler:eq1}.

\section{Explanation of PBR Attribute Modeling}
\label{app:meshsampler:mlp}

Since Gaussian points are generated in real time from the underlying mesh, directly modeling these attributes as learnable parameters is impractical due to the varying number of Gaussian points during training. Instead, as described in Eq.~\ref{eq:mlptexture} (Sec.~\ref{sec:method_render}), we employ multi-resolution hash grids $\mathcal{E}_\mathrm{d},\mathcal{E}_\mathrm{s}$ to construct attribute fields.
Specifically, the hash grids incorporate the multi-resolution hash encoders introduced in~\citep{muller2022instant}, followed by small MLP headers. We implement them using tiny-cuda-nn\citep{tiny-cuda-nn}. Detailed parameters can be found in Table~\ref{app:meshsampler:mlp:config}.

\begin{table}[h]
\centering
\resizebox{0.8\textwidth}{!}{
    \begin{tabular}{lll}
        \toprule
        Module & Parameter & Value \\
        \midrule
        & Number of levels & $16$\\
        & Max.entries per level (hash table size) & $2^{19}$\\
        $\mathcal{E}_\mathrm{d}$/$\mathcal{E}_\mathrm{s}$~HashEnc & Number of feature dimensions per entry & $2$\\
        & Coarsest resolution & $32$\\
        & Finest resolution & $4096$\\
        \midrule
        & MLP layers & $32\times32\times32\times6$\\
        $\mathcal{E}_\mathrm{d}~\text{MLP}$ & Initialization & Kaiming-uniform\\
        & Final activation & Sigmoid\\
        \midrule
        & MLP layers & $32\times32\times3$\\
        $\mathcal{E}_\mathrm{s}~\text{MLP}$ & Initialization & Kaiming-uniform\\
        & Final activation & None\\
        \bottomrule
    \end{tabular}
}
\caption{Parameters of Spatial MLP}
\label{app:meshsampler:mlp:config}
\vspace{-3mm}
\end{table}

\section{Details of Loss Functions}
\label{app:loss}

\subsection{Photometric Term}
\label{app:loss:photometric}

During the training stage, for the $i$-th view , {\name} differentiably renders a RGB image $\image_\mathrm{pred}^{(i)}\in \mathbb{R}^{H\times W\times3}$ and takes the alpha channel as the mask $\mask_\mathrm{pred}^{(i)}\in \mathbb{R}^{H\times W\times1}$. Given the ground truth $\image_\mathrm{gt}^{(i)}$ and $\mask_\mathrm{gt}^{(i)}$ for view $i$, the photometric loss is computed as:
\begin{equation}
\label{eq:app:photo-loss}
\begin{aligned}
\loss_\mathrm{photo} =& \loss_{1} + \lambda_\mathrm{ssim}\loss_\mathrm{SSIM} + \lambda_\mathrm{mask}\loss_\mathrm{mask}\\
=& \|\image_\mathrm{gt}^{(i)}-\image_\mathrm{pred}^{(i)}\|_1 + \lambda_\mathrm{ssim} \mathrm{SSIM}(\image_\mathrm{gt}^{(i)},\image_\mathrm{pred}^{(i)}) + \lambda_\mathrm{mask}\|\mask_\mathrm{gt}^{(i)}-\mask_\mathrm{pred}^{(i)}\|^2_2\eqstop
\end{aligned}
\end{equation}
Here, $\lambda_\mathrm{ssim}=0.2$ and $\lambda_\mathrm{mask}=5.0$ for all the cases.

\subsection{Entropy Regularization Term}

Following DMTet and FlexiCubes~\citep{shen2021dmtet,shen2023flexicubes}, we add an entropy loss to constrain the shape. Specifically, we employ FlexiCubes as the underlying geometric representation, which defines a scalar function $\sdfFunc: \mathbb{R}^3 \rightarrow \mathbb{R}$ on the underlying cube grids $\mathcal{G}(\mathcal{V},\mathcal{E})$ and then extracts isosurfaces via the differential Dual Marching Cubes introduced by~\citep{shen2023flexicubes}. Given an edge $(v_i, v_j)$ from edge set $\mathcal{E}$, the SDF values defined on the endpoints $v_i,v_j$ are respectively $\sdfFunc(v_i)$ and $\sdfFunc(v_j)$.

Then, we can compute the regularization term as:

\begin{equation}
\begin{aligned}
\loss_\mathrm{sdf} &= \sum_{(v_i, v_j)\in\mathcal{E}, \text{sgn}(\sdfFunc(v_i)) \ne \text{sgn}(\sdfFunc(v_j))} \mathcal{H}(\sdfFunc(v_i), \text{sgn}(\sdfFunc(v_j))) + \mathcal{H}(\sdfFunc(v_j), \text{sgn}(\sdfFunc(v_i)))
\end{aligned}
\end{equation}

Here, $\mathcal{H}$ denotes the binary cross entropy. By encouraging the same sign of $\sdfFunc$, such a regularization term penalize internal geometry and floaters.

\subsection{Smoothness Regularization Term}

Following prior works~\citep{Munkberg_2022_CVPR,hasselgren2022nvdiffrecmc,R3DG2023}, we apply smoothness regularization on albedo, roughness, and metallic to prevent dramatic high-frequency variations. Given the positions $\centerPos$ of Gaussian points, the albedo, roughness and metallic attributes are generated by the hash grids:

\begin{equation}
\begin{aligned}
\diffAlbedo = \mathcal{E}_\mathrm{d}(\centerPos)\eqcomma
\left(\materialRoughness, \materialMetalness\right) = \mathcal{E}_\mathrm{s}(\centerPos)\eqstop
\end{aligned}
\end{equation}

While applying a small perturbation $\Delta\centerPos\sim \mathcal{N}(0,\sigma^2)$ on $\centerPos$ can yield a different set of attributes:

\begin{equation}
\begin{aligned}
\diffAlbedo' = \mathcal{E}_\mathrm{d}(\centerPos+\Delta\centerPos)\eqcomma
\left(\materialRoughness', \materialMetalness'\right) = \mathcal{E}_\mathrm{s}(\centerPos+\Delta\centerPos)\eqstop
\end{aligned}
\end{equation}

The smoothness is then computed as:

\begin{equation}
\begin{aligned}
\loss_{smooth}=\|\diffAlbedo-\diffAlbedo'\|_1 + \|\materialRoughness-\materialRoughness'\|_1 + \|\materialMetalness-\materialMetalness'\|_1
\end{aligned}
\end{equation}

Here, $\sigma=0.01$.

\subsection{Light Regularization Term}

Following NVdiffrecmc~\citep{hasselgren2022nvdiffrecmc}, we add a light regularization that is based on monochrome image loss between the demodulated lighting terms and the reference image. Given the demodulated diffuse lighting $L_\mathrm{d}$, the specular lighting $L_\mathrm{s}$, and the reference image $I_\mathrm{gt}$, the regularization is computed as follows:

\begin{equation}
    \begin{aligned}
    \loss_\mathrm{light}=\|Y(L_\mathrm{d}+L_\mathrm{s})-V(I_\mathrm{gt})\|
    \end{aligned}
\end{equation}

Here, $Y(\x)=(\x_r+\x_g+\x_b)/3$ is the luminance operator, and $V(\x) = \max (\x_r, \x_g, \x_b)$ is the HSV value component. As discussed in the original paper~\citep{hasselgren2022nvdiffrecmc}, this regularization is based on the assumption that the demodulated lighting is mostly monochrome, \ie, $Y(\x)\sim V(\x)$, and it is proven to be effective for shadow disentanglement.

\subsection{Final Loss}

The final loss $\loss$ is computed as:

\begin{equation}
\begin{aligned}
\loss = \loss_\mathrm{photo} + \lambda_\mathrm{sdf}\loss_\mathrm{sdf} + \lambda_\mathrm{smooth}\loss_\mathrm{smooth} + \lambda_\mathrm{light}\loss_\mathrm{light}
\end{aligned}
\end{equation}

Here, $\lambda_\mathrm{sdf}$ is initially set to $0.2$ at the start of the training stage and is linearly decreased to $0.01$ by the midpoint of the training, with $\lambda_\mathrm{smooth}=0.03$ and $\lambda_\mathrm{light}=0.15$.

\section{Explanation of Appearance Refinement}
\label{app:appref}

In this section, we provide a detailed explanation of the implementation of our appearance refinement technique. Specifically, we enable {\name} to transition to deferred shading and optimize 3DGS attributes to adjust their displacement, including positions, scales, rotations, and opacities.

\paragraph{Deferred Shading}

As described in Sec.~\ref{sec:method_render}, we extract per-Gaussian PBR attributes and conduct Monte Carlo sampling to obtain Gaussian-wise colors. Then, we utilize alpha-blending to rasterize Gaussian-wise colors into screen-space pixels, which constitute the forward shading:
\begin{align}
\mathrm{c}_\mathrm{GS}&=\text{PBR}(\centerPos, \normal, \diffAlbedo, \materialRoughness, \materialMetalness)\eqcomma&I_\mathrm{RGB}&=\text{Rasterize}(\mathrm{c}_\mathrm{GS})\eqstop
\end{align}
During the appearance Refinement, we conduct deferred shading instead. Specifically, we render per-Gaussian attributes into screen-space attribute map. 
\begin{equation}
\begin{aligned}
    I_\mathrm{\mu}&=\text{Rasterize}(\centerPos)\eqcomma&
    I_\mathrm{n}&=\text{Rasterize}(\normal)\eqcomma\\
    I_\mathrm{a}&=\text{Rasterize}(\diffAlbedo)\eqcomma&
    I_\mathrm{\rho}&=\text{Rasterize}(\materialRoughness)\eqcomma&
    I_\mathrm{m}&=\text{Rasterize}(\materialMetalness)\eqstop
\end{aligned}
\end{equation}
Then, PBR is conducted in screen-space:
\begin{equation}
    I_\mathrm{RGB}=\text{PBR}(I_\mathrm{\mu}, I_\mathrm{n}, I_\mathrm{a}, I_\mathrm{\rho}, I_\mathrm{m})\eqstop
\end{equation}

\paragraph{Optimization Details}

Since the transition from forward shading to deferred shading is not seamless and may introduce artifacts, we perform a slight optimization using a learning rate of 0.001 for 100 steps (approximately 1 minute). This optimization adjusts the displacement of the 3DGS. During this process, we unlock the geometry constraints, allowing the 3DGS to modify its positions, scales, rotations, normals, and opacities. As shown in Table~\ref{app:meshsampler:appref}, this adjustment does not significantly alter these attributes, ensuring that the Gaussian-mesh consistency remains well-preserved.

\begin{table}[h]
    \centering
    \resizebox{0.9\textwidth}{!}{
        \begin{tabular}{cccccc}
            \toprule
            \textbf{Dataset} & positions $\Delta\centerPos$ & scales $\Delta\threeDScaling$ & rotations $\Delta\threeDRotation$ & opacities $\Delta\opacity$ & normals $\Delta\normal$ \\
            \midrule
            TensoIR Synthetic Dataset& 0.0014& 0.0003 & 0.0148 & 0.0131 & 0.0146\\
            Shiny Blender Dataset& 0.0011& 0.0002 & 0.0118 & 0.0168 & 0.0087\\
            \bottomrule
        \end{tabular}
    }
    \caption{We measure the average L1 difference between the unoptimized and optimized attributes to validate that this technique does not significantly degrade geometry quality.}
    \label{app:meshsampler:appref}
\end{table}
\section{Discussion on Mask Requirements}
\label{app:maskanalysis}

Benefiting from the explicit mesh normals and the opaque mesh surface, our {\name} achieves improved light transport modeling compared to previous 3DGS-based inverse rendering approaches. However, as mentioned in Sec.~\ref{sec:method_implement}, optimizing explicit surfaces is quite challenging, and existing isosurface-based approaches~\cite{Munkberg_2022_CVPR, hasselgren2022nvdiffrecmc} typically require an object mask loss for background removal. Similar to these methods, our {\name} also relies on input masks. In this section, we discuss how to mitigate this limitation and adapt our approach for real-world applications in the absence of ground truth masks.

\begin{figure}[h]
	\vspace{-4mm}
	\begin{center}
	\setlength{\tabcolsep}{1pt}
	\setlength{\fboxrule}{1pt}
	\begin{tabular}{c}
		\begin{tabular}{ccc}
			\includegraphics[width=0.32\linewidth,trim={4cm 0cm 4cm 3cm},clip]{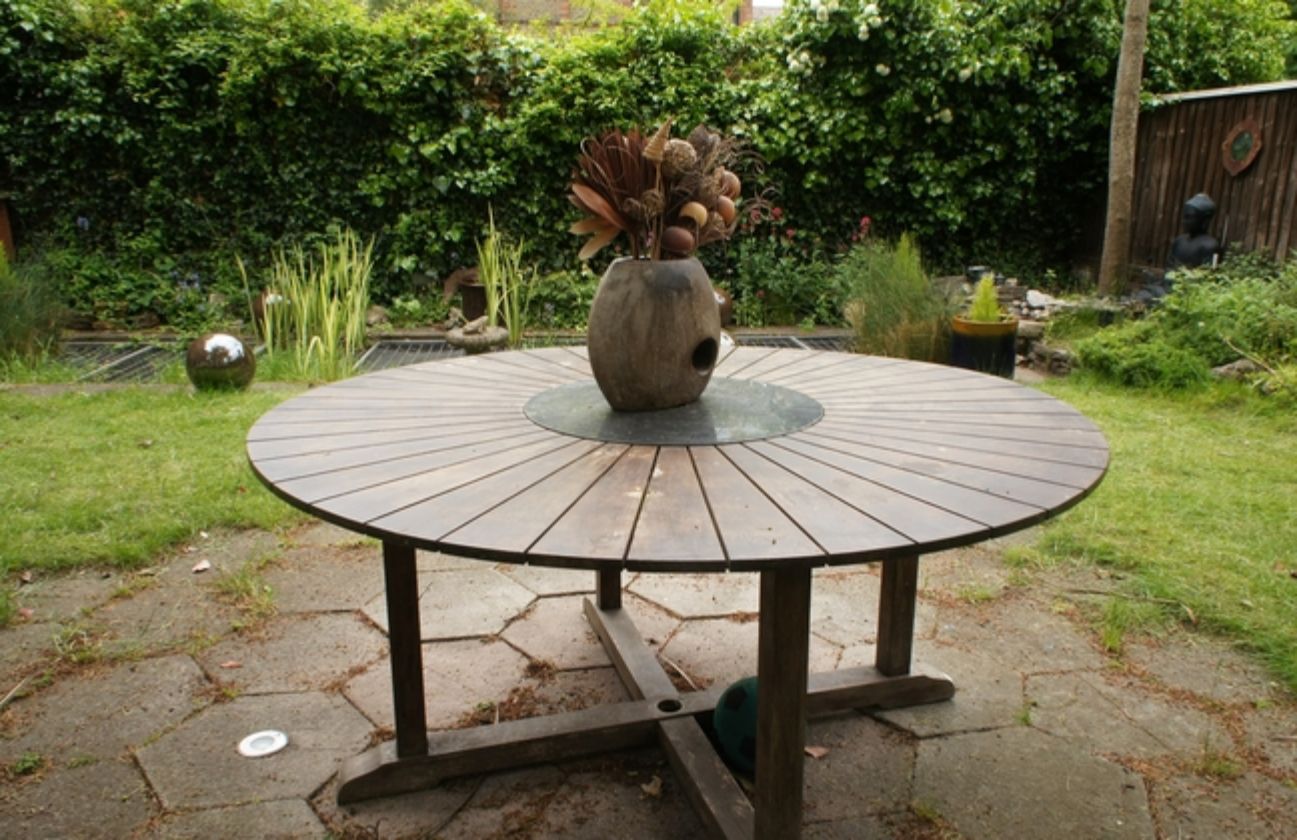} &
			\includegraphics[width=0.32\linewidth,trim={4cm 0cm 4cm 3cm},clip]{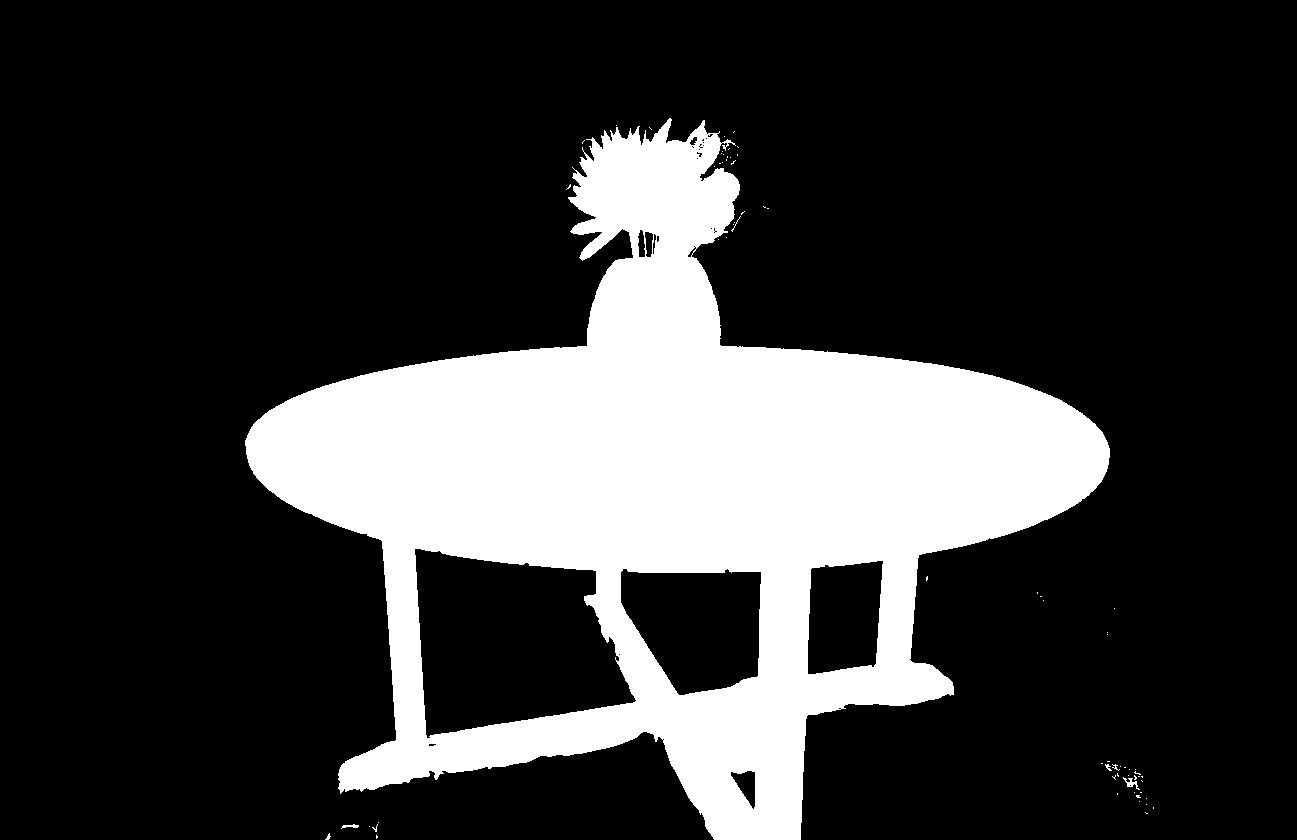} &
			~\includegraphics[width=0.32\linewidth,trim={4cm 0cm 4cm 3cm},clip]{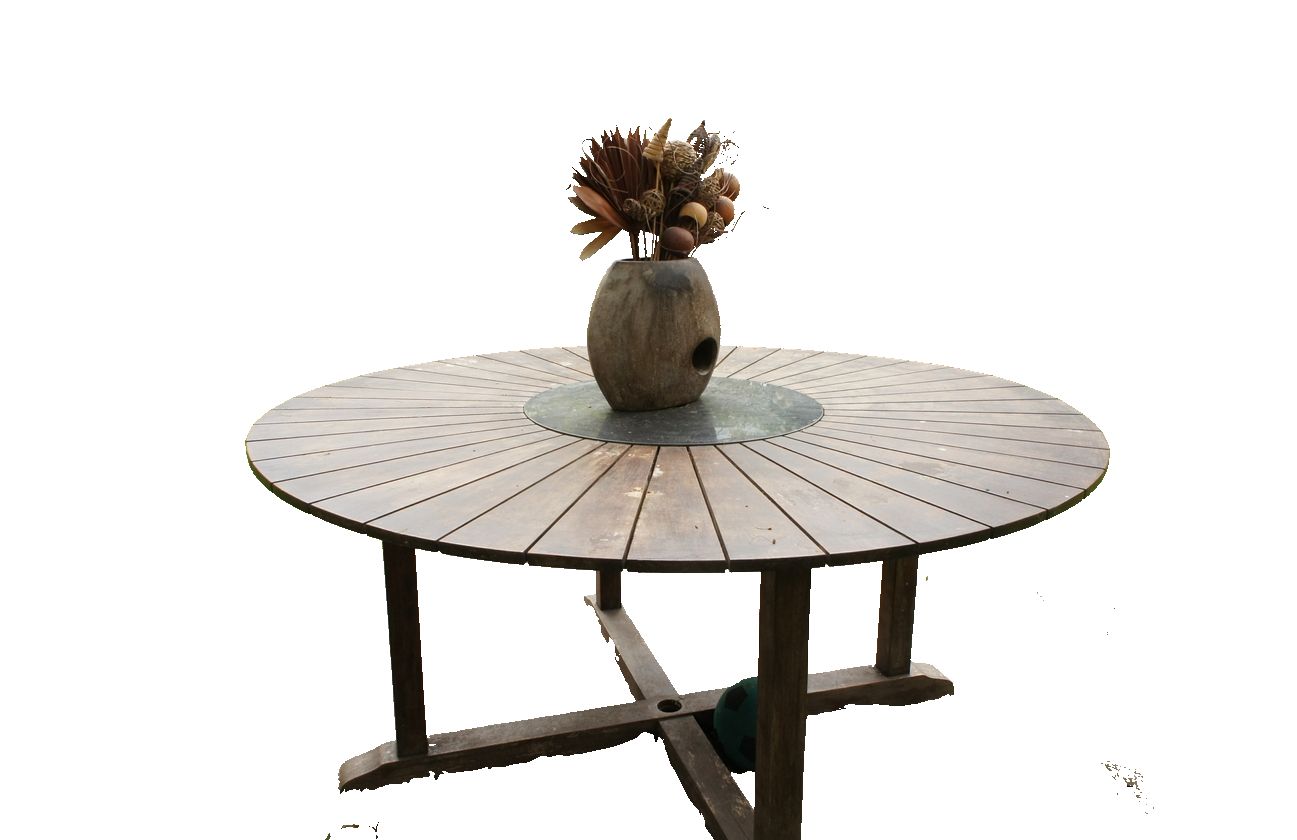} \\
			{(a) Original Capture} &
			{(b) Binary Mask} &
			{(c) Masked Capture}
		\end{tabular}
	\end{tabular}
	\end{center}
	\vspace{-4mm}
	\caption{ \label{fig:garden:mask}
		Coarse mask estimated on \textit{Garden} from the Mip-NeRF 360 Dataset~\cite{barron2022mip}.
	}
	\vspace{-1mm}
\end{figure}

\begin{figure}[h]
	\begin{center}
	\setlength{\tabcolsep}{1pt}
	\setlength{\fboxrule}{1pt}
	\begin{tabular}{c}
		\begin{tabular}{ccc}
			\includegraphics[width=0.32\linewidth,trim={4cm 0cm 4cm 3cm},clip]{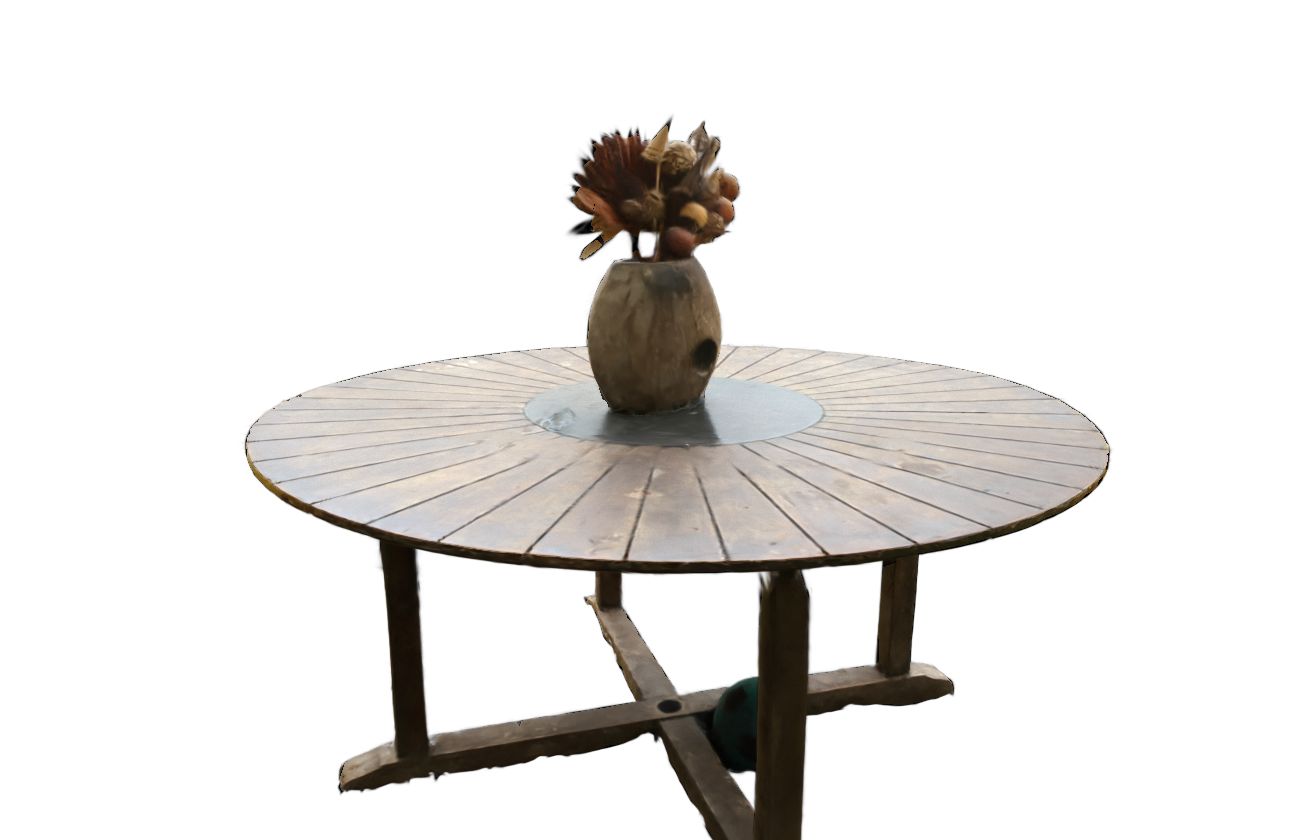} &
			\includegraphics[width=0.32\linewidth,trim={4cm 0cm 4cm 3cm},clip]{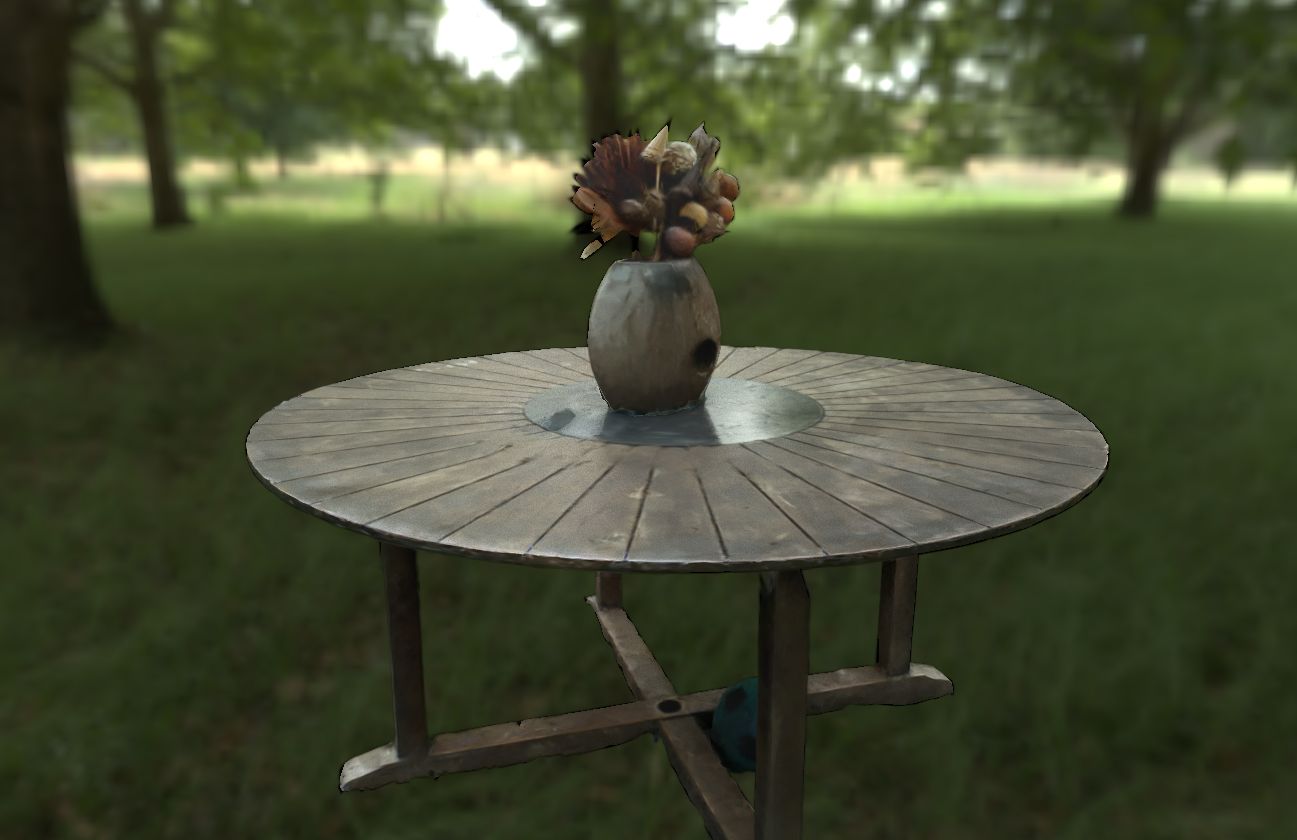} &
			~\includegraphics[width=0.32\linewidth,trim={4cm 0cm 4cm 3cm},clip]{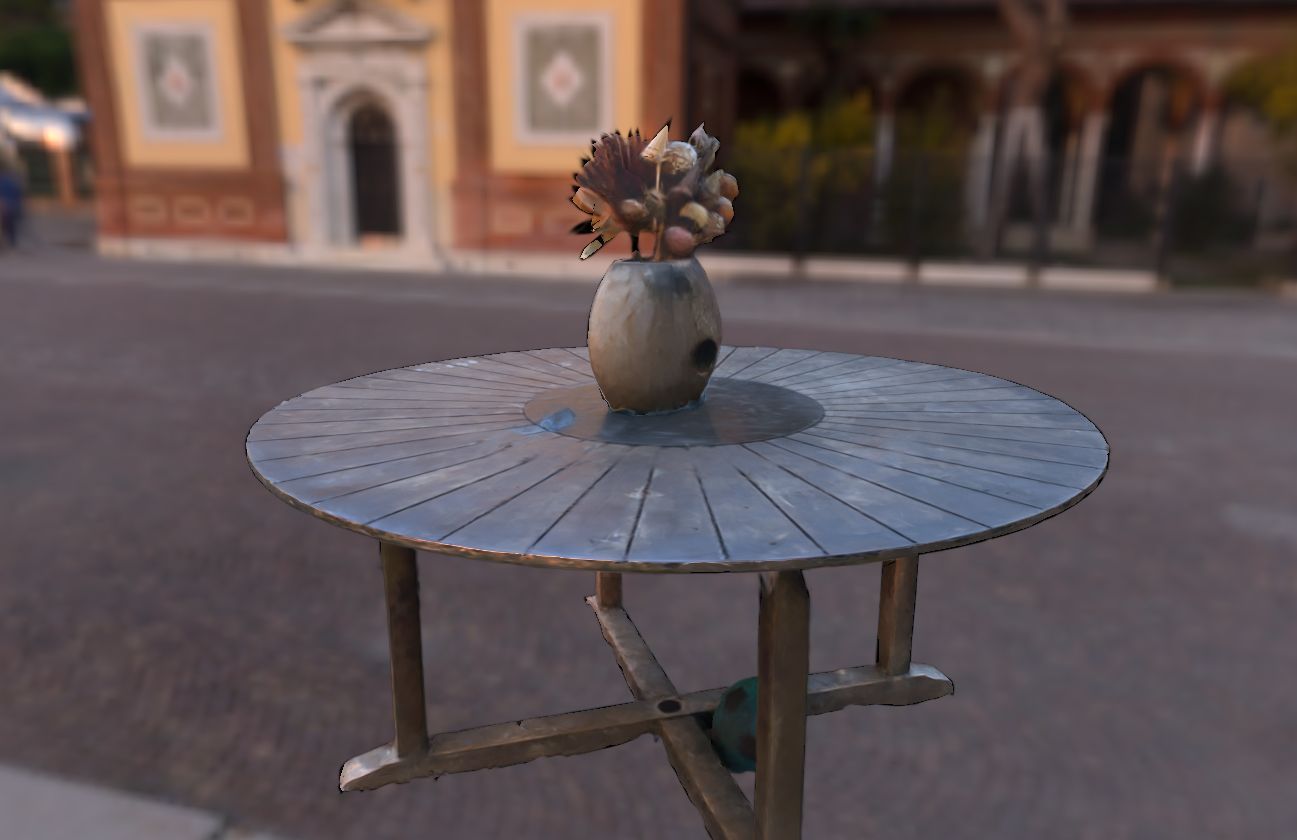} \\
            {(a) NVS} &
			{(b) Relit 1} &
			{(c) Relit 2} \\
            \includegraphics[width=0.32\linewidth,trim={4cm 0cm 4cm 3cm},clip]{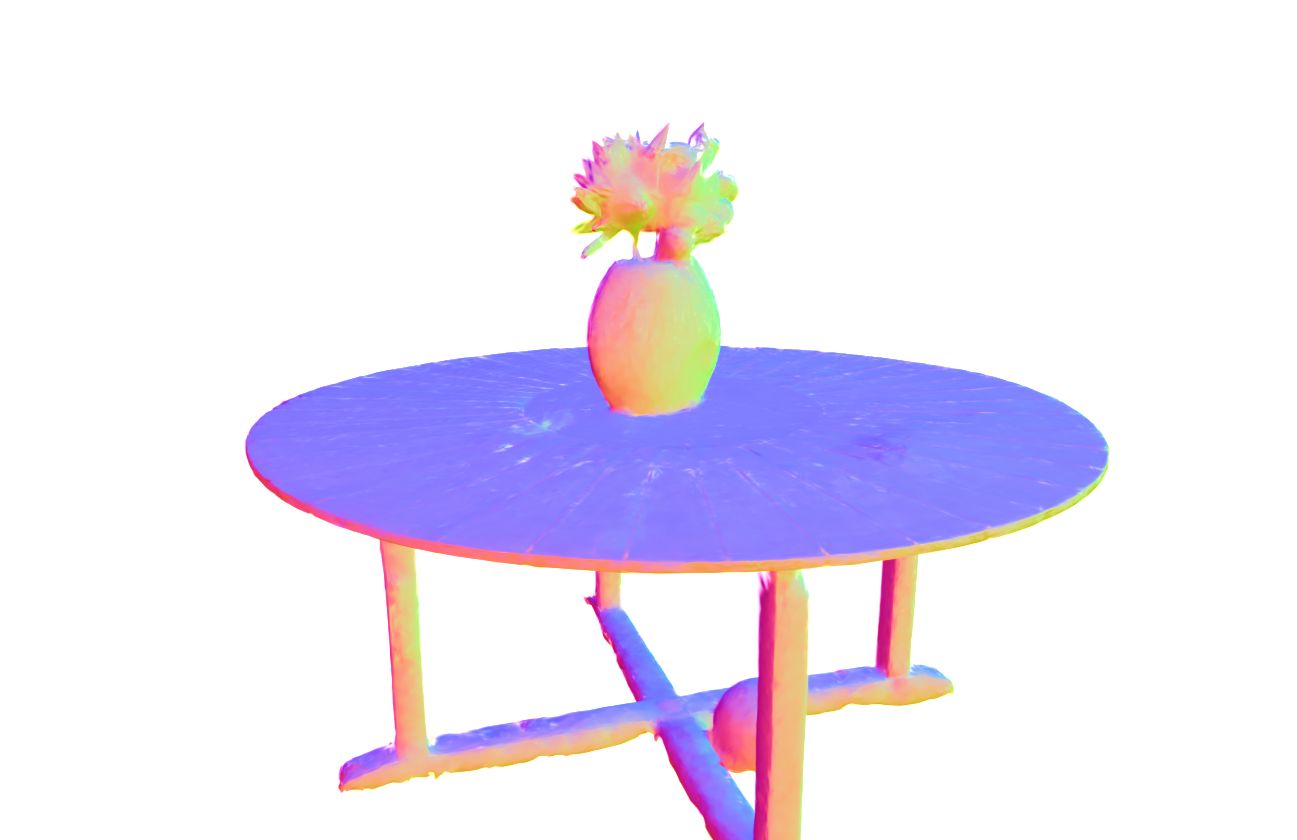} &
			\includegraphics[width=0.32\linewidth,trim={4cm 0cm 4cm 3cm},clip]{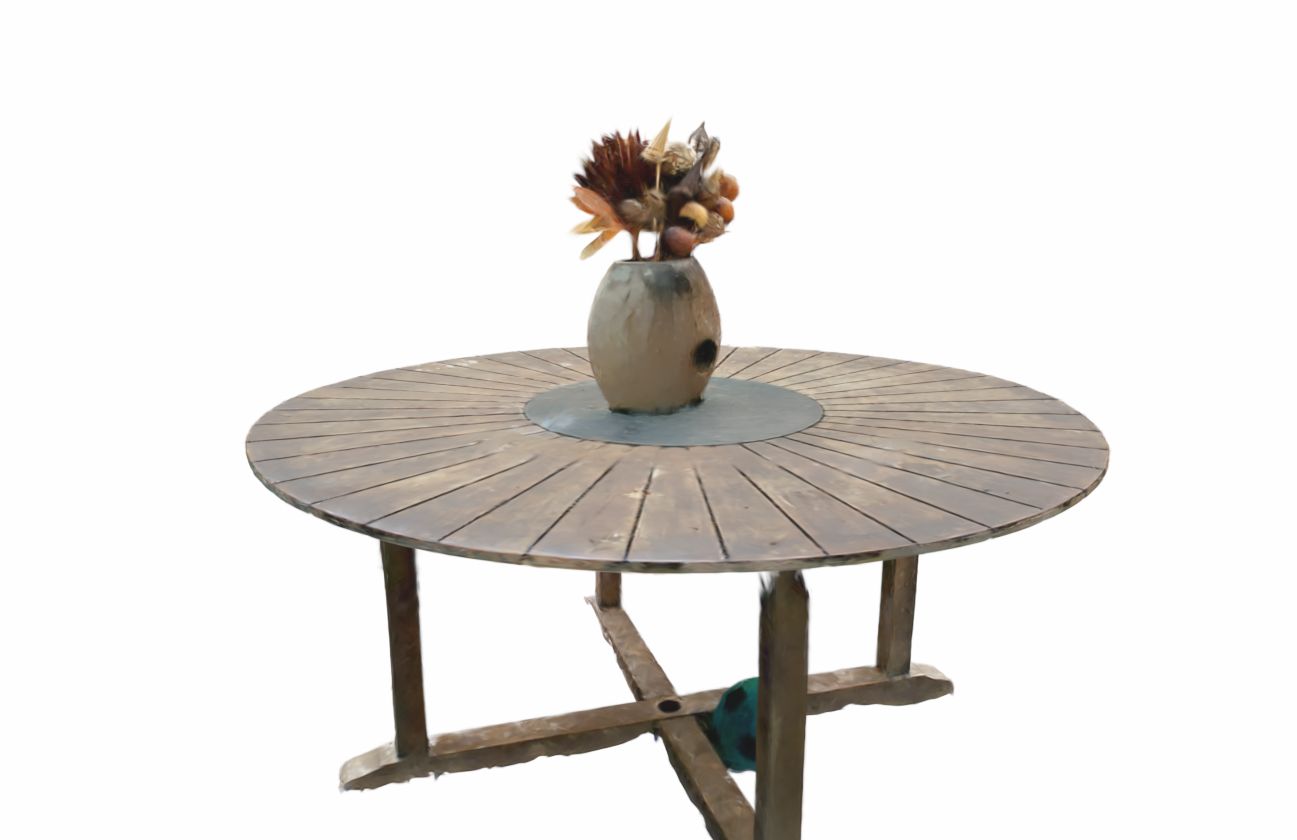} &
			~\includegraphics[width=0.32\linewidth,trim={4cm 0cm 4cm 3cm},clip]{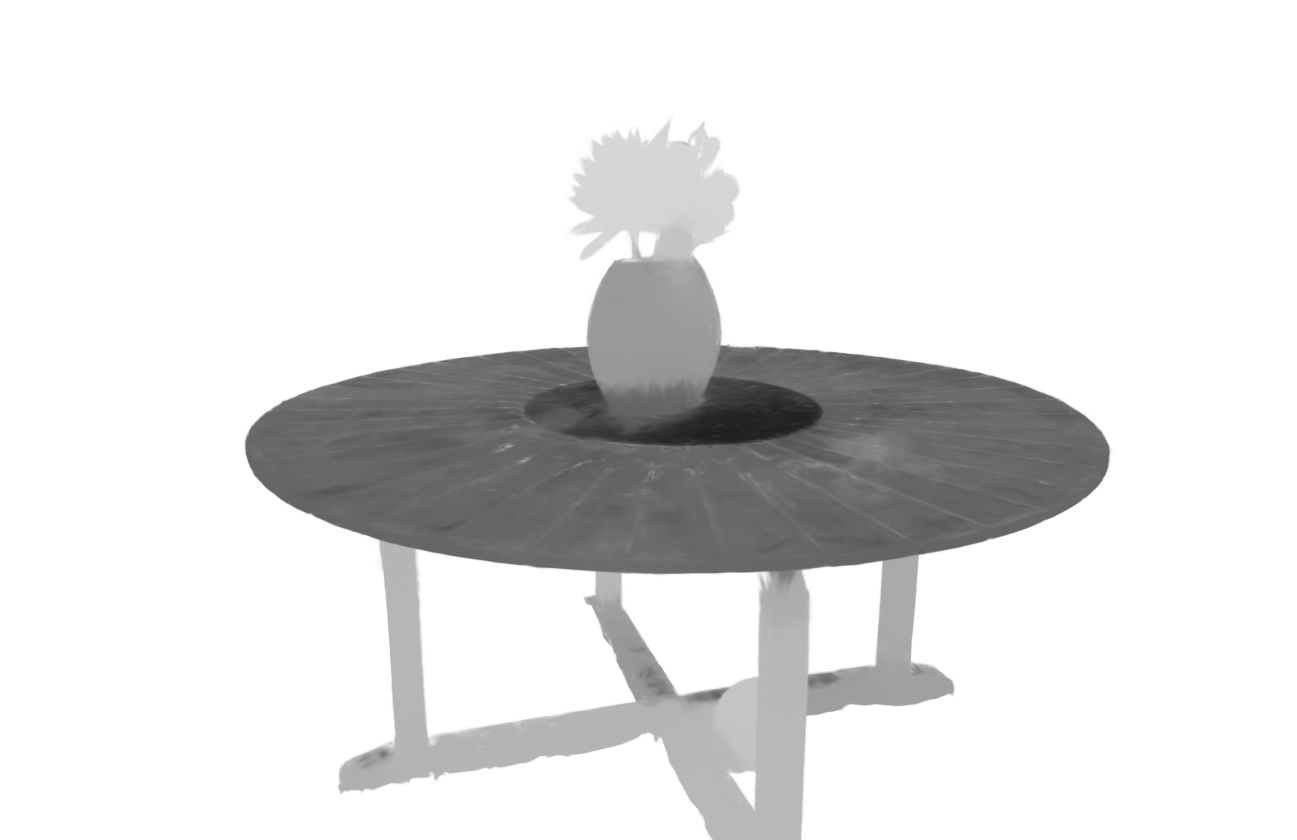} \\
			{(d) Normal} &
			{(e) Albedo} &
			{(f) Roughness} \\
		\end{tabular}
	\end{tabular}
	\end{center}
	\vspace{-6mm}
	\caption{ \label{fig:garden:decomp}
		Decomposition and relighting results of \textit{Garden} from the Mip-NeRF 360 Dataset~\cite{barron2022mip}.
	}
	\vspace{-3mm}
\end{figure}

As analyzed in prior work~\cite{Munkberg_2022_CVPR}, isosurface techniques are not highly sensitive to mask quality, and a coarse input mask is sufficient. Therefore, we use vanilla 3DGS to develop a simple solution that effectively approximates object masks for wild captures. Specifically, given a set of multi-view captures, we first train a standard 3DGS for only 5000 steps (2-4 minutes) to obtain Gaussian points. Then, we simply choose a bounding box and clip the Gaussian points outside the bounding box. As illustrated in Fig.~\ref{fig:garden:mask}, the rendering results of the remaining Gaussian points naturally form a coarse mask, which can be used for background removal. Based on this mask estimation trick, our {\name} can be applied to object decomposition for wild captures. As shown in Fig.~\ref{fig:garden:decomp}, we demonstrate the decomposition and relighting results for the \textit{Garden} scene.

\section{Analysis of Resolution}
\label{app:resolution}

\begin{figure}[h]
	\vspace{-2mm}
	\begin{center}
	\setlength{\tabcolsep}{1pt}
	\setlength{\fboxrule}{1pt}
    \resizebox{\textwidth}{!}{
	\begin{tabular}{c}
		\begin{tabular}{cccc}
            \rotatebox{90}{\Large{~~~~~$96^3$}} &
			\includegraphics[width=0.3\linewidth,trim={0cm 4cm 0cm 1cm},clip]{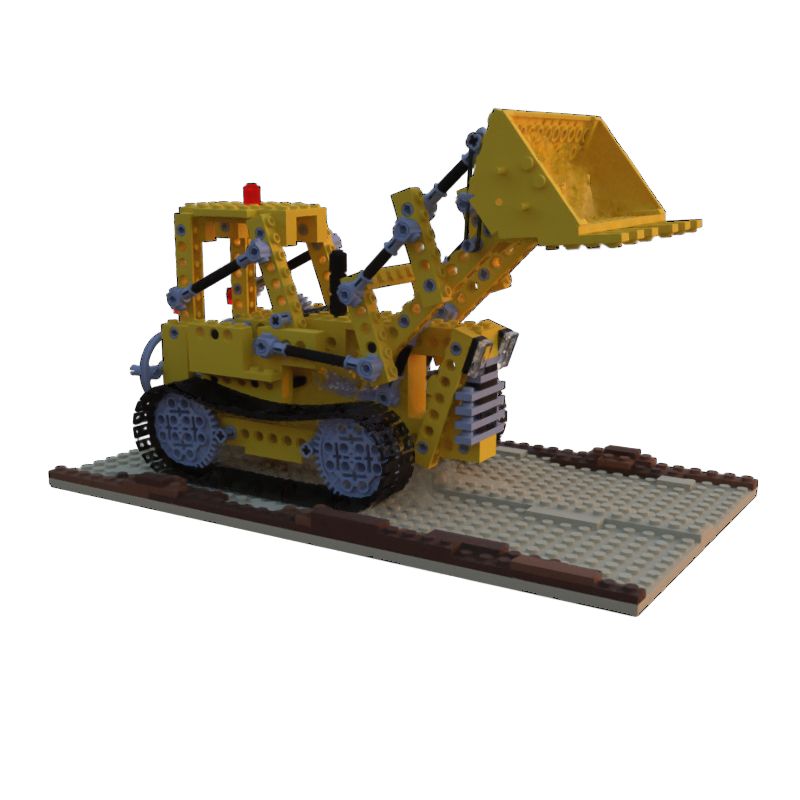} &
			\includegraphics[width=0.3\linewidth,trim={0cm 4cm 0cm 1cm},clip]{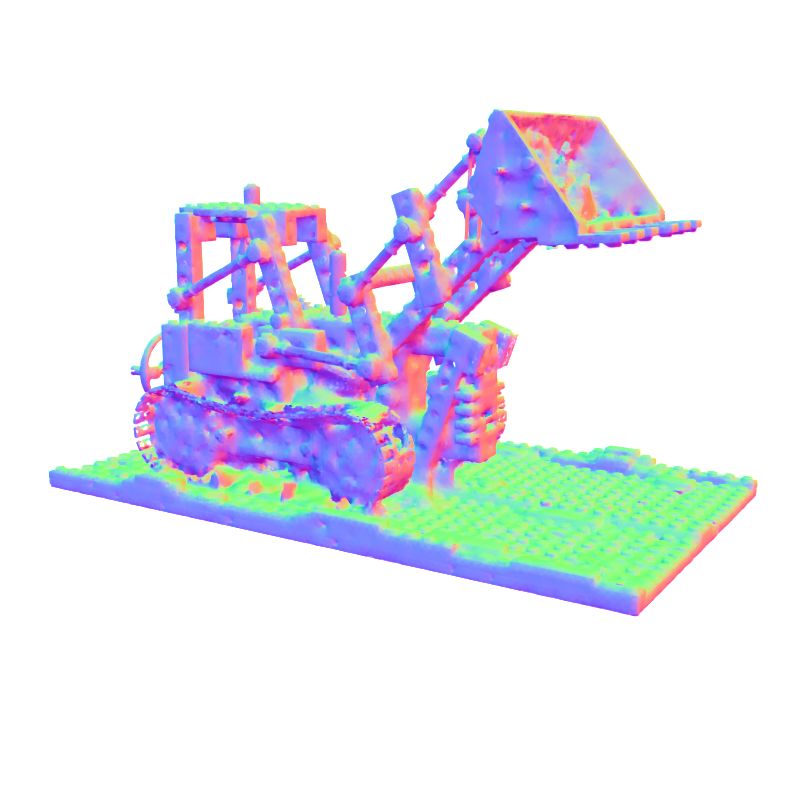} &
			~\includegraphics[width=0.3\linewidth,trim={0cm 4cm 0cm 1cm},clip]{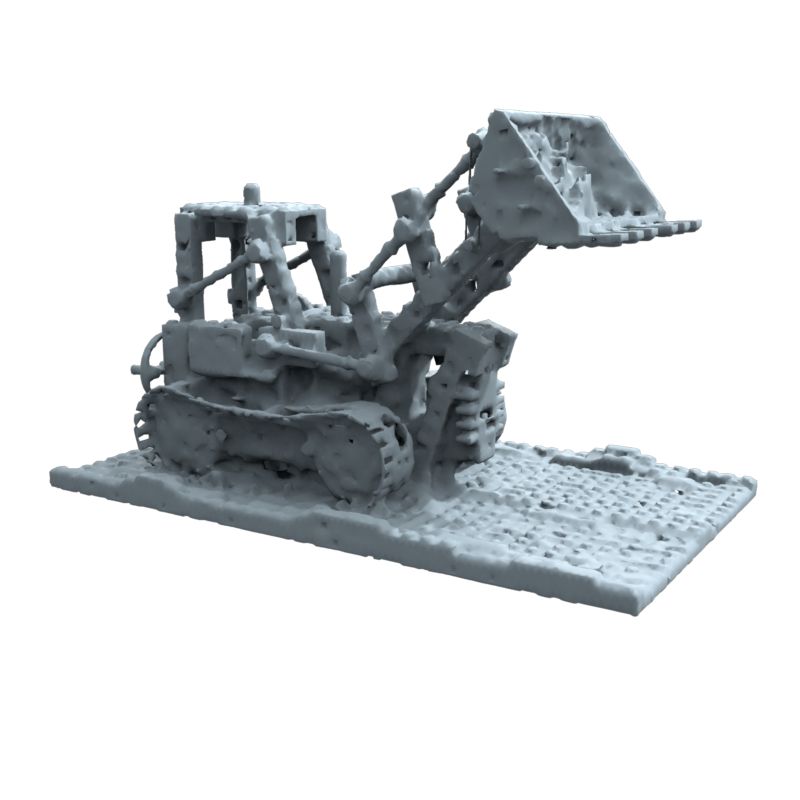} \\
            \rotatebox{90}{\Large{~~~~~$128^3$}} &
            \includegraphics[width=0.3\linewidth,trim={0cm 4cm 0cm 1cm},clip]{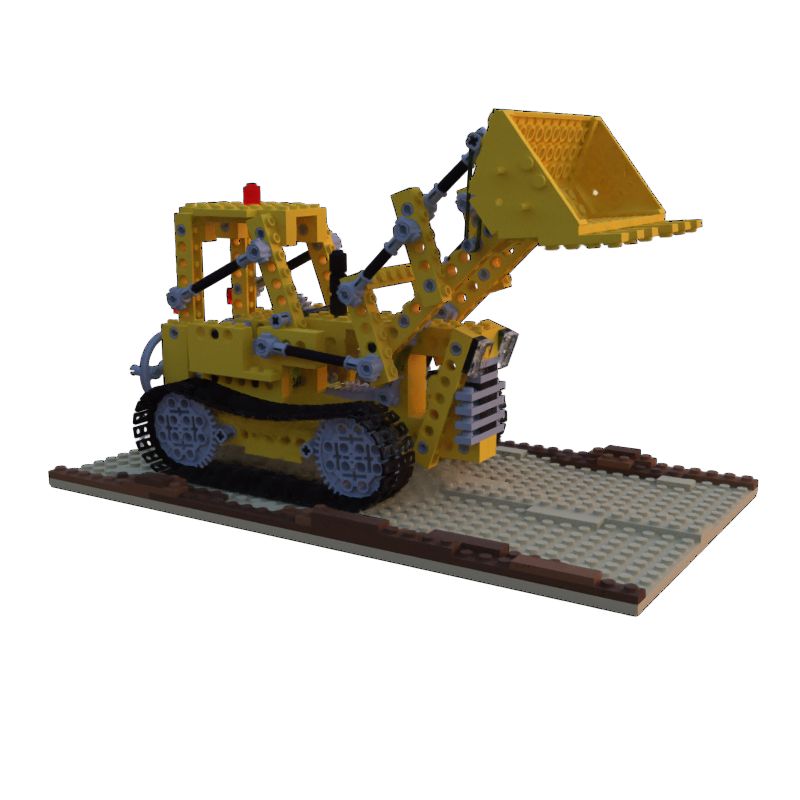} &
			\includegraphics[width=0.3\linewidth,trim={0cm 4cm 0cm 1cm},clip]{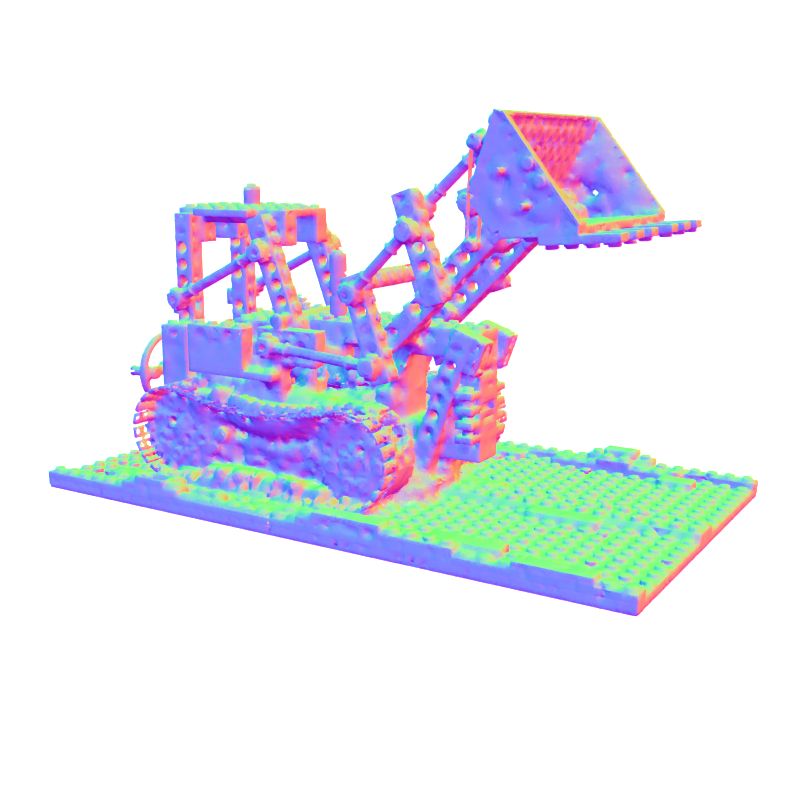} &
			~\includegraphics[width=0.3\linewidth,trim={0cm 4cm 0cm 1cm},clip]{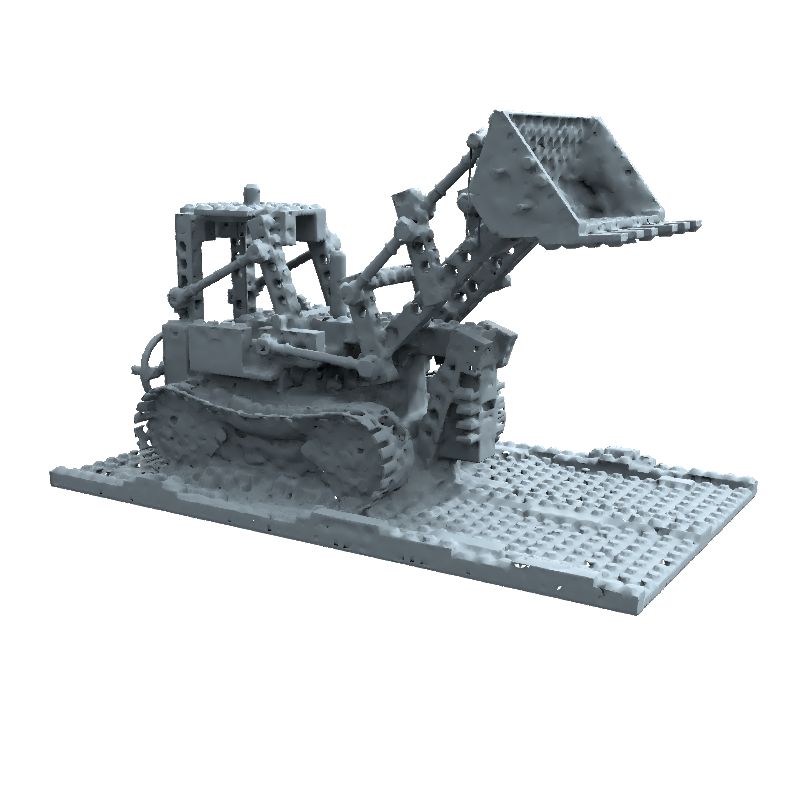} \\
            &
			{Render} &
			{Normal} &
			{Mesh Guidance} \\
		\end{tabular}
	\end{tabular}
    }
	\end{center}
	\vspace{-4mm}
	\caption{ \label{fig:app:resoluion}
		Resolution comparisons of \textit{Lego} from the TensoIR Synthetic Dataset.
	}
	\vspace{-2mm}
\end{figure}

As discussed in Sec.~\ref{sec:conclusion}, one of the most significant limitations of our {\name} is the resolution constraint. Due to the limited grid resolution of the isosurface technique (FlexiCubes), {\name} struggles with thin structures and surfaces featuring complicated geometry. This issue can be mitigated by simply increasing the FlexiCubes resolution. We present a comparison in Fig.~\ref{fig:app:resoluion} to demonstrate the impact of different resolutions on the recovery of thin geometry ($R=96$ and $R=128$, respectively).

However, due to the space and time complexity of $O(R^3)$, the optimization process with $R=96$ can be easily conducted on a single NVIDIA RTX 4090 (24GB CUDA memory) within 15-20 minutes. In contrast, the $R=128$ setting typically takes more than 1 hour to train and requires more than 60GB of CUDA memory. Therefore, in Sec.~4, when presenting experimental comparisons, we use $R=72$ for simple shapes and $R=96$ for others. Furthermore, this resolution limitation also restricts our method from adapting to scene-level decomposition. A promising direction for future work is to explore how adaptive resolution techniques could be applied to accommodate detailed geometry, enabling the extension of our approach to scene-level tasks.
\section{Optimization with An Existing Mesh}
\label{app:imesh}

While the underlying geometry is modeled from scratch using isosurfaces in the standard pipeline, {\name} also supports initializing the optimization process with an existing mesh. For instance, given multi-view images, one can first extract a triangle mesh using methods like NeuS~\cite{wang2021neus} or GOF~\cite{Yu2024GOF}. Subsequently, {\name} can leverage this initial mesh by converting it into Gaussian points via {\meshsampler}. The remaining steps then follow the standard {\name} pipeline.

\begin{figure}[h]
\vspace{-2mm}
\centering
\setlength{\tabcolsep}{1pt}
\setlength{\fboxrule}{1pt}
\resizebox{\linewidth}{!}{
\begin{tabular}{ccc}
  \vspace{-0.5mm}
  \includegraphics[width=0.4\linewidth,trim={0cm 0.5cm 2.5cm 0cm},clip]{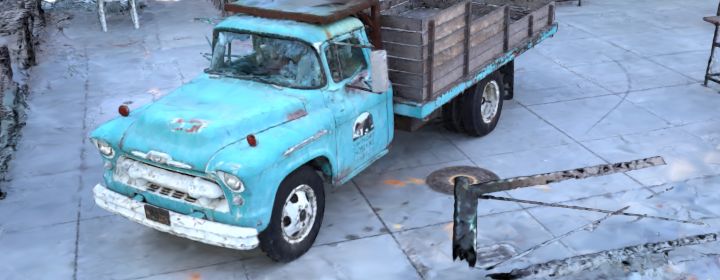} &
  \includegraphics[width=0.4\linewidth,trim={0cm 0.5cm 2.5cm 0cm},clip]{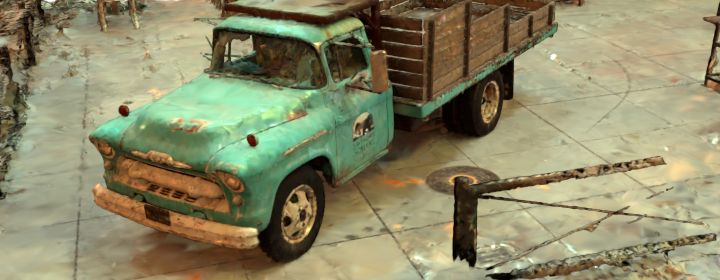} &
  \includegraphics[width=0.4\linewidth,trim={0cm 0.5cm 2.5cm 0cm},clip]{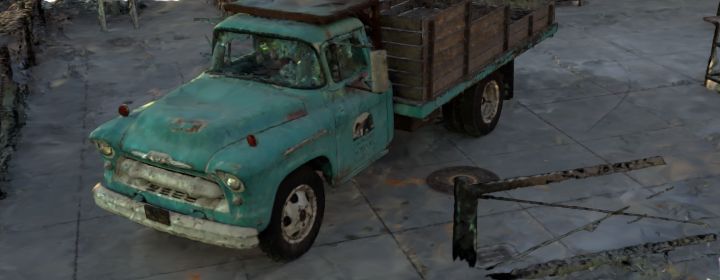} \\
  \includegraphics[width=0.4\linewidth,trim={2cm 0.5cm 0.5cm 0cm},clip]{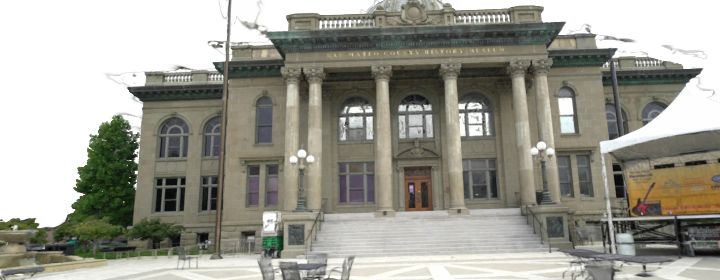} &
  \includegraphics[width=0.4\linewidth,trim={2cm 0.5cm 0.5cm 0cm},clip]{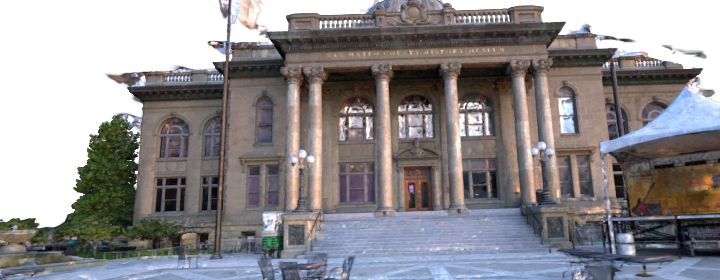} &
  \includegraphics[width=0.4\linewidth,trim={2cm 0.5cm 0.5cm 0cm},clip]{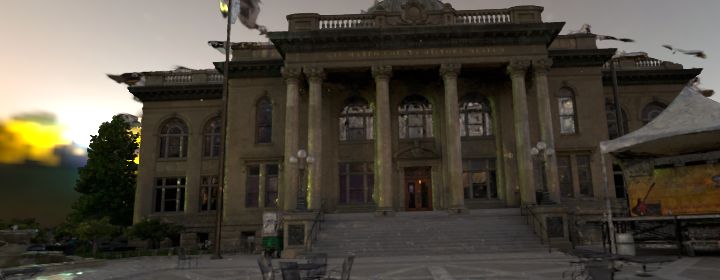} \\
  \textit{NVS} & \textit{Relit 1} & \textit{Relit 2}
\end{tabular}
}
\vspace{-2mm}
\caption{Extending {\name} to scene levels by providing initial mesh.}
\label{fig:app:scene}
\vspace{-4mm}
\end{figure}

The differentiable nature of {\meshsampler} enables the refinement of the initial mesh through gradients derived from 3DGS rendering. A significant advantage of this approach is the ability to circumvent the limitations associated with isosurfaces, such as restricted resolution and mask dependencies. This capability inherently extends {\name}'s applicability to scene-level inverse rendering, with illustrative results presented in Figure~\ref{fig:app:scene}.

It is important to note, however, that this two-stage extension is not applicable to specular objects. This limitation arises because surface reconstruction methods typically lack effective PBR modeling, struggling to extract accurate geometry for specular objects and leading to noisy meshes. As a result, {\name} is unable to rectify geometric inaccuracies from the previous stage. Actually, the superior performance of {\name} in reflective scenarios, as demonstrated in Figure~\ref{fig:exp:normal} of the main paper, is primarily attributable to its joint optimization of geometry and BRDF material.

\section{Path Tracing Adaption}
\label{app:pt}

\begin{figure}[h]
\centering
\small
\setlength{\tabcolsep}{1pt}
\setlength{\fboxrule}{1pt}
\begin{tabular}{ccccc}
  \vspace{-1mm}
  \includegraphics[width=0.19\linewidth,trim={5cm 9cm 4cm 3cm},clip]{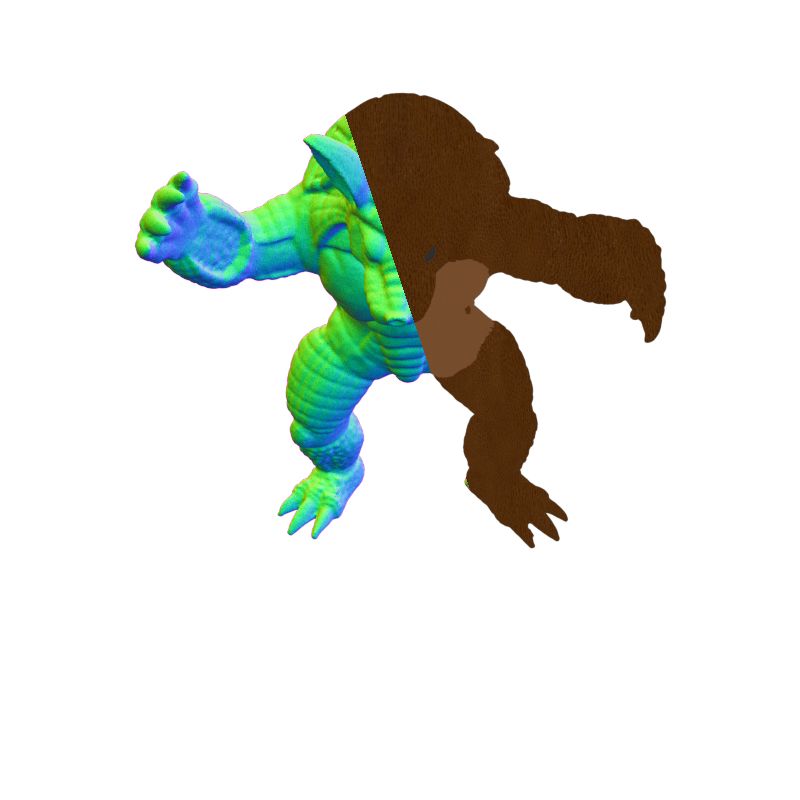} &
  \includegraphics[width=0.19\linewidth,trim={5cm 9cm 4cm 3cm},clip]{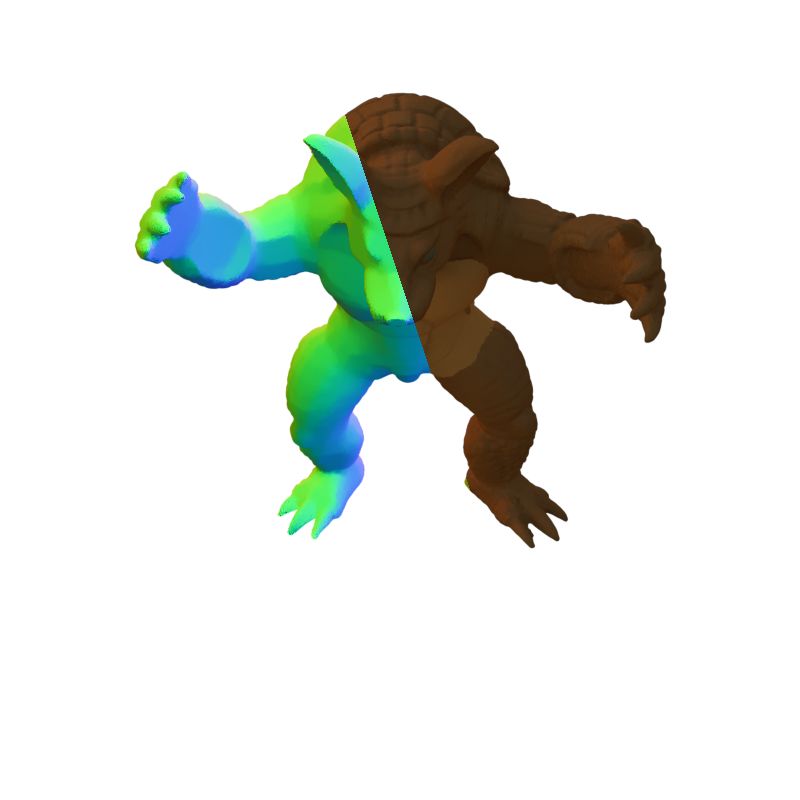} &
  \includegraphics[width=0.19\linewidth,trim={5cm 9cm 4cm 3cm},clip]{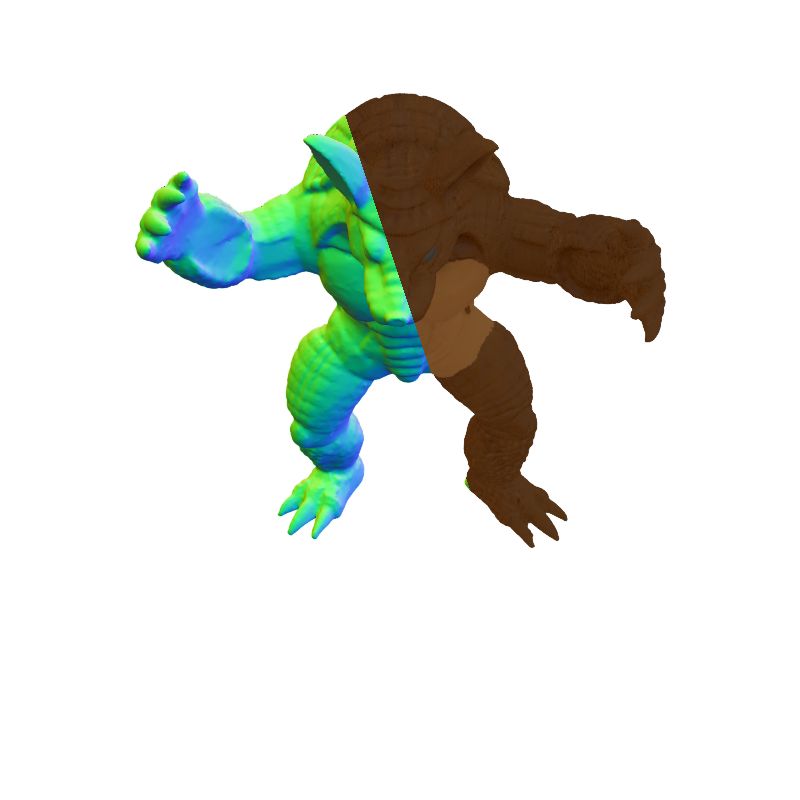} &
  \includegraphics[width=0.19\linewidth,trim={5cm 9cm 4cm 3cm},clip]{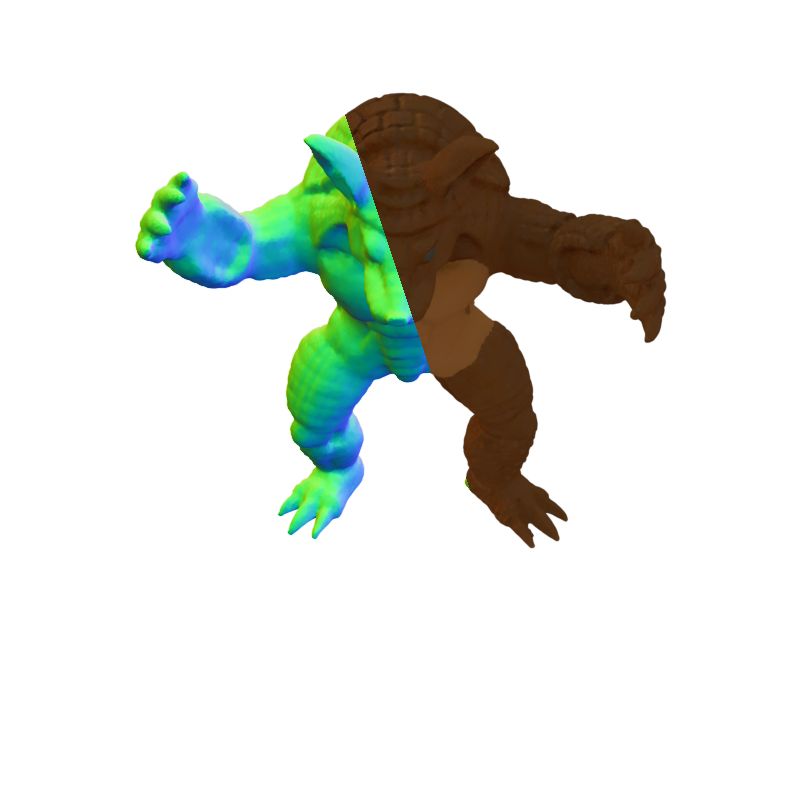} &
  \includegraphics[width=0.19\linewidth,trim={5cm 9cm 4cm 3cm},clip]{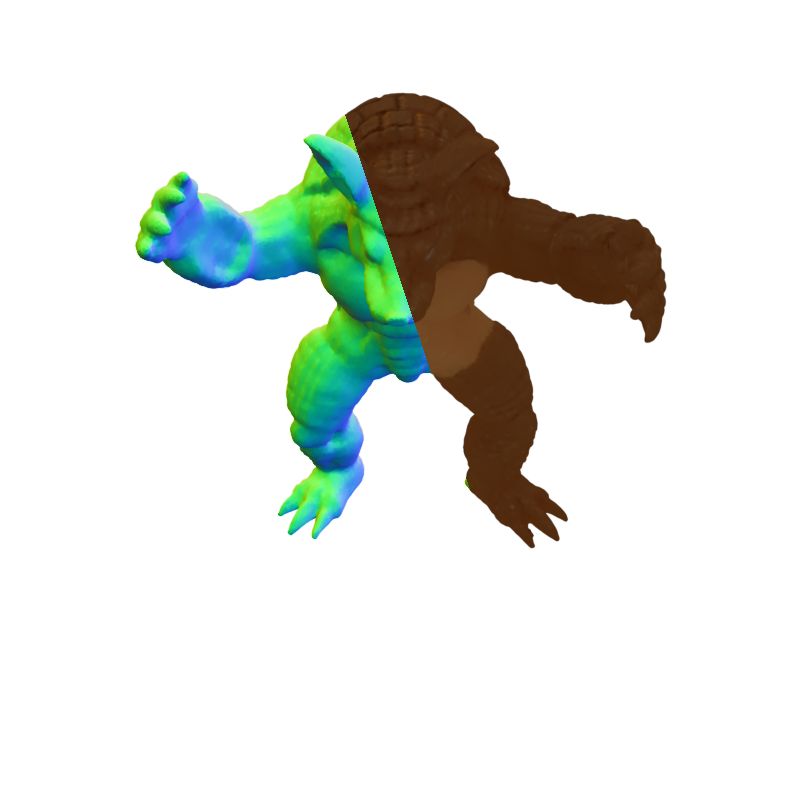} \\
  \vspace{-1mm}
  \includegraphics[width=0.19\linewidth,trim={4cm 4cm 4cm 2cm},clip]{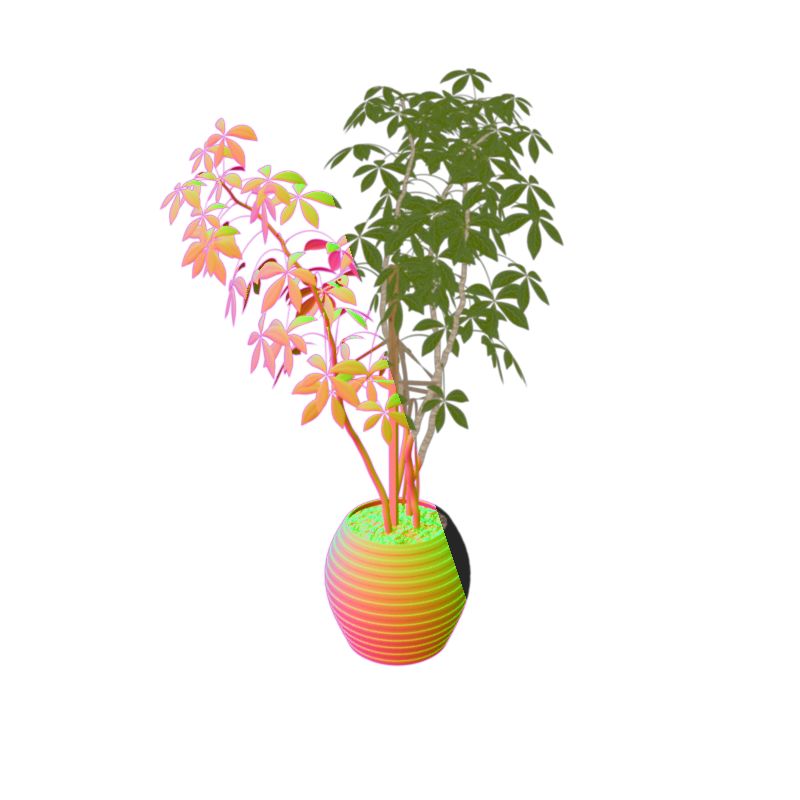} &
  \includegraphics[width=0.19\linewidth,trim={4cm 4cm 4cm 2cm},clip]{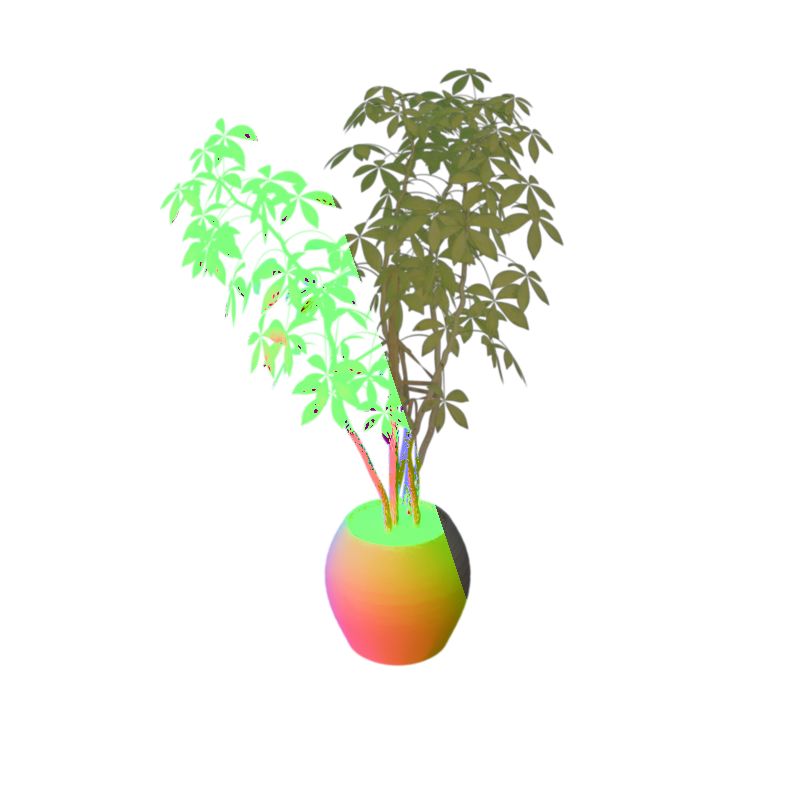} &
  \includegraphics[width=0.19\linewidth,trim={4cm 4cm 4cm 2cm},clip]{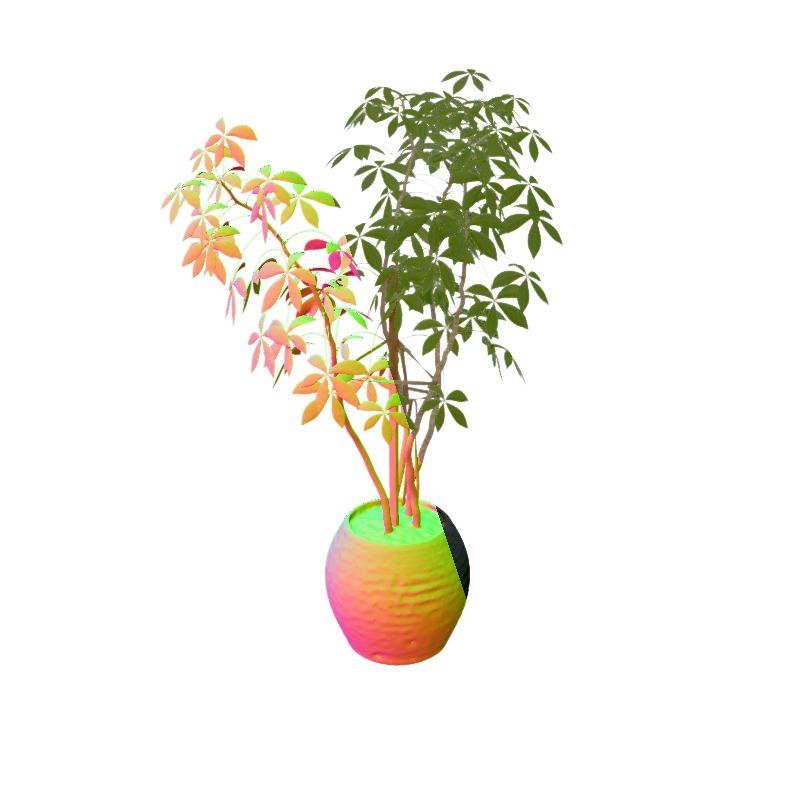} &
  \includegraphics[width=0.19\linewidth,trim={4cm 4cm 4cm 2cm},clip]{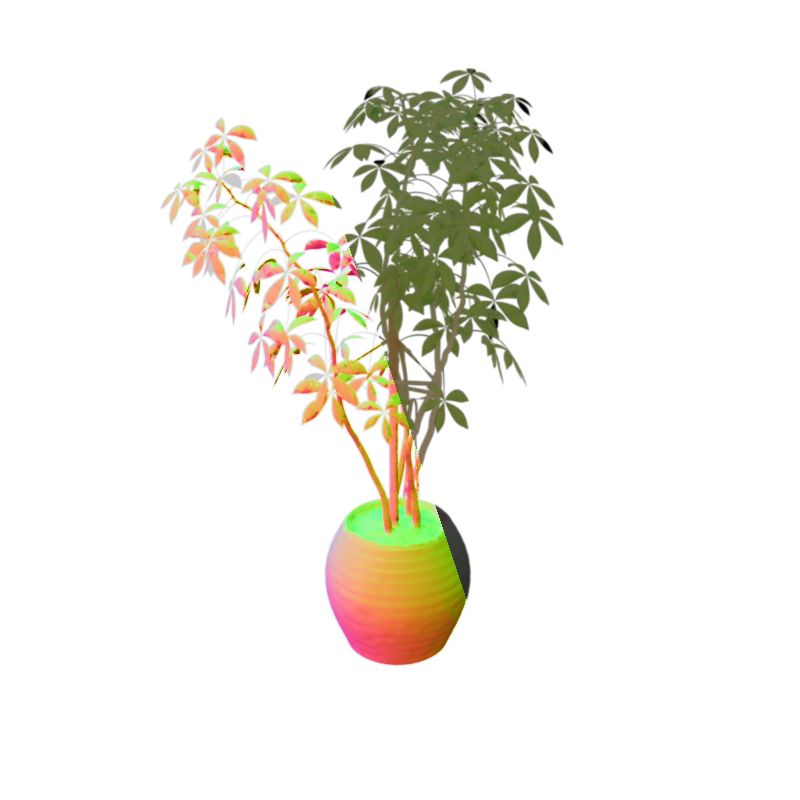} &
  \includegraphics[width=0.19\linewidth,trim={4cm 4cm 4cm 2cm},clip]{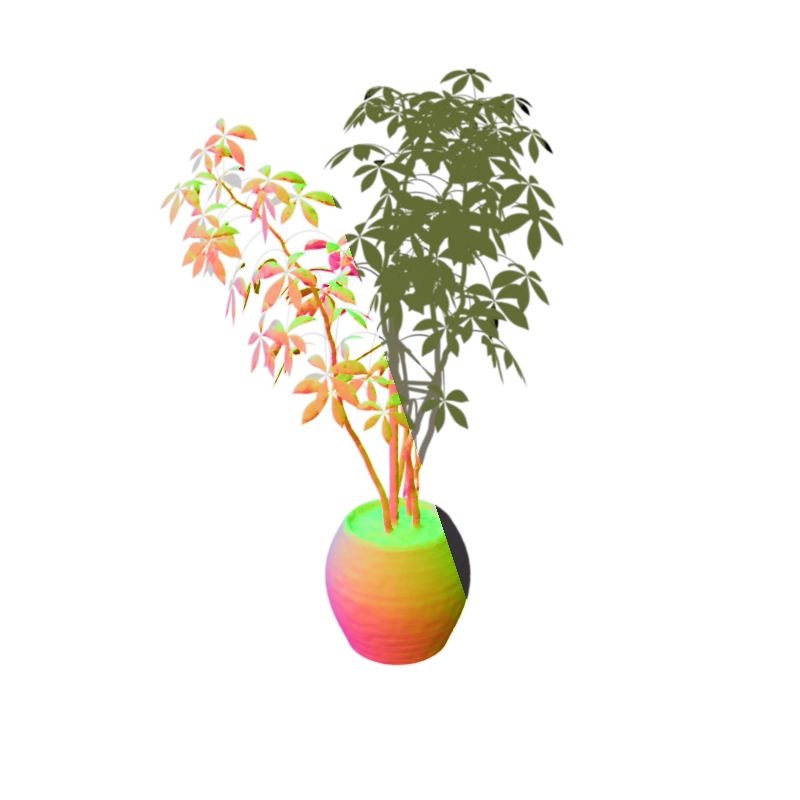} \\
  \vspace{-1mm}
  \includegraphics[width=0.19\linewidth,trim={1cm 8cm 2cm 4cm},clip]{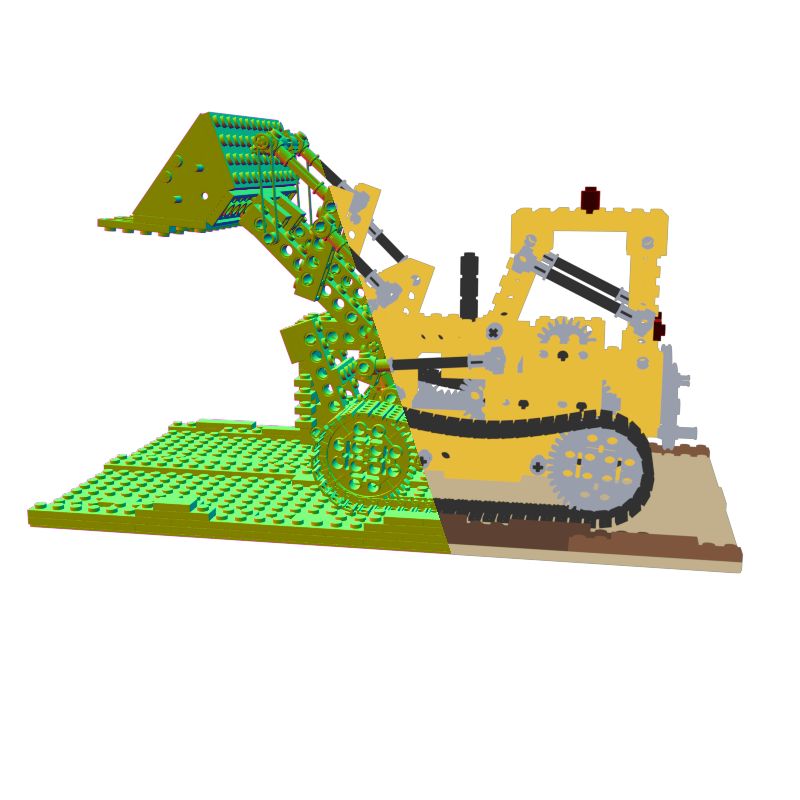} &
  \includegraphics[width=0.19\linewidth,trim={1cm 8cm 2cm 4cm},clip]{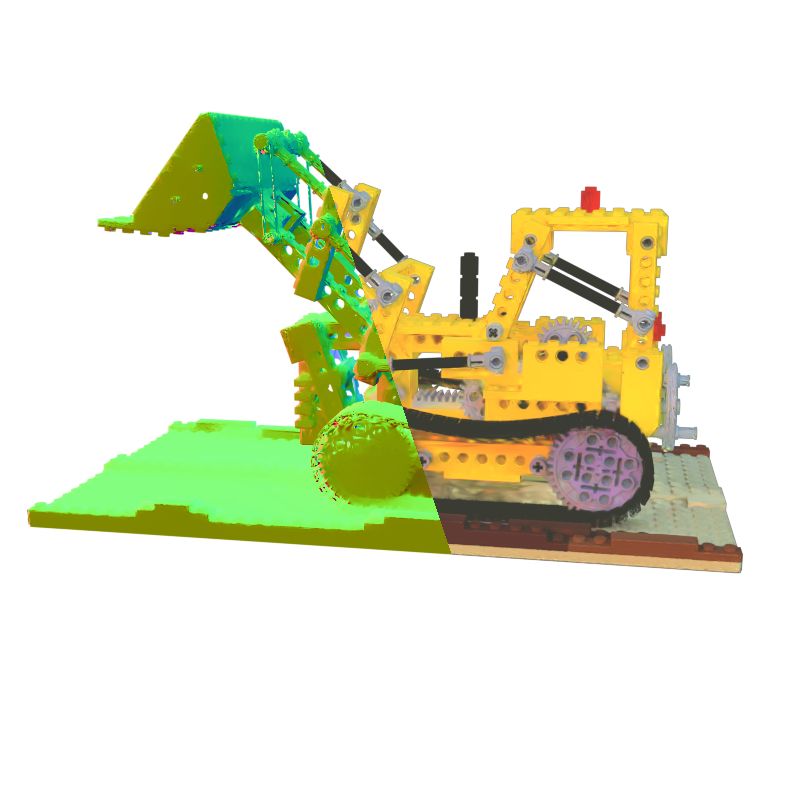} &
  \includegraphics[width=0.19\linewidth,trim={1cm 8cm 2cm 4cm},clip]{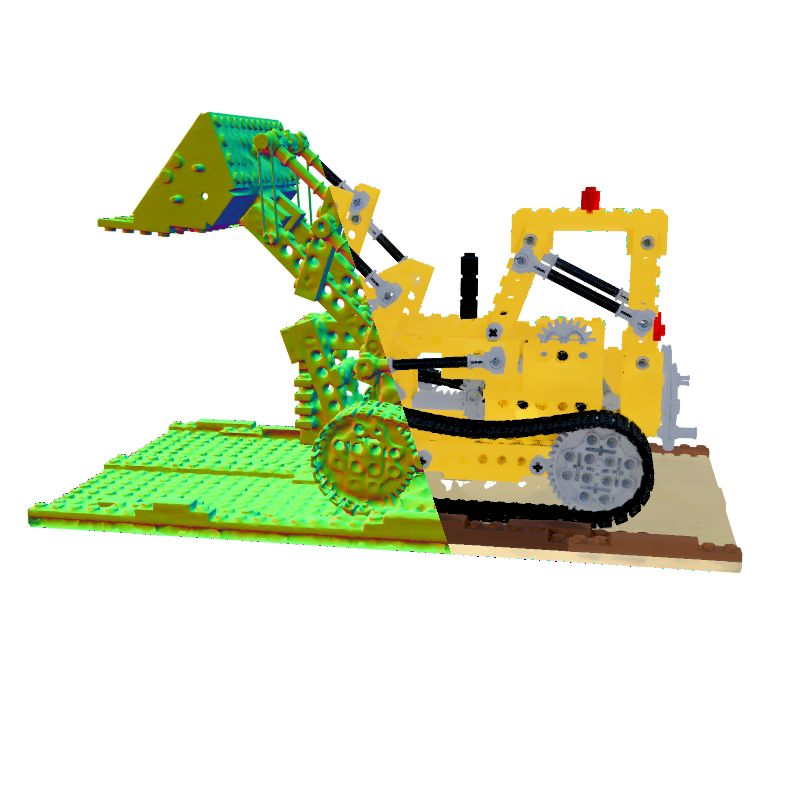} &
  \includegraphics[width=0.19\linewidth,trim={1cm 8cm 2cm 4cm},clip]{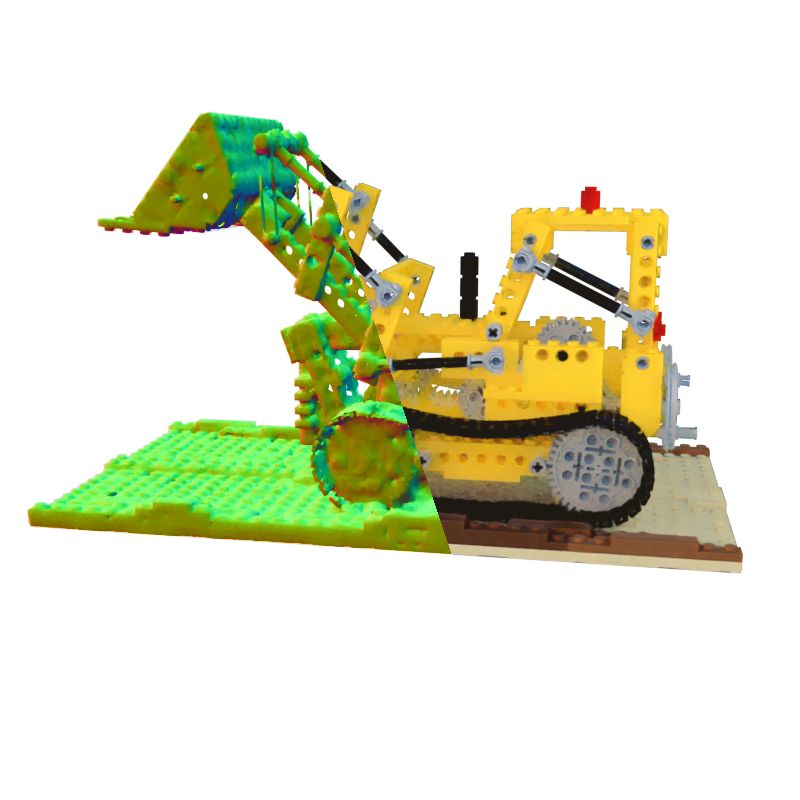} &
  \includegraphics[width=0.19\linewidth,trim={1cm 8cm 2cm 4cm},clip]{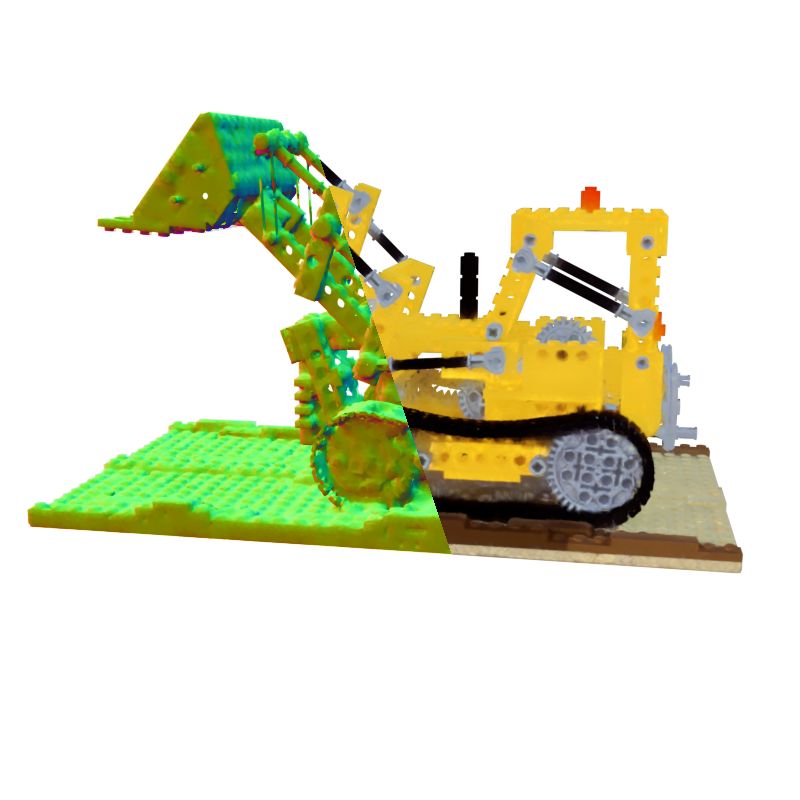} \\
  \textit{GT} & \textit{R3DG} & \textit{MIRRES} & \textit{Ours} & \textit{Ours+PT}
\end{tabular}
\vspace{-2mm}
\caption{\textbf{Decomposition Results.} Left: Normal; Right: Albedo.}
\label{fig:app:restir:qual}
\vspace{-2.5mm}
\end{figure}

In the standard {\name} pipeline, we follow NVDiffrecmc~\cite{hasselgren2022nvdiffrecmc} to use one-bounce ray tracing to evaluate the rendering equation, leaving indirect lighting terms represented as learnable SH coefficents. While this is effective, it can be also improved by incorporating path tracing for physically correct indirect lighting. Following MIRRES~\cite{dai2025mirres}, we incorporate ReSTIR~\cite{bitterli2020spatiotemporal} to conduct efficient path tracing. As shown in Table~\ref{tbl:app:restir:quan} and Figure~\ref{fig:app:restir:qual}, this improves performance at the cost of ~3x slower training speed.

\begin{table}[h]
  \centering
  \setlength{\tabcolsep}{3pt}
  \setlength{\fboxrule}{1pt}
  \begin{tabular}{r|ccccc}
  \toprule
  PSNR $\uparrow$ & Ours & Ours+PT & R3DG~\cite{R3DG2023} & MIRRES~\cite{dai2025mirres}\\
  \midrule
  \textit{TensoIR Synthetic} Albedo & 29.45 & \textbf{29.97} & 28.74 & 29.24 \\
  \textit{TensoIR Synthetic} Relit & 29.59 & 30.04 & 28.55 & \textbf{31.44} \\
  \textit{Shiny Blender} PBR & 31.46 & \textbf{31.60} & 28.83 & 26.53 \\
  \midrule
  Training Time & $\sim$\textbf{14min} % & $\sim$27min
  & $\sim$40min & $\sim$110min & $\sim$240min\\
  \bottomrule
  \end{tabular}
  \vspace{-1mm}
  \caption{Quantitative Comparisons.}
  \label{tbl:app:restir:quan}
\end{table}

\end{document}